%% file: main.tex
\documentclass{article}
\usepackage{graphicx} 

\usepackage[utf8]{inputenc}
\usepackage[T1]{fontenc}

\usepackage{multirow}
\usepackage{booktabs}
\usepackage{graphicx}
\usepackage{subcaption}
\usepackage[justification=centering]{caption}
\usepackage[dvipsnames]{xcolor}

\usepackage{pgfplots}
\usetikzlibrary{calc}

\usepackage{amssymb} 
\usepackage{amsfonts}
\usepackage{amsmath}
\usepackage{amsthm}
\usepackage{hyperref} 
\hypersetup{
        colorlinks,
        citecolor=MidnightBlue,
        linkcolor=black,
        urlcolor=BrickRed,
}
\usepackage{xurl}  
\usepackage{mwe} 
\usepackage{lettrine}

\usepackage[switch,displaymath]{lineno}
\usepackage{rotating}
\usepackage{setspace}
\usepackage{tocloft}
\usepackage{subcaption}
\usepackage{graphicx}
\usepackage{colortbl}
\usepackage{doi}
\usepackage{caption}
\usepackage{hhline} 
\usepackage{authblk}
\usepackage{geometry}
\title{Leveraging point annotations in segmentation learning with boundary loss} 

\author[1]{Eva Breznik}
\affil[1]{Centre for Image Analysis, Department of Information Technology, Uppsala University, Sweden}

\author[2]{Hoel Kervadec}
\affil[2]{Department of Radiology and Nuclear Medicine, Erasmus MC, Rotterdam, Netherlands}
\author[1]{Filip Malmberg}
\author[3,4]{Joel Kullberg}
\author[3,4]{H\r{a}kan Ahlstr\"{o}m}
\affil[3]{Section of Radiology, Department of Surgical Sciences, Uppsala University, Sweden}
\affil[4]{Antaros Medical, M\"{o}lndal, Sweden}
\affil[5]{Department of Computer Science, University of Copenhagen, Denmark}
\author[2,5]{Marleen de Bruijne}
\author[1]{Robin Strand}

\date{}

\usepackage[dvipsnames]{xcolor}

\usepackage{todonotes}

\begin{document}
\maketitle

                \begin{abstract}
                      This paper investigates the combination of intensity-based distance maps with boundary loss  for point-supervised semantic segmentation.
                        By design the boundary loss imposes a stronger penalty on the false positives the farther away from the object they occur. 
                        Hence it is intuitively inappropriate for weak supervision, where the ground truth label may be much smaller than the actual object and a certain amount of false positives (w.r.t. the weak ground truth) is actually desirable. Using intensity-aware distances instead may alleviate this drawback, allowing for a certain amount of false positives without a significant increase to the training loss. 
                        The motivation for applying the boundary loss directly under weak supervision lies in its great success for fully supervised segmentation tasks, but also 
                        in not requiring extra priors or outside information that is usually required---in some form---with existing weakly supervised methods in the literature.
                        This formulation also remains potentially more attractive than existing CRF-based regularizers, due to its simplicity and computational efficiency.
                        We perform experiments on two multi-class datasets; ACDC (heart segmentation) and POEM (whole-body abdominal organ segmentation).
                        Preliminary results are encouraging and show that this supervision strategy has great potential. On ACDC it outperforms the CRF-loss based approach, and on POEM data it performs on par with it. The code for all our experiments is openly available at \url{https://github.com/EvaBr/geodesic_bl}. 
                  \end{abstract}



        \setlength{\parskip}{3pt}

        \section{Introduction}

 \lettrine{C}{onvolutional} neural networks (CNNs) have now long been a method of choice for various image processing tasks including image segmentation. However, they typically require a large amount of annotated ground truth data for training. For semantic segmentation---in the medical domain in particular---such annotations require expert knowledge and are very costly to obtain. This increased research interest in weakly supervised training, aiming to utilize approximate labels that are cheaper and faster to produce.

                Weak supervision comes in many different flavors, for example image-level labels (\cite{wei_2017_stc, Fan_2020_CVPR, dubost_2020_weak, Huang_2018_CVPR, ahn_2019_interpixel, xu_2015_variousweak, dubost2020weakly}), bounding boxes (\cite{kervadec2020bounding, dai_2015_boxsup, rajchl_2017_deepcut, xu_2015_variousweak, khoreva_2017_simple} ), scribbles (\cite{lin_scribblesup_2016, kervadec2019constrained, tang_2018_normalizedcut, xu_2015_variousweak, ji_2019_scribble}) or point annotations (\cite{qu_2019_nuclei, xu_2015_variousweak, lin_scribblesup_2016, bearman_whats_2016, dorent_2021_extreme}), to name a few.
                Most existing weakly supervised methods for image segmentation require some outside prior, additional information or regularization.  For instance, 
                \cite{kervadec2019constrained} and \cite{zhou_2019_organ} incorporate object size estimates or their distributions, 
                while \cite{kervadec2020bounding} introduces a tightness prior that has some strictness guarantee on the drawn bounding boxes. 
               
                With the aim to improve the accuracy of segmentation methods, distance maps have been employed in a multitude of ways (\cite{ma_2020_dtboost}).
                An example of a very direct inclusion of a distance map to guide the CNN training is boundary loss by \cite{kervadec2021boundary}, introduced in the context of fully supervised segmentation (focused on but not limited to imbalanced tasks). It uses a Euclidean distance transform of the ground truth to minimize the distance between the ground truth and the predicted segmentation boundaries, at training time.
                This proved to be very effective while remaining computationally lightweight and compatible with different network architectures, optimization strategies and other losses.

                In this paper, we investigate the use of boundary loss in a weakly supervised setting and propose a way to make it directly compatible with training on point annotations without requiring architectural changes or modifications to the training procedure. 
                We propose to replace the Euclidean distance map used in the original boundary loss paper with intensity-aware ones, taking pixel intensities into account when computing the distance from the point annotations.
                In applications where the object to segment has a fairly homogeneous intensity  and a decent contrast around the object boundary, this can provide the network with a better notion of the region extent and shape, and thus address the incompatibility between boundary loss and weak labels. It allows for end-to-end training without needing to explicitly formulate priors or additional data-dependent  information.

                Our proposed approach is evaluated using various intensity-aware distances, on two multi-class segmentation tasks with artificially created point annotations. We show that training  with the combination of cross entropy and boundary loss is not only compatible with point-level supervision but reaches competitive results compared to a CRF-loss based training 
                and even compared to full supervision. 

        \section{Related works}



            \subsection{Point and scribble supervision}
                While learning from weak annotations has seen an increased amount of research interest in recent years, segmentation methods developed specifically for point supervision are still scarce. On the other hand, many more methods are available for scribble supervision, of which point supervision can be viewed as a special case (as the scribble lengths tend to zero).  

                A separate line of works focusing on point annotations requires the points to be extremal, meaning they lie at the boundary (\cite{roth_2021_extremes, dorent_2021_extreme}). This usually requires multiple points per object and can be viewed as introducing  additional (extent) information. Moreover, it  induces a higher annotation cost. We thus omit it in this study.
                \cite{xu_2015_variousweak} present a unified approach for various types of weak supervision like bounding boxes, image tags and partial labels. Their approach is based on first oversegmenting the image to superpixels, then using max-margin clustering with weak annotations as constraints. 
                ScribbleSup from (\cite{lin_scribblesup_2016}) makes use of a superpixel-based graphical model for annotation propagation jointly with a CNN based segmentation model, optimizing them alternately. While the method was developed specifically for training with scribbles, they report good results even on point annotations. 
                However, creating synthetic and inaccurate ground truth to use during training can easily propagate errors, possibly creating instabilities.
                Similarly \cite{ji_2019_scribble} create pixel-level labels from scribbles using Graph Cut and use them for training a U-Net. However they do not iterate this step but instead apply a global-label guided clustering on the output. 
                Using point-level supervision, \cite{bearman_whats_2016} introduce a special loss based on both global image-level labels and weighted full supervision on annotated pixels. The loss is further supplemented by an objectness prior, which requires independent supervised pretraining.
                   
                Conditional random field (CRF) based losses presented in \cite{tang2018regularized} and \cite{zheng_conditional_2015} implement the seminal work of \cite{krahenbuhl_efficient_2011}, and using such losses to add regularization has been shown to achieve good results under weak supervision. 
                \cite{tang_2018_normalizedcut} propose a kernel cut loss combining CRF and normalized cut terms and apply it together with cross entropy loss for semantic segmentation tasks with scribble annotations.                
                In \cite{zheng_conditional_2015}, dense CRFs are reformulated as RNNs, to achieve an end-to-end trainable segmentation system. 
                Both mentioned CRF based methods however significantly slow down the training time per iteration.
                While many works (e.g. \cite{lin_scribblesup_2016, ji_2019_scribble}) actually also use CRF as a postprocessing step to further improve the segmentation results, \cite{tang2018regularized} show that incorporating it as loss can still be advantageous.            
                Also based on point annotations and CRF loss, \cite{qu_2019_nuclei} create two coarse segmentations to be used in training: a voronoi diagram based one as a lower bound (undersegmentation) and a clustering based one as an upper bound (oversegmentation). They show that training with both segmentations jointly, with the addition of a CRF loss, performs well on nuclei segmentation in histopathology images.

                \subsection{Non-Euclidean distance maps in segmentation tasks}
                Among works employing distance maps to boost segmentation, 
                \cite{criminisi_2008_geos} apply a Geodesic distance based filtering operator to produce a set of smooth segmentation proposals, a viable subset of which is then searched for the best, energy minimizing  labelling.
                In \cite{bai_2009_geomatting} and \cite{gulshan_2010_star} the authors make use of Geodesic maps for scribble based segmentation, but in an interactive setting. While \cite{gulshan_2010_star} introduce a geodesic star convexity shape constraint computed on (intensity/colour-based) likelihood maps, the geodesic matting framework from \cite{bai_2009_geomatting} calculates the Geodesic map on the space of the class probability densities and models class distributions based on the user scribble statistics. 
                Also in a purely interactive setting, \cite{wang_2019_deepigeos} use the Geodesic maps from user provided scribbles as additional input channels during CNN refinement.            
                \cite{mortazi_2019_geodesicprior} propose using Geodesic map priors to improve robustness when training with noisy labels. Their method consists of first training an autoencoder to regress ground truth annotations based on the Geodesic maps, then using the mean square error between the encodings of segmentor network probabilities and distance maps as a part of the loss during segmentor network training. 
                Their approach requires additional training time and while they work with noisy labels, they don't cover severe label degradation such as scribble or point annotations. 
                In \cite{onvu_2020_mbd} the authors propose a vectorial Dahu pseudo-distance, based on the Minimum barrier distance. While developed and evaluated mainly for saliency detection, it is successfully used also in white-matter segmentation with simple seeding and thresholding.
                
        \section{Method}
        \subsection{The boundary loss}
                Boundary loss from \cite{kervadec2021boundary} has been so far  successfully applied in CNN training for fully supervised segmentation tasks.  It is calculated by computing a signed distance map of the ground truth, and then multiplying it with the network output.

                Formally, we have $\Omega \subset \mathbb R^{D}$ as $D$-dimensional image space, with $X: \Omega \rightarrow \mathbb R^M$ an image with $M$ modalities, and $Y: \Omega \rightarrow \mathcal K$ its corresponding multi-class ground truth, with $\mathcal K = \{0: \verb|background|, 1: \verb|first class|, ..., K: \verb|last class|\}$.
                For simplicity, we will denote $Y^{(k)} := \{ i \in \Omega | Y(i) = k \}$ the subset of $\Omega$ containing all the voxels belonging to the class $k$. It follows that $\cup\{Y^{(k)}\}_{k\in \mathcal{K}}=\Omega$, and $ Y^{(i)}\cap Y^{(j)} = \emptyset$ for any pair of classes $i\neq j; i,j \in \mathcal K$ (i.e. they do not overlap).

                In the original work on boundary loss\footnote{The original paper did not evaluate multi-class settings, but the formulation was already compatible with such settings.}, the signed distance map for each class $k$  is computed as follows:
                \begin{equation}
                        \label{eq:levelset}
                        \forall i \in \Omega: \phi^{(k)}_Y(i) = \begin{cases}
                                        -D_\text{euc}^{(k)}(i) & \text{if } i \in Y^{(k)}; \\
                                        D_\text{euc}^{(k)}(i) & \text{otherwise,}
                                \end{cases}
                \end{equation}
                with $D_\text{euc}^{(k)}(\cdot): \Omega \rightarrow \mathbb R_+$ denoting the Euclidean distance map from the boundary of the ground truth annotation for class $k$. Strictly at the annotation boundary $\partial Y^{(k)}$, the value of the distance map is $0$ (with the boundary $\partial Y^{(k)}$ here denoting all elements of $Y^{(k)}$ that have a neighbor outside this set). 

                Then, this distance map is used as-is inside the boundary loss:
                \begin{equation}
                        \label{eq:boundaryloss}
                        \mathcal L_B(s_{\boldsymbol\theta}, Y) = \sum_{k \in \mathcal K} \sum_{i \in \Omega} s_{\boldsymbol\theta}^{(i, k)}\phi_Y^{(k)}(i),
                \end{equation}
                where $s_{\boldsymbol\theta}$ represents the network output probabilities.

                \subsection{From full- to point annotation}
\begin{figure*}
                        \centering
                        \begin{subfigure}[b]{0.19\textwidth}
                                \centering
                                \includegraphics[width=\textwidth]{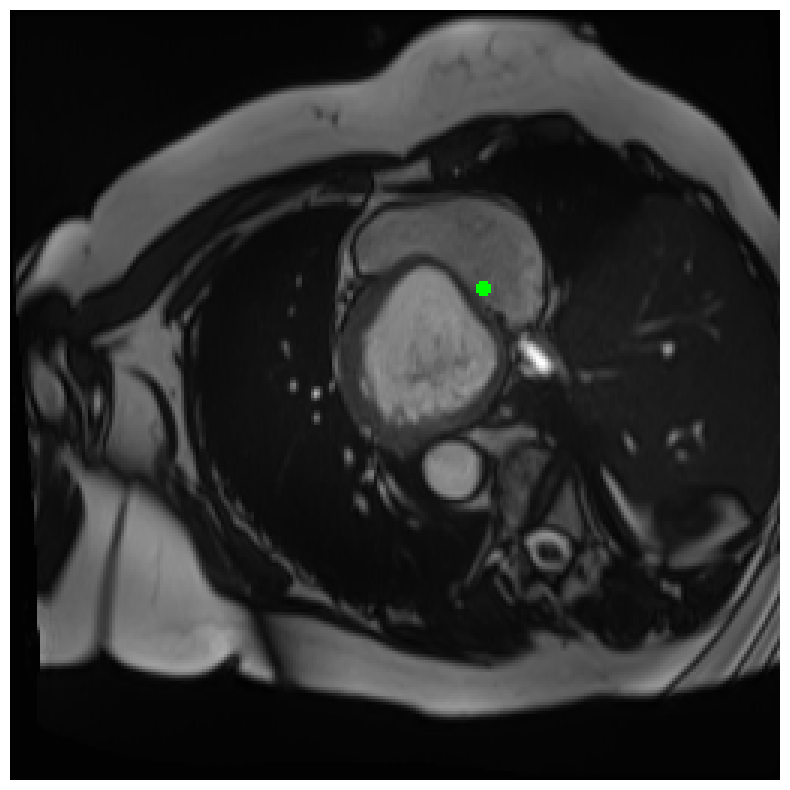}\\
                                \includegraphics[width=\textwidth]{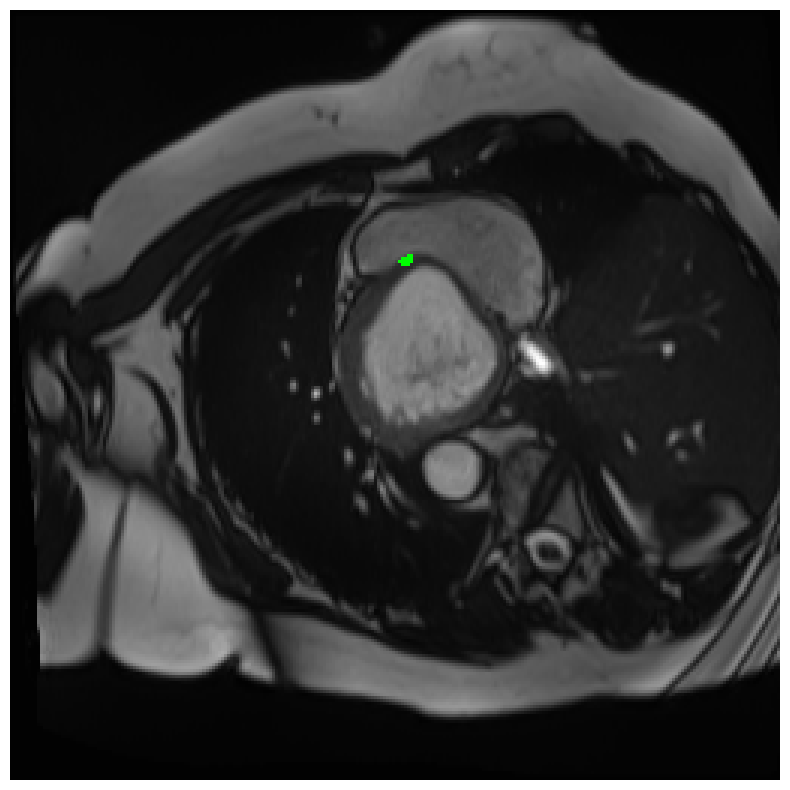}\\
                                \includegraphics[width=\textwidth]{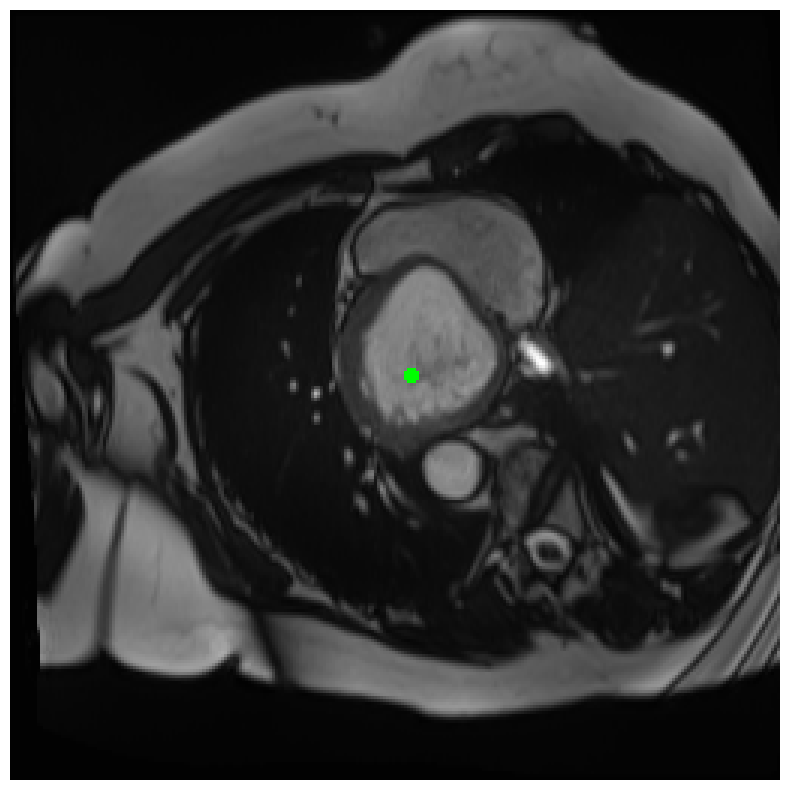}
                                \caption{Point ground truth}
                                \label{subfig:gt}
                        \end{subfigure}
                        \begin{subfigure}[b]{0.19\textwidth}
                                \centering
                                \includegraphics[width=\textwidth]{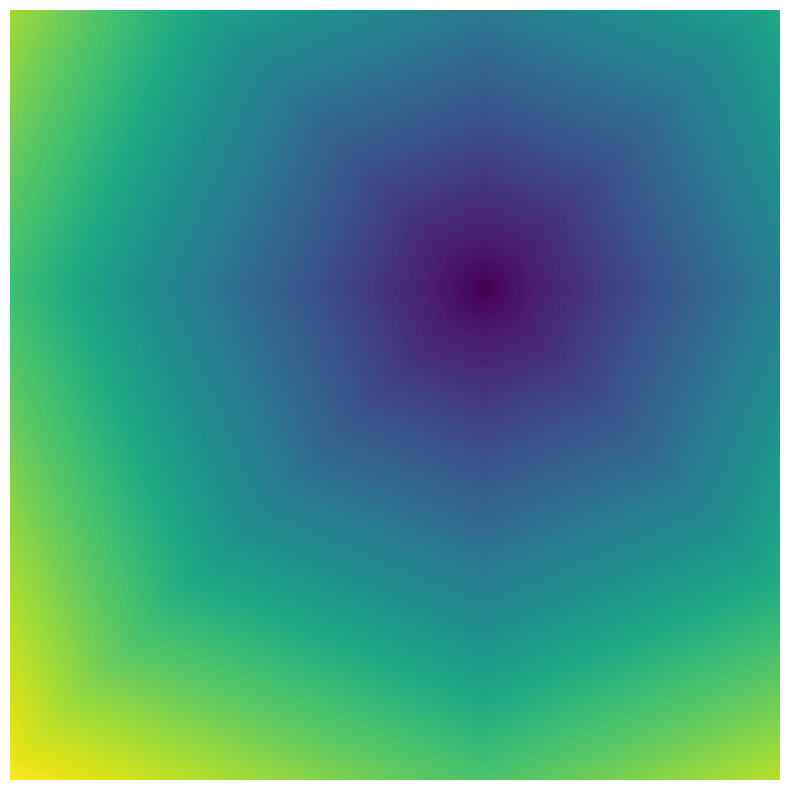}\\
                                \includegraphics[width=\textwidth]{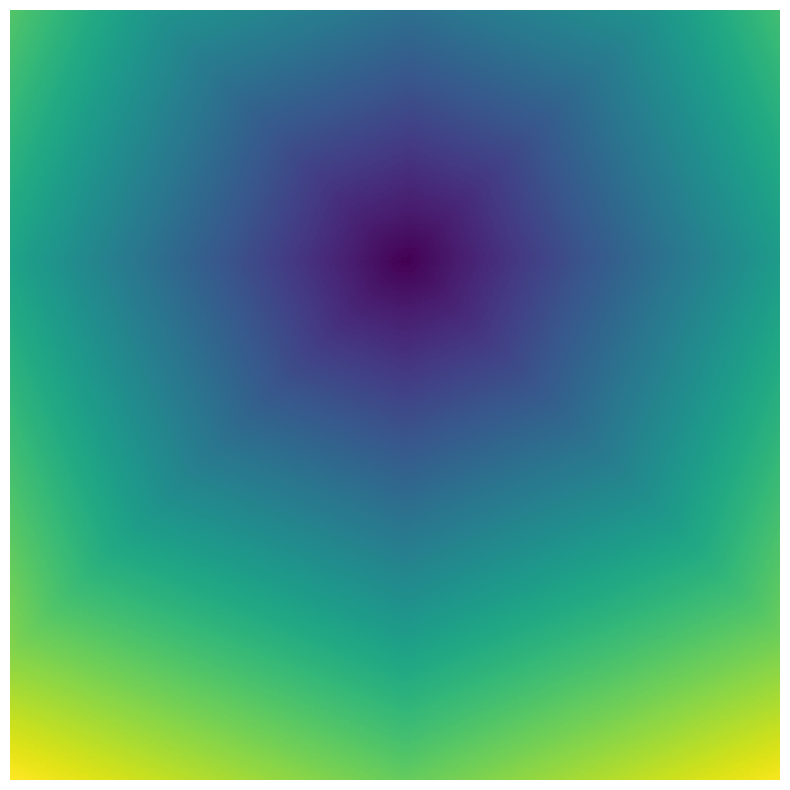}\\
                                \includegraphics[width=\textwidth]{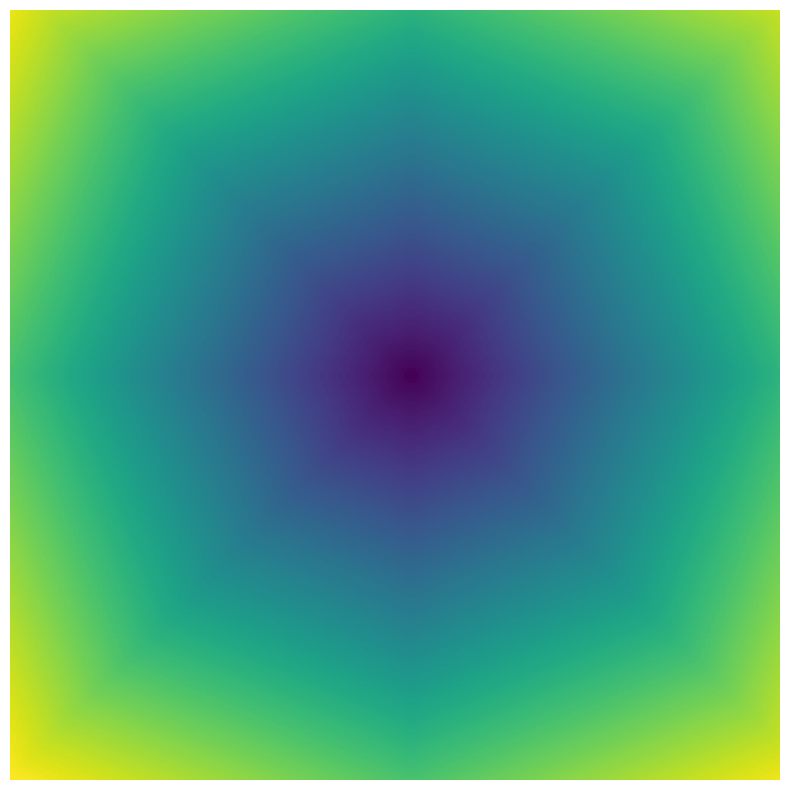}
                                \caption{$D_\text{euc}$ maps}
                                \label{subfig:ecl}
                        \end{subfigure}
                        \begin{subfigure}[b]{0.19\textwidth}
                                \centering
                                \includegraphics[width=\textwidth]{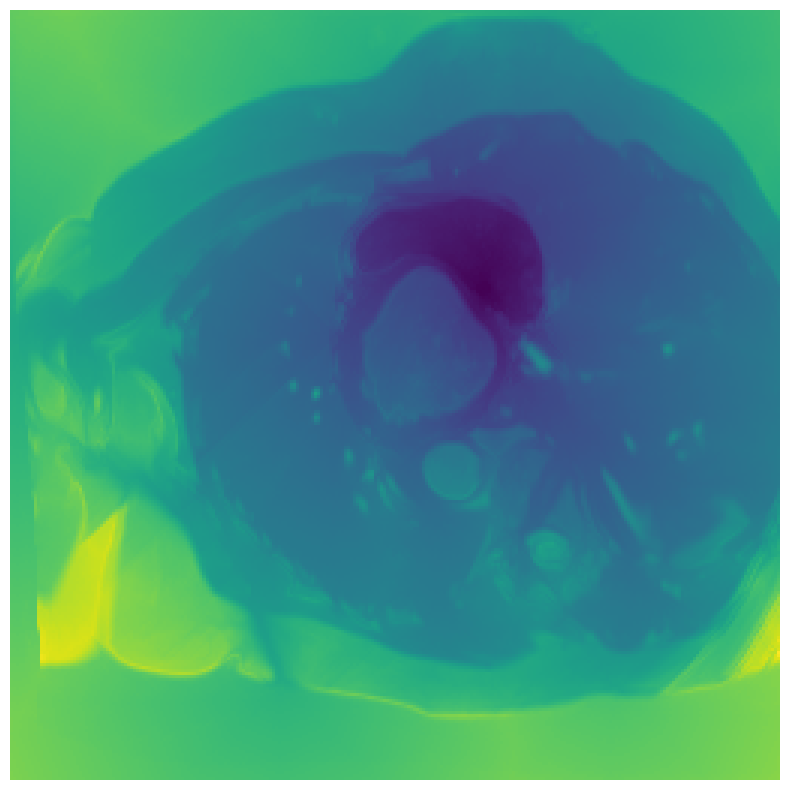}\\
                                \includegraphics[width=\textwidth]{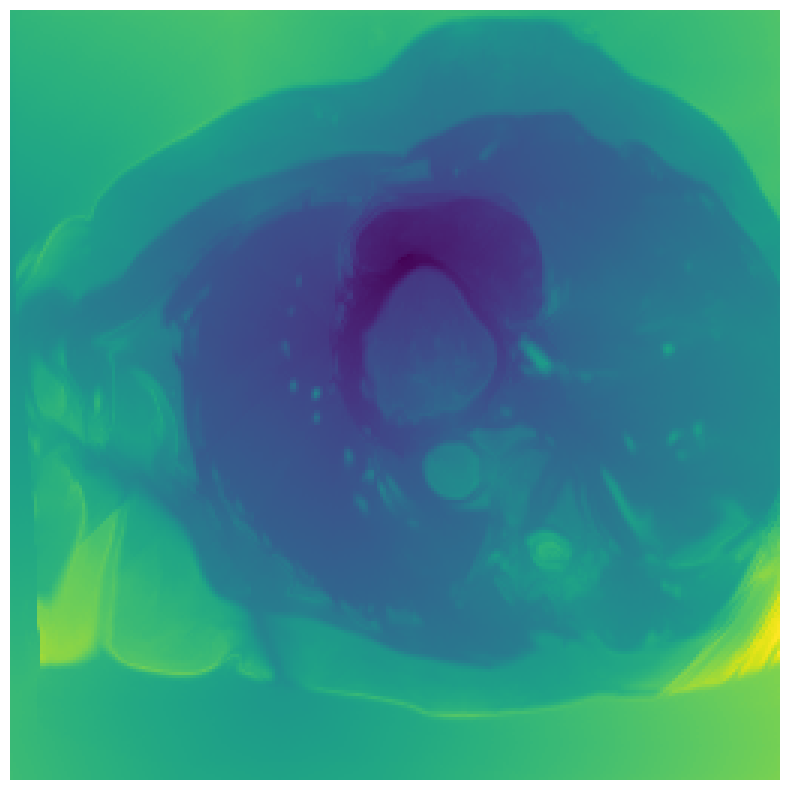}\\
                                \includegraphics[width=\textwidth]{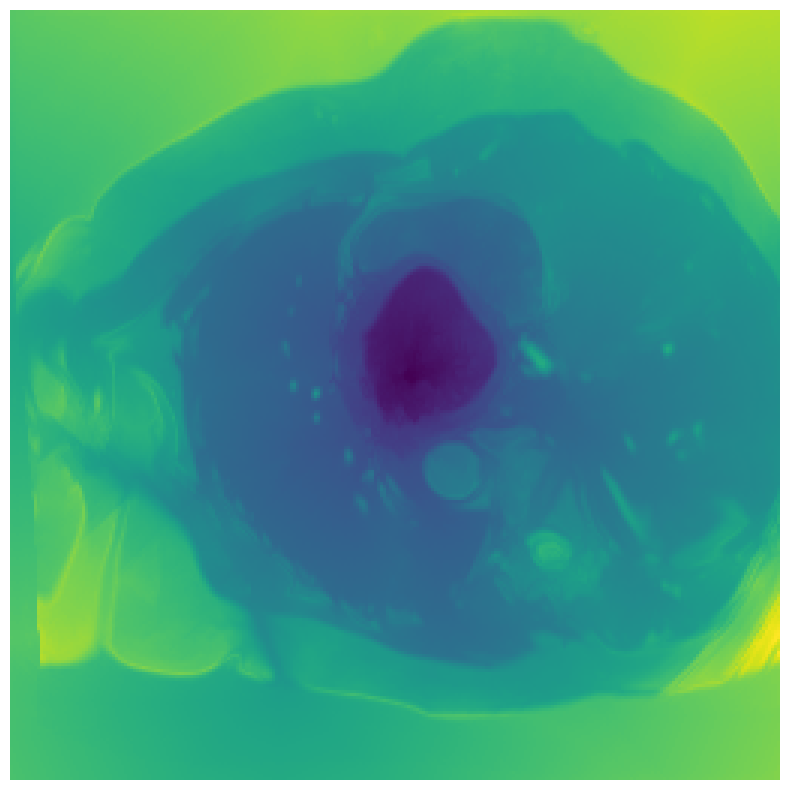}
                                \caption{$D_\text{geo}$ maps}
                                \label{subfig:geo}
                        \end{subfigure}
                        \begin{subfigure}[b]{0.19\textwidth}
                                \centering
                                \includegraphics[width=\textwidth]{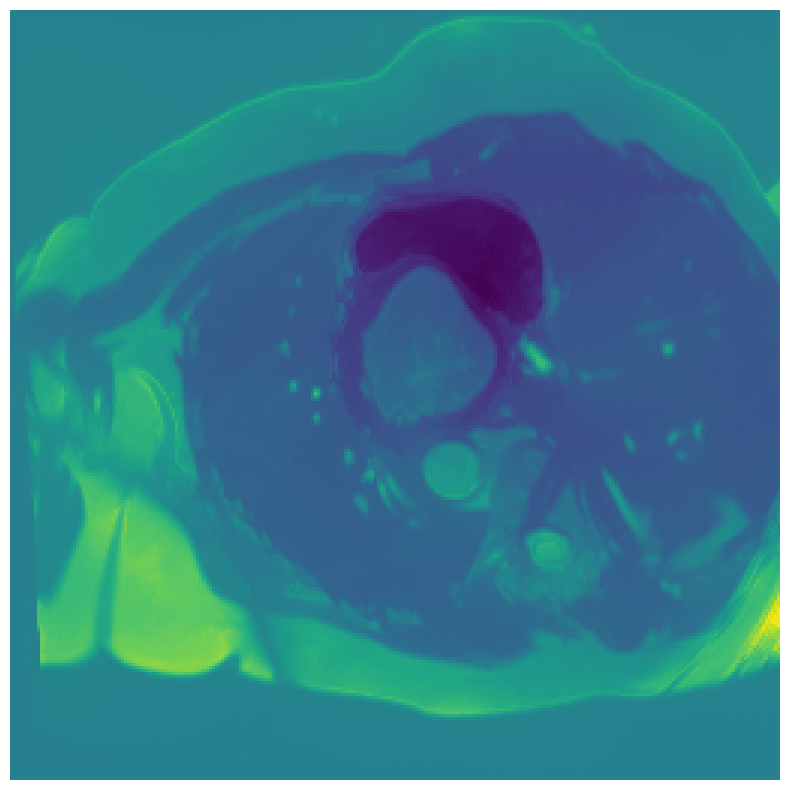}\\
                                \includegraphics[width=\textwidth]{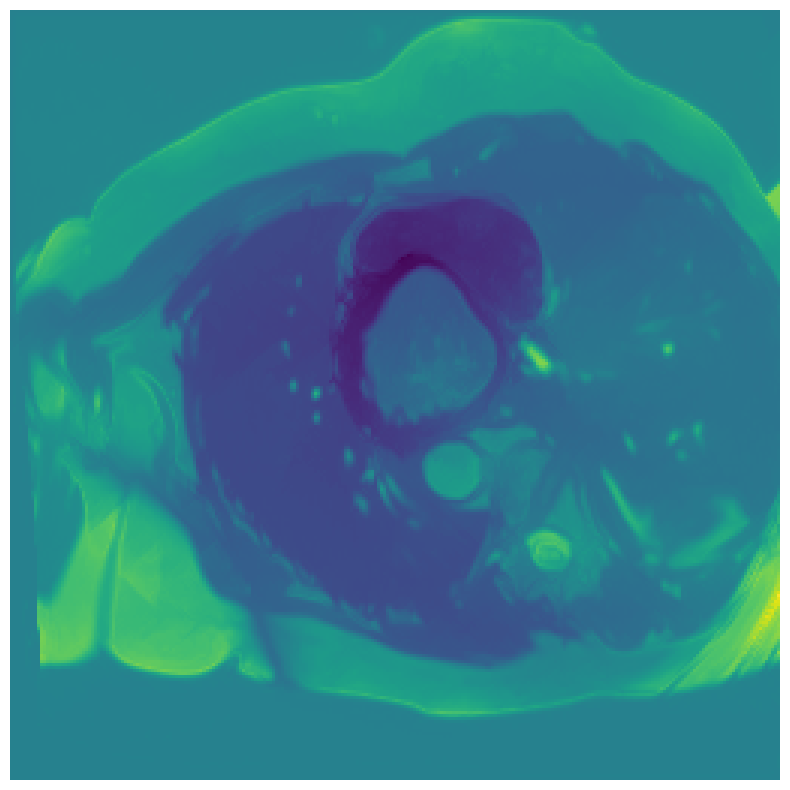}\\
                                \includegraphics[width=\textwidth]{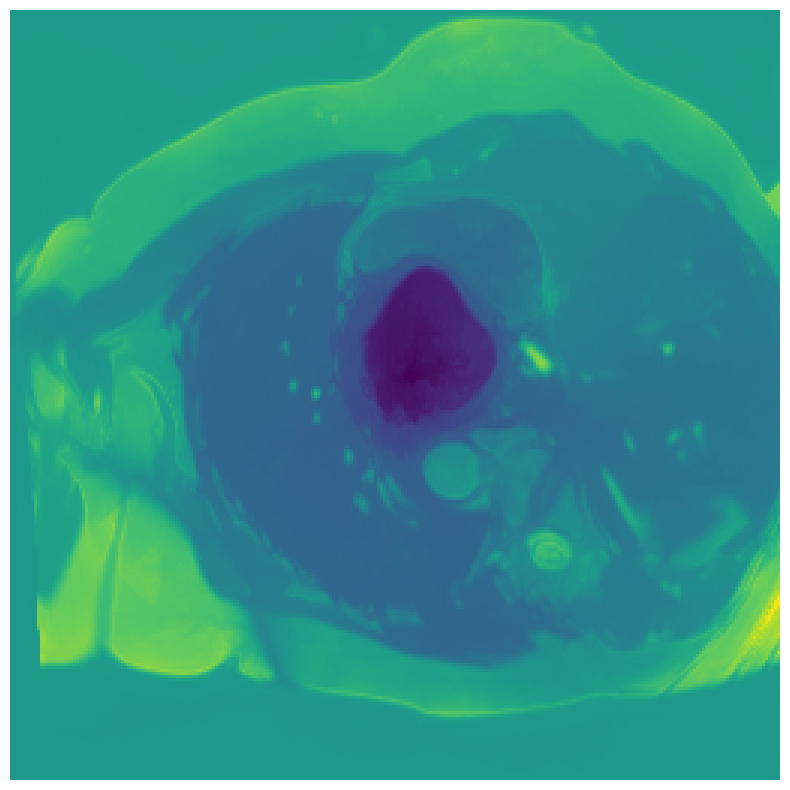}
                                \caption{$D_\text{int}$ maps}
                                \label{subfig:int}
                        \end{subfigure}
                        \begin{subfigure}[b]{0.19\textwidth}
                                \centering
                                \includegraphics[width=\textwidth]{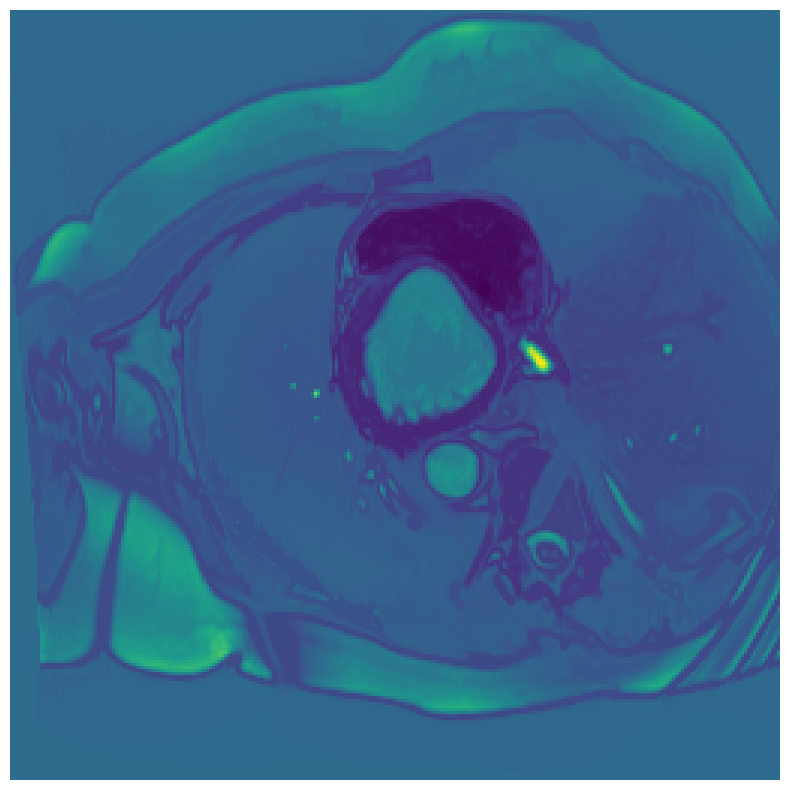}\\
                                \includegraphics[width=\textwidth]{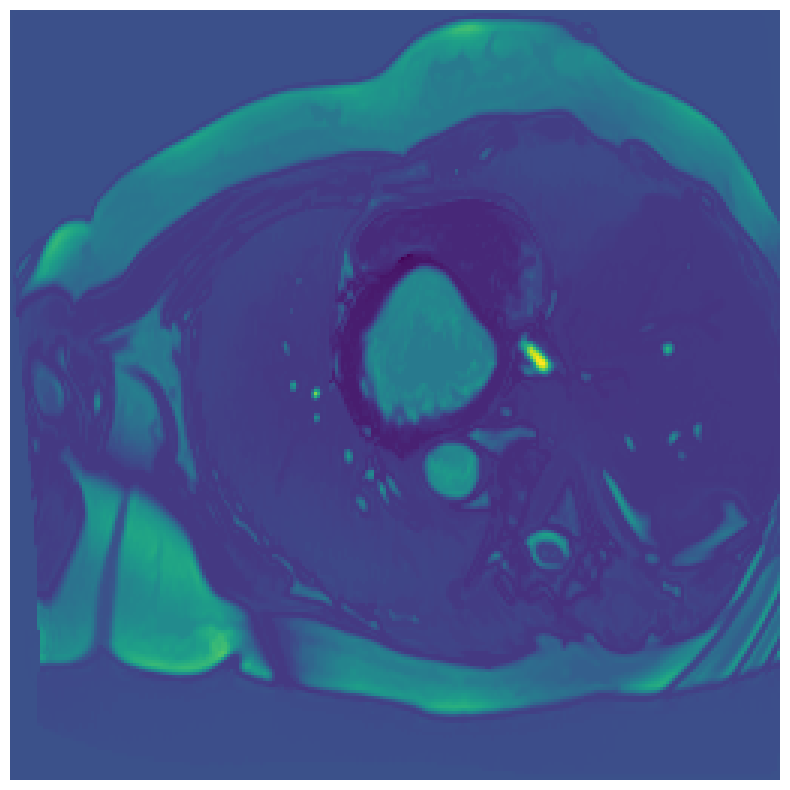}\\
                                \includegraphics[width=\textwidth]{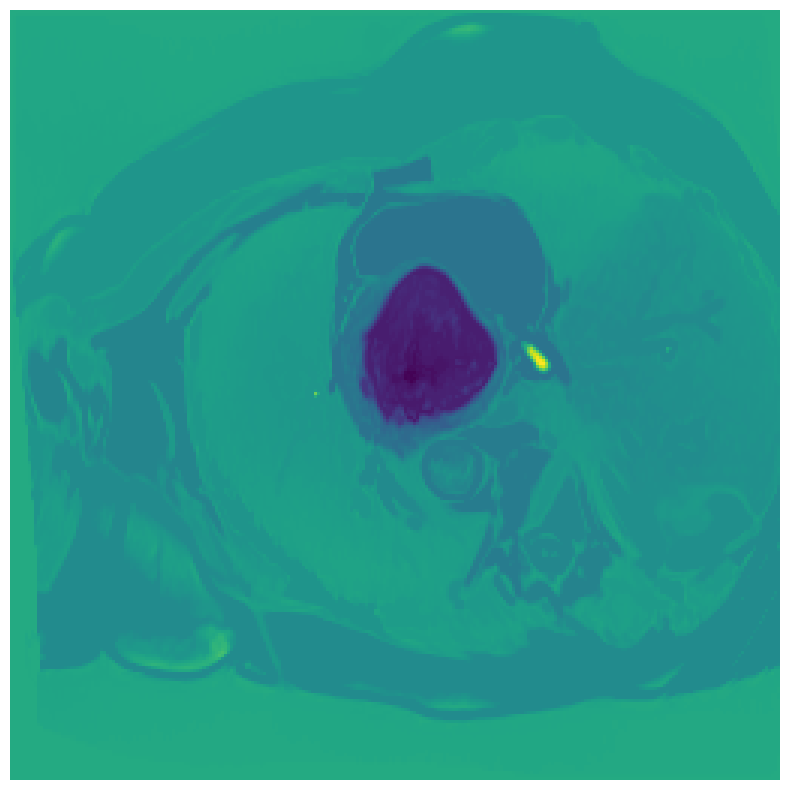}
                                \caption{$D_\text{mbd}$ maps}
                                \label{subfig:mbd}
                        \end{subfigure}
                \caption{Comparison of the different computed distance maps on an example slice from the ACDC dataset (\cite{bernard2018deep}). The maps are computed from the point labels shown in first column. The three rows show the ground truth and corresponding distance maps for the right ventricle, myocardium and left ventricle respectively.}
                \label{fig:distmaps}
                \end{figure*}

                Generally, given an exact, full annotation of an object, such a Euclidean signed distance map encodes also information on the shape of an object. But when using a weak ground truth  $\widetilde Y: \Omega \rightarrow K $, where $\widetilde Y^{(k)} \subset Y^{(k)}$ and $\cup \{ \widetilde Y^{(k)} \}_{k \in \mathcal K} \neq \Omega$, 
                inferring a correct extent and shape of the objects is nontrivial. As shown in \autoref{subfig:ecl}, Euclidean distance calculated from a point label source grows radially, regardless of the actual shape of the object, and thus makes little sense from an information point of view.
%
%
%
                Under the assumption of intra-object homogeneity and inter-object contrast (w.r.t intensities), this problem can be circumvented to a degree by using a distance function that takes also intensity values into account. 
                An example of a commonly used distance measure with an intensity component is the Geodesic distance (\cite{toivanen_1996_geo}). Let $\pi_{x,y} = \langle x=p_0, \ldots,p_i,p_{i+1},\ldots, p_n=y \rangle$ denote a path between $x,y\in\Omega$, with $p_i$ and $p_{i+1}$ being neighbors under a chosen adjacency relation. Reusing the notation from before, a Geodesic distance map from the boundary of the ground truth class $k$, $D_\text{geo}^{(k)}(\cdot): \Omega \rightarrow \mathbb R_+$, can be defined as 
                 \begin{equation}
                        \label{eq:geodesic}
                        D_\text{geo}^{(k)}(\cdot) = \min_{\pi\in\Pi_k(\cdot)}\sum_{p_n, p_{n+1}\in \pi} \sqrt{(1-\lambda)\Delta I_{n,n+1}^2+\lambda c_{n,n+1}^2}
                \end{equation}
                where $\Pi_k(\cdot) := \cup_{j\in \partial Y^{(k)}}\pi_{\cdot,j}$ and $\Delta I_{n,n+1}=I_n-I_{n+1}$ is the intensity difference between $p_n, p_{n+1}\in\Omega$. The value of $c_{n,n+1}$ is the distance between $p_n$ and $p_{n+1}$ in space, and depends on the type of adjacency between them (in 3D for example, it is equal to $1$, $\sqrt{2}$ or $\sqrt{3}$ if $p_n$ and $p_{n+1}$ share a face, an edge or a vertex respectively). The parameter $\lambda\in[0,1]$ allows for balancing the contributions of intensities and spatial proximity. 
               
                In practice, the Geodesic distance is often implemented using a weighted $L_1$ distance instead. That means changing the expression under the sum in Eq.\eqref{eq:geodesic} to $(1-\lambda)|\Delta I_{n,n+1}|+\lambda c_{n,n+1}$.
                This definition is adopted also throughout this paper, as it is easier and faster to compute and locally approximates the original one up to the next greater integer number (see for example the implementation of \cite{asad2022fastgeodis}). 
                
                While setting the $\lambda$ parameter to $0$ actually results in a taxicab distance, which can be seen as a discrete approximation of a Euclidean distance, setting $\lambda = 1$ focuses purely on image intensities. We call the latter simply Intensity distance, $D_\text{int}$. From here on we shall use the term Geodesic distance, $D_{geo}$, to denote the setting of $\lambda=0.5$, and Euclidean, $D_{euc}$, to denote the setting $\lambda=0$.

                Entirely on the other side of the spectrum from the Euclidean are the fully intensity based distances, such as the Minimum barrier distance (MBD) from \cite{strand_2013_mbd}. It is not only calculated exclusively on the image intensity space, but also effectively independent of the path length in space. 
                Given the definition of $\Pi_k(\cdot)$ above, the MBD map from the boundary of ground truth class $k$, $D_\text{mbd}^{(k)}(\cdot): \Omega \rightarrow \mathbb R_+$, can be defined as:

                \begin{equation}
                        \label{eq:mbd}
                        D_\text{mbd}^{(k)}(\cdot) = \min_{\pi\in\Pi_k(\cdot)}\left(\max_{p_i\in\pi}I_{p_i} - \min_{p_j\in\pi}I_{p_j}\right) .
                \end{equation}

                MBD is actually only a pseudo-distance, as the reflexivity property may be violated (i.e. it may be that $D_\text{mbd}^{(k)}(x)=0$ even when $x\notin Y^{(k)}$). As opposed to full supervision, this can be a very desirable property in the case of weak labels.


                Both the Intensity and the Minimum barrier distance are defined exclusively on the image intensity space. However, from the examples of $D_\text{int}$ distance map in \autoref{subfig:int}, we can notice that the values still increase somewhat radially from the annotation. This behaviour is similar to the one of the Geodesic distance in \autoref{subfig:geo} (which actually includes the spatial proximity in its definition), and is due to the summing operator in the general Geodesic distance definition in \autoref{eq:geodesic}. While the intensities of two neighboring pixels on a path may be the same, that will rarely be the case in real life, noise riddled images. This makes the Intensity distance function approximately monotonically increasing with increasing length of the path (in space), even on paths where the intensity is mildly fluctuating (e.g. consider a path with even pixels intensity value of $n$ and odd pixels intensity value of $n+1$). One could thus argue that such a definition of a distance, despite being based exclusively on intensities, is still capable of loosely encoding the spatial distance information. 

                On the other hand, while we can see that the MBD based maps are similar to Geodesic and Intensity ones (\autoref{subfig:mbd}) with respect to the object shape recovery, they have a less pronounced and smooth increase in the values outward from the source point. 

                In contrast to Euclidean distance, the Geodesic, Intensity and Minimum barrier distance maps all encode contrast sensitivity and preserve the object structures by harnessing the intensity information of the underlying image. This holds even when calculated from point sources. In practice, using such maps for network training could mean a lower penalty for false positives that occur farther from the point annotation but are close to it in intensity. Thus still enabling the propagation of a sort of shape information (as it can be inferred from the raw image intensities).
                
                 If we were to use the boundary loss on the point labels alone, very few pixels would be \emph{positively} supervised: their probability is pushed \emph{up} only when the distance map is negative, i.e. on the exact dot annotation. We solve this minor issue by combining it with a partial cross-entropy, $\mathcal L_{\widetilde{\text{CE}}}$, which results in the following model to optimize:
                \begin{equation}
                        \min_{\boldsymbol\theta} \sum_{k \in \mathcal K} \sum_{i \in Y^{(k)}} -\log(s_{\boldsymbol\theta}^{(i, k)}) + \alpha \mathcal L_B,
                \end{equation}
                with $\alpha \in \mathbb R$ balancing the two losses. 
                To differentiate between the boundary loss computation with different distances, we use $\mathcal L^d_\text{B}$ to denote the boundary loss computed with the distance metric $d \in \{$euc, int, geo, mbd$\}$.

                \subsection{Practical considerations in distance choice}
                \paragraph{General} 
                The presented definitions of the distances are specific to single-channel images, and their extensions to multi-channel images are not always trivial (see e.g. \cite{onvu_2020_mbd, vectorialMBD}), and can depend on the application.
                In addition, all of them depend on the chosen connectivity pattern (i.e. which neighbourhood of pixels or voxels is considered in the path search). While varying the connectivity might not adversely affect the use of the distances within boundary loss supervision, we suggest using the full connectivity to mimic the way tissues expand over images. To make sure the distance is not biased toward a specific direction, image sampling (voxel size) needs to be taken into account for those distances that accumulate over paths ($D_\text{euc}$, $D_\text{geo}$, $D_\text{int}$). Since many medical datasets have a larger spacing in the third dimension, this is particularly important when computing the distances om 3D volumes.
                
                Depending on the imaging data, a certain amount of robustness to noise or blur may be desired.
                In $D_\text{int}$, the intensity values (and with that, the present noise) accumulate over paths. This can result in high sensitivity to noise.
                Similar holds for $D_\text{geo}$, albeit to a lower extent, as the intensity contribution is smaller.
                As shown in \cite{strand_2013_mbd}, the MBD is more robust to noise and blur, but may bleed out/propagate low distance values when the contrast between neighbouring classes is not large.
                Another thing to consider when dealing with point annotations (i.e. point sources for computing the distance maps) is the distance stability with respect to the change of point positions within the objects. For large objects, the radially increasing $D_\text{euc}$ will produce severely different maps when calculated from points at opposite ends of the object. On the other hand, given a reasonably homogeneous intensity profile of the object the $D_\text{int}$ and $D_\text{mbd}$ should be more agnostic towards the point locations.

                \paragraph{Ranges}
                The value ranges of the different distance maps depend on the image size and/or intensity range. 
                Given a one channel square image of size $N\times N$ with intensity range $[0,255]$ and using full connectivity, the maximal possible range of   $D_\text{mbd}$ will be $[0, 255]$, while the ranges of $D_\text{int}$, $D_\text{geo}$ and $D_\text{euc}$ all depend on $N$.  While the maximal possible value of a $D_\text{euc}$ is of order $N$, it can reach up to the order of $255\cdot N$ for $D_\text{int}$ and $D_\text{geo}$.

                
                Moreover, assuming that the point annotation denoting the object of interest is somewhere close to the center, the possible $D_\text{euc}$ and $D_\text{int}$ value ranges effectively halve.  $D_\text{geo}$ range shrinks as well while $D_\text{mbd}$ range is unaffected by the image size or object position. 

                This is important to note as the difference in ranges can affect the training speed or hyperparameter (learning rate) settings when used in the boundary loss.
                The differences between the distance ranges can get even more pronounced when using 3D images, due to the increase in size. 

                To make results comparable over trainings with different distance maps, the image intensities can be rescaled prior to distance computation, such that the ranges of space- and intensity-based distances are more similar. Such rescaling is particularly important also within the Geodesic map computation, as extreme differences in the intensity and space ranges will result in one of these contributions being severely dominant. In \cite{gulshan_2010_star} for example, the authors rescale the intensities such that the average intensity and Euclidean differences between neighbouring pixels are the same. Alternatively, the contributions can be balanced by a careful tuning of $\lambda$.

                Even within a single distance map, different images (even different slices of the same subject) can attain very different ranges. But those differences carry important information, hence it is not advisable to directly normalize them; especially not on individual slice image basis.
                
                \paragraph{Slice vs. volume}
                As described also in \cite{kervadec2021boundary}, training with boundary loss can actually be improved by calculating the distance maps in entire 3D volumes instead of on individual slices. However, depending on the distance choice, there may be drawbacks connected to it. 
                
                For example, due to the noise in the image MBD is prone to propagate values out of the objects, and might thus actually find much shorter (but biologically nonsensical) paths inside the 3D volume. 
                And for the $D_\text{euc}$, $D_\text{geo}$  and $D_\text{int}$, where the local contributions are summed up throughout the path, the distance values can grow unmanageably large.
                
                On the other hand, working with 2D slices of volumetric scans means not all classes might  be present in the ground truth of every image. And it is not clear what value to set the distance map from an absent object to. While setting it to infinity for all pixels seems sensible from a theoretical perspective, it does not work well with the fact that the slices come from volumes where the object in question actually exists. Additionally, using distance maps with such extremely large values with CNNs may again lead to exploding gradients and unstable training.
                Simply setting the map to zero (which is often done in publicly available implementations, see e.g. \cite{asad2022fastgeodis}), however, incurs zero penalty for arbitrarily large oversegmentations.

        \section{Experiments}
                \subsection{Datasets}
                        \paragraph{ACDC} (\cite{bernard2018deep}) The Automated Cardiac Diagnosis Challenge (ACDC)  is a public benchmark multi-class heart segmentation dataset. It contains cine-MR images of 150 patients (of which 100 are available for training and the rest 50 comprise a test set), covering healthy scans and four types of pathologies in equal amounts, 
                        with annotations for the right ventricle (RV), myocardium (Myo) and left ventricle (LV) heart structures.
                        We split the training set randomly, using 65 subjects for training, 10 for validation and 25 as a hold-out test set.
                        Due to the large and varying interslice gap, we work with 2D slices instead of 3D volumes directly. This includes distance map computation.

\begin{figure}
                        \centering
                        \begin{subfigure}[b]{0.2\columnwidth}
                                \centering
                                \includegraphics[width=\textwidth]{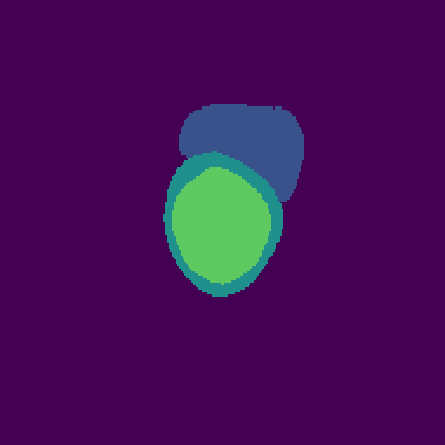}
                                \caption{Ground truth}
                                \label{subfig:gtPOEM}
                        \end{subfigure}
                        \begin{subfigure}[b]{0.2\columnwidth}
                                \centering
                                \includegraphics[width=\textwidth]{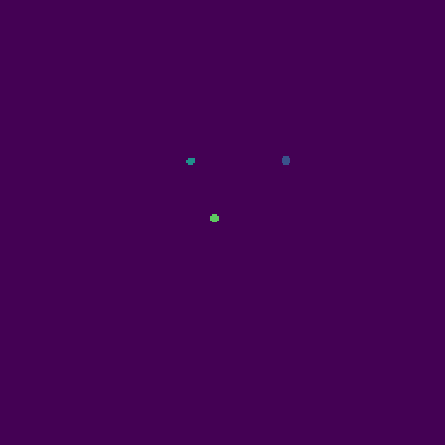}
                                \caption{Point annotations}
                                \label{subfig:point}
                        \end{subfigure}
                        \caption{Comparison between a full ground truth and synthetically generated point annotations, on the ACDC dataset \cite{bernard2018deep}.}
                        \label{fig:gt_dots}
                \end{figure}

                        We normalize the volumes and resize the slices to $256\times256$ pixels. As the official dataset comes with full annotations, we create a synthetic point ground truth. This is done by first randomly choosing the centers of the point annotations within the class masks, followed by filling an ellipse with axes lengths of $4$ and $2$ (in pixels) around each center. The intersections of these elliptic discs with the underlying full annotations are then used as our point ground truth. 
                        See \autoref{fig:gt_dots} for an example of the created weak annotation mask. The point annotations are created for every slice, one for each foreground object present in the slice.   
                       
                         \paragraph{POEM} The Prospective investigation of Obesity, ENergy production and Metabolism (POEM) is a local (not currently publicly available; PI: L. Lind, see \cite{lind_2013_poem} for details) cohort of whole-body fat/water separated MR images. Full annotations of the liver, kidneys, bladder, pancreas and spleen are available for 50 subjects, providing a challenging segmentation dataset with heavily imbalanced classes of varying shapes. The resolution of the data is anisotropic, with reconstructed voxel size of $2.07 \times 2.07 \times 8.0 \text{mm}^3$ in left-right, anterior-posterior and foot-head directions, respectively.
                        For additional technical details regarding the acquisition and specifications of the images see \cite{lind_2013_poem}.

                        The images contain two channels, one for water and one for fat content. 
                        For training, we normalize the volumes (per channel) and use 2D slices in the coronal plane, sized $256\times 256$. 
                        The weak annotations are created synthetically, following the same procedure as described for the ACDC dataset.

              
        \subsection{Distance map computation}
                For computing the intensity-aware distance maps, the image intensities are scaled to 0-255, in order to ensure that the spatial distances and intensity differences between voxels are comparable in 2D. We use full connectivity  in all distance computations. 

                For both datasets, the distance maps are computed prior to training, and on POEM data, they are computed on the fat content channel only. 
                For the approximate Euclidean, Geodesic and Intensity distance, we use the FastGeodis implementation of \cite{asad2022fastgeodis}. For the Minimum barrier distance, we use our own implementation\footnote{The base code is available at \url{https://github.com/FilipMalmberg/DistanceTransforms}}. The computation time for the signed versions (as used in the boundary loss definition, \autoref{eq:boundaryloss}) of all the distances is reported in \autoref{tab:timetocompute}. It should be noted that the required times for 2D map computations appear lower for POEM due to a large number of slices containing only few or no foreground classes.  

                For ACDC, we compute the maps on 2D slices. In the case of POEM, however, the majority of 2D slices contain only background, and even those slices that contain foreground never contain all of the classes. As a lack of a class in an image results in a zero distance map (by implementation design), that particular class can be arbitrarily segmented without an increase in the loss.
                To circumvent this issue, we use a simple map of ones for every absent class, assuring some minimal amount of penalty. In addition, considering that the POEM data is highly anisotropic and its coronal slices may contain unconnected regions belonging to the same class, we run a separate set of experiments with distance maps calculated in 3D volumes, using all slice-wise point annotations. In 3D, since all classes are present in every volume, the problem of zero distance maps and no supervision is avoided entirely.

                \begin{table}[h!]
                \centering
                \caption{Average time it takes to compute the signed distance maps for all foreground classes on 2D slices of size $256 \times 256$ and entire 3D volumes of the ACDC and POEM datasets, given in seconds. The computation times are based on the implementations of \cite{asad2022fastgeodis} for $D_\text{euc}$, $D_\text{geo}$, and $D_\text{int}$; and our own implementation 
                for $D_\text{mbd}$.}
                \begin{tabular}{lcccc}
                        \toprule
                        & \multicolumn{2}{c}{ACDC} & \multicolumn{2}{c}{POEM} \\
                        Distance & in 2D & in 3D & in 2D & in 3D\\
                        \midrule
                        \midrule
                        $D_\text{euc}$  & $0.021$ & $0.091 $ & $0.007$ & $1.814$\\
                        $D_\text{geo}$  & $0.023$ & $0.092 $ & $0.006$ & $2.265$\\
                        $D_\text{int}$  & $0.019$ & $0.084$ & $0.006$ & $2.127$ \\
                        $D_\text{mbd}$  & $0.125$ & $7.203$ & $0.016 $& $98.673$ \\
                        \bottomrule
                \end{tabular}
                \label{tab:timetocompute}
                \end{table}

                \subsection{Baselines and settings}
                             
                               We compare to a fully supervised cross-entropy $\mathcal L_\text{CE}$ as an upper bound.
                        As a lower bound, we train a partial cross-entropy $\mathcal L_{\widetilde{\text{CE}}}$ with just the point annotations.

                        In addition, we compare the results to a state-of-the-art method for weak supervision, using a combination of the partial cross entropy and a CRF-loss from \cite{tang2018regularized}. We choose this work in particular because a number of successful works in various weakly supervised segmentation tasks either build upon it (e.g. \cite{tang_2018_normalizedcut, qu_2019_nuclei}) or use CRFs for postprocessing (e.g. \cite{ji_2019_scribble, lin_scribblesup_2016}). 

                        For both datasets, we use the lightweight E-Net by \cite{enet} and train the model with Adam optimizer, on a combination of partial cross entropy and boundary loss. The starting learning rate was set to $0.0005$ and halved every time the validation performances did not improve within 20 epochs. The training was carried out for 200 epochs on ACDC and 300 epochs on POEM data, using a batch size of 8. In training with a combination of the partial cross entropy and boundary loss, both losses are calculated only on the foreground classes and their contributions are weighted equally ($\alpha=1$). For the competing method of \cite{tang2018regularized}, the chosen parameters are $w=2e-9$, $\sigma_\text{rgb}=15$, $\sigma_\text{xy}=100$, and the scale factor is set to $0.5$.
                        
                        During training, we monitor the batch Dice score on a fully annotated validation set. the model that performs best according to this measure is used for subsequent evaluations on the test dataset. Each experiment is run 3 times for increased repeatability, and evaluation results are averaged over runs. 

                        All the experiments are run on an NVIDIA GeForce GTX 3080 Ti using cuDNN v11.8. The code is available at \url{https://github.com/EvaBr/geodesic_bl}.

                \subsection{Evaluation metrics}
                        We evaluate the performance of the methods through two standard segmentation metrics; Dice score (DSC) and 95th percentile Hausdorff distance (HD95). While training is performed on 2D slices, the evaluation metrics are reported on the entire 3D scans.
                        \paragraph{DSC}
                        The Dice similarity score measures the overlap between the ground truth volume $G$ and the output segmentation volume $S$, and is defined as $\frac{2|G\cap S|}{|G|+|S|}$, where $|\cdot|$ denotes the cardinality (in this case the nonzero element count). We report the Dice scores for every foreground organ/tissue, and over all classes. 
                        \paragraph{HD95} The Hausdorff distance is a dissimilarity measure, representing the distance between the surfaces of $G$ and $S$. As it is sensitive to outliers, we use the 95th percentile instead of the maximum for computing the directed distances. Again we report the metrics on each foreground class separately, as well as over all classes.

        \section{Results}
        \subsection{Segmentation of cardiac structures}
\begin{figure}
                        \centering
                        \includegraphics[height=8cm]{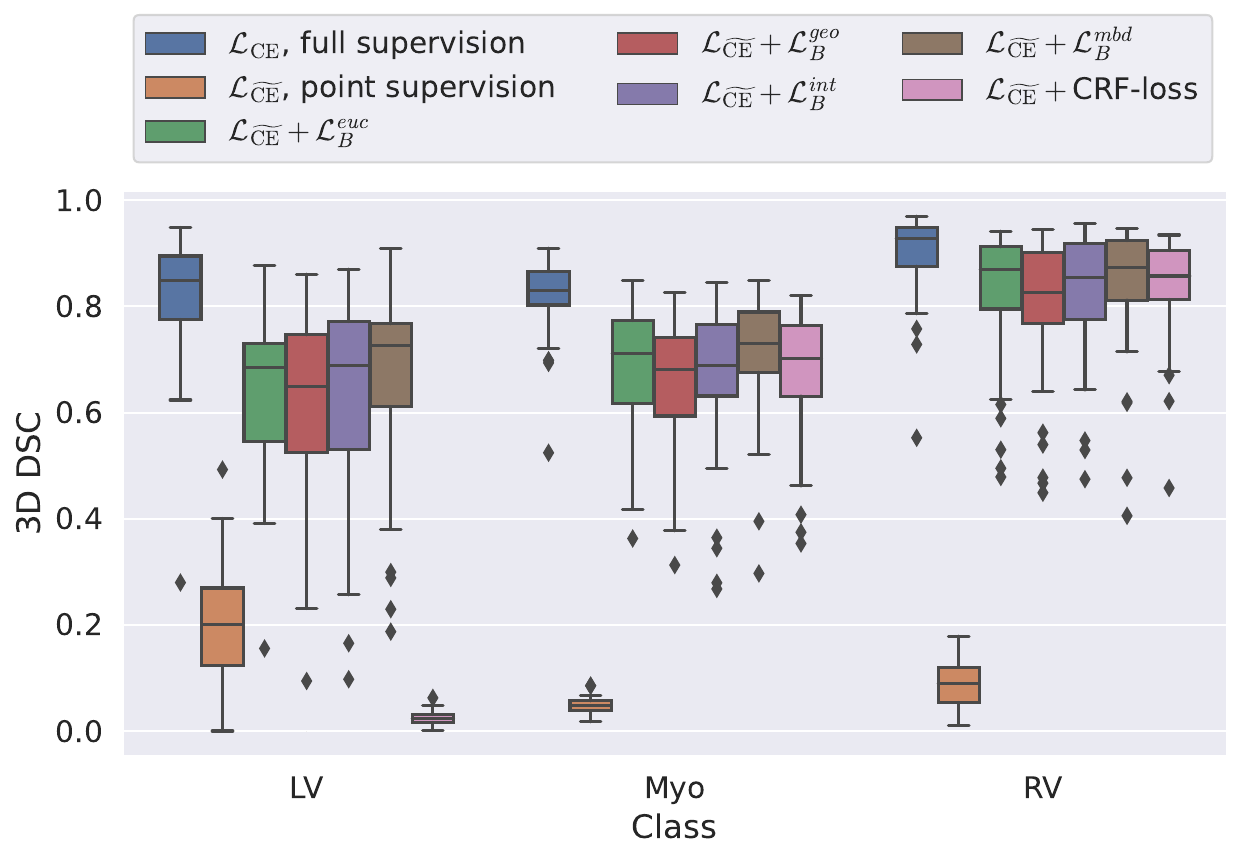}
                        \caption{Boxplots for the per-class 3D test Dice scores on the ACDC dataset. Results are shown for a single run. The labels LV, Myo and RV stand for left ventricle, myocardium and right ventricle respectively.}
                        \label{fig:boxplot_acdc_dsc}
                \end{figure}

\begin{figure}
                        \centering
                        \includegraphics[height=8cm]{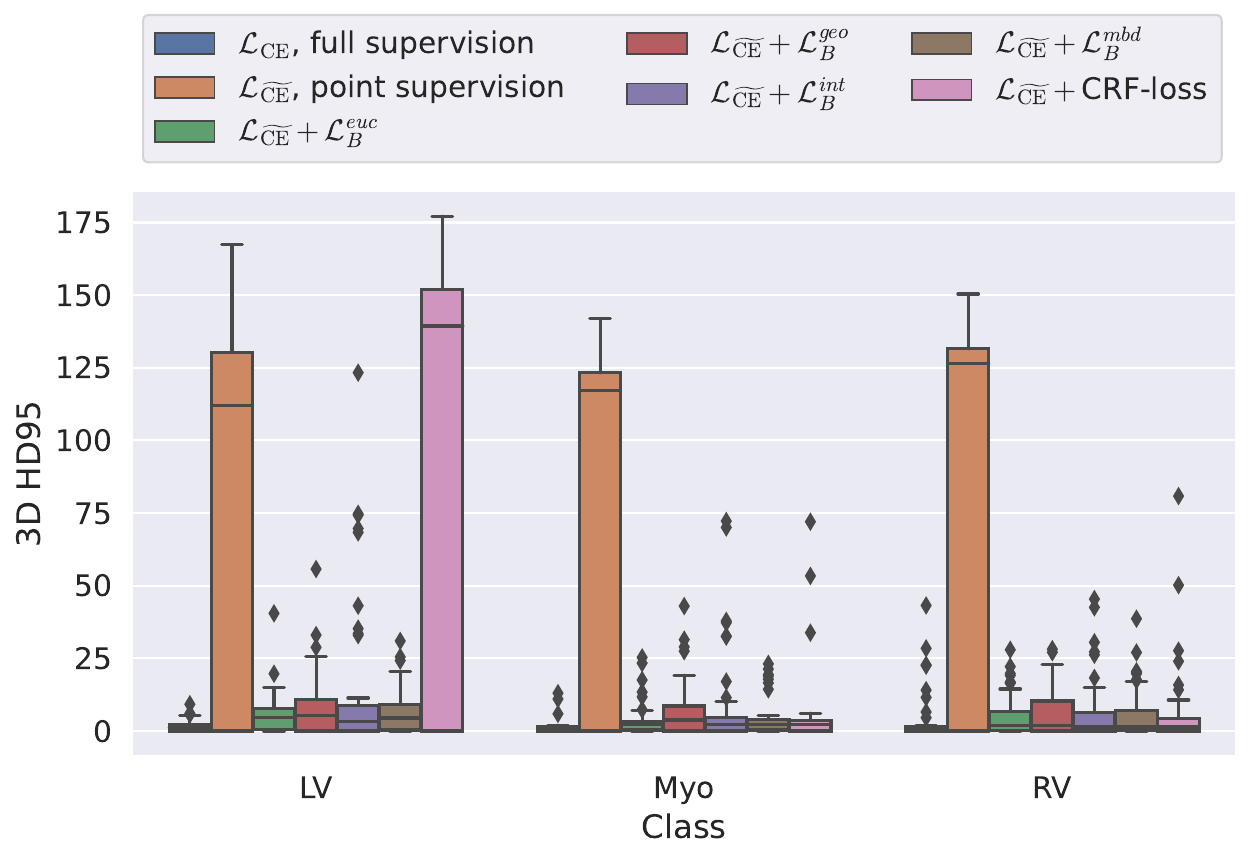}
                        \caption{Boxplots for the per-class 3D test HD95 values on the ACDC dataset. Results are shown for a single run. The labels LV, Myo and RV stand for left ventricle, myocardium and right ventricle respectively.} 
                        \label{fig:boxplot_acdc_hd95}
                \end{figure}

            The average 3D Dice scores and HD95 values on the ACDC test set are given in Table \ref{tab:acdcresults}, and boxplots in figures \ref{fig:boxplot_acdc_dsc} and \ref{fig:boxplot_acdc_hd95} show the distributions. 
            We see that, in terms of DSC, the proposed strategy of using intensity-aware distances withing boundary loss performs better than simply using the Euclidean distance, with the best results achieved by using the strictly intensity based MBD. The HD95 however favors the original version of $\mathcal{L}_B^{euc}$, which may be do to its smoother predictions and less fragmentation and oversegmentation. The CRF-loss results are significantly worse in both metrics.
            
            \input{acdccurves}
            
            \input{tableACDC}
            
            In Figure \ref{fig:acdc_curve} we show the 3D DSC validation curve evolution for a single run. The CRF-loss seems to have converged to a low DSC value, while all settings combining CE and boundary loss reach values close to the full supervision in the beginning of the training and then slowly collapse towards to the point annotations. The MBD version stands out, degrading slower, thus providing a wider range of potentially good models for evaluation.

            \paragraph{\bf Qualitative comparison}
                \input{examplesACDC}
            In Figure \ref{fig:ACDCout} we provide qualitative results on a number of randomly chosen test set slices. Upon visual inspection, we can observe that training with the intensity-aware distances (particularly with $\mathcal{L}_B^{int}$ and $\mathcal{L}_B^{mbd}$) follows the image gradients better and is better at recovering the underlying shape than the Euclidean version. The CRF-loss seems to recover the shape of the myocardium and left ventricle to some extent, but fails entirely on the right ventricle.  


        \subsection{Abdominal organ segmentation}

\begin{figure}
                        \centering
                        \includegraphics[height=8cm]{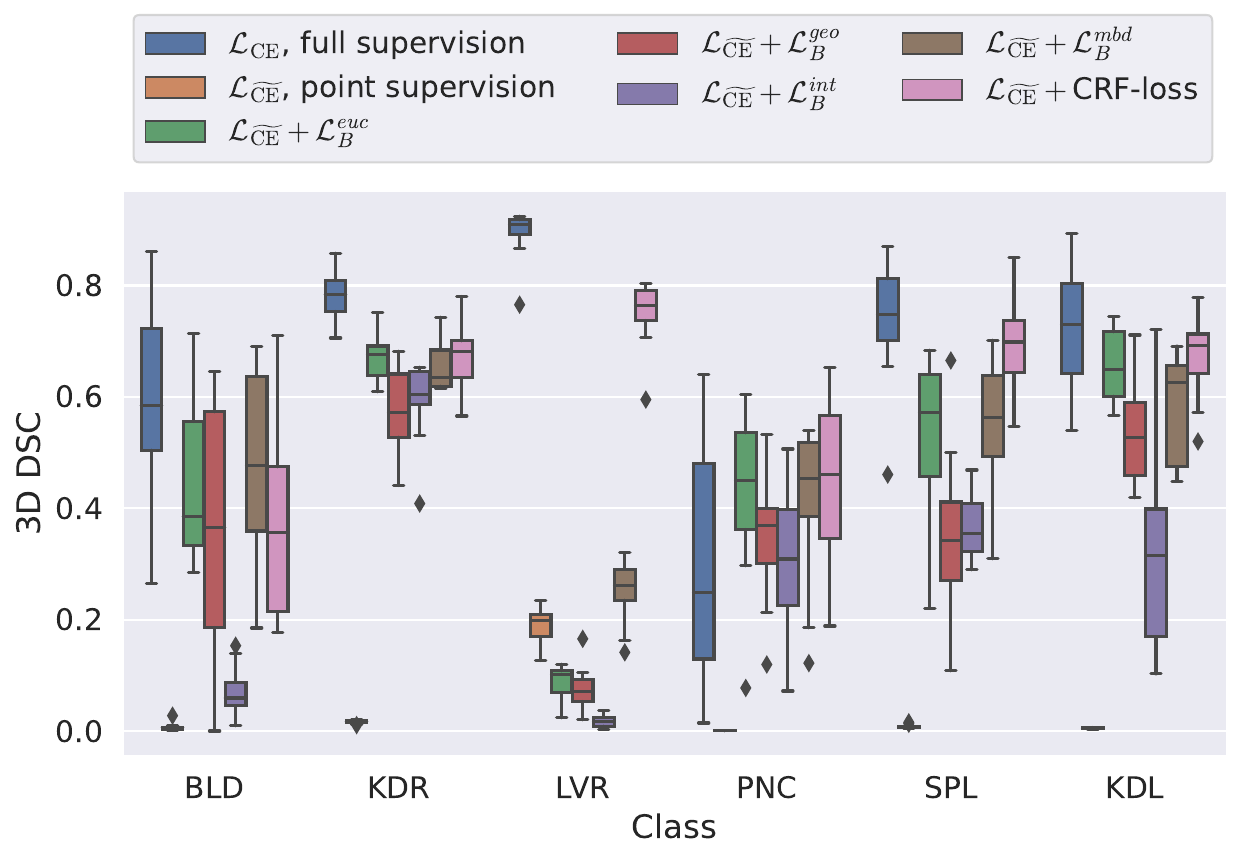}
                        \caption{Boxplots for the per-class 3D test Dice scores on the POEM dataset. Results are shown for a single run. The labels BLD, KDR, LVR, PNC, SPL and KDL stand for bladder, right kidney, liver, pancreas, spleen and left kidney respectively.}
                        \label{fig:boxplot_poem_dsc}
                \end{figure}

\begin{figure}
                        \centering
                        \includegraphics[height=8cm]{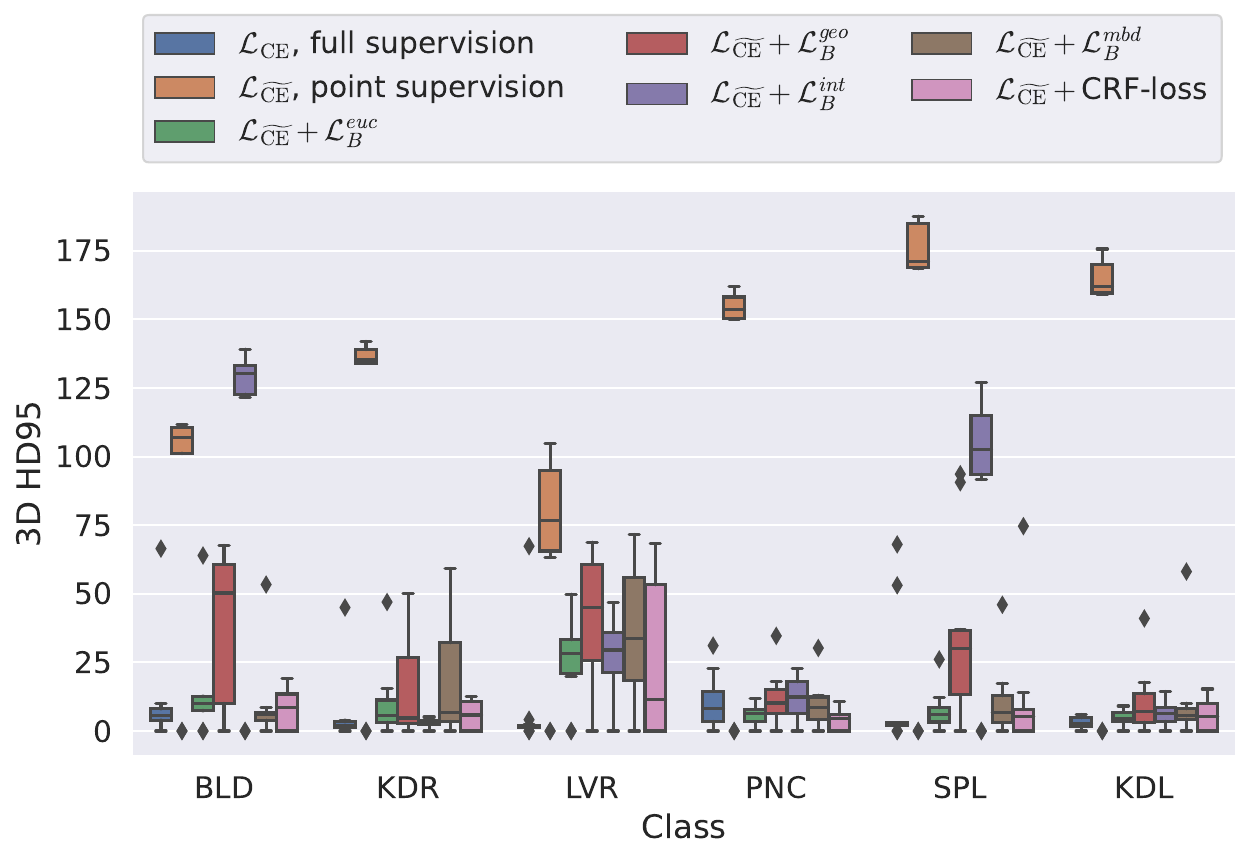}
                        \caption{Boxplots for the per-class 3D test HD95 values on the POEM dataset. Results are shown for a single run.} 
                        \label{fig:boxplot_poem_hd95}
                \end{figure}

                \paragraph{\bf Using 2D distance maps}
                In Table \ref{tab:poemresults}, the average DSC and HD95 results are shown (both using 2D and 3D distance maps) for the task of abdominal organ segmentation in POEM data (for boxplots see figures \ref{fig:boxplot_poem_dsc} and \ref{fig:boxplot_poem_hd95}). We see that training with $\mathcal{L}_B^{euc}$ and $\mathcal{L}_B^{mbd}$ (with distances calculated on 2D slices) performs comparably, while using $\mathcal{L}_B^{int}$ and $\mathcal{L}_B^{geo}$ produces lower scores in both DSC and HD95 metric.  On this dataset, the CRF-loss is able to compete with the boundary loss-based training strategies, even outperforming them on most classes. Most notably, all models trained with boundary loss appear to have a hard time segmenting the liver. We hypothesize this may be due to extremely severe class imbalance, as the liver covers a very large area compared to the rest of the classes. It is thus also more strongly affected by undersegmentations. 
                \input{tablePOEM}
                According to the validation curves in Figure \ref{fig:poem_curve}, training on this dataset is less stable, and slower, than on the ACDC one, for all methods. Using Euclidean or MBD maps appears to reach full-supervision scores, while the other methods lag behind. However, due to the long computation times on 3D data from the POEM cohort, these curves now show the evolution of the 2D Dice, which is less representative of the true success of the methods. 
                 \input{poemcurves}

            \paragraph{\bf Using 3D distance maps}
            Inspecting the values of training with $\mathcal L_\text{B} $ on  distance maps computed in 3D from Table \ref{tab:poemresults}, we notice that the results generally improve over values achieved by using 2D distance maps. Most notable decreases are visible in HD95 values, as using volume-calculated distance maps provides more global information and additionally penalizes spatially unreasonable segmentations. The methods based on $\mathcal L_\text{B} $ are now  able to compete with the CRF-loss, in particular the  $\mathcal L_\text{B}^{geo} $ one.
            \input{poemcurves3d}
            In Figure \ref{fig:poem_curve3d}, the validation curve evolution is plotted. Comparing it to the one with using 2D-computed distance maps, in Figure \ref{fig:poem_curve}, we see that the curves for all the methods training with $\mathcal L_\text{B}$ improve, with the exception of $\mathcal L_\text{B}^{mbd}$ based one. The lack of improvement here could be attributed to the MBD bleeding through object boundaries (due to noise) and propagating low distances further away in the volume, causing under-penalization. This is also suggested by the degradation in performance from 2D to 3D maps in Table \ref{tab:poemresults}. But at the same time, it allows for better segmentation of large and/or elongated (homogeneous) objects, which is also confirmed by the large improvement of liver segmentation scores in Table \ref{tab:poemresults}. 
            
            \subsubsection{Qualitative comparison}
            In Figure \ref{fig:POEMout} and \ref{fig:POEMout2} we show the same random slices in cases of calculating the boundary loss on 2D- and 3D-based distances, respectively. Comparing the two figures again indicates that the intensity-aware distances offer most improvement when calculated in 3D over 2D. The exception here is the MBD, which seems to even slightly degrade for most classes. 

            \input{examplesPOEM}
            \input{examplesPOEM2}

            \subsection{Computational complexity}
                While CRF-loss appears to perform very well on POEM data, it is also very slow to learn according to the validation curves in Figure \ref{fig:poem_curve}. In Table \ref{tab:traintimes} a quantitative comparison of the times per training iteration is given. We see that training by means of boundary loss with different distances incurs a small increase in iteration time, compared to training with only CE. However, the increase in time for the CRF-loss is much larger.
                
                \begin{table}[h]
                        \centering
                        \caption{Time per training iteration (s/iter), on each dataset, using a batch size of 8.}
                        \label{tab:traintimes}
                        \resizebox{0.35\columnwidth}{!}{%
                        \begin{tabular}{lcc}
                                \toprule
                                Method & ACDC & POEM \\
                                \midrule
                                $\mathcal L_\text{CE}$ (fully supervised) & $0.143$ & $0.196$ \\
                                \midrule
                                $\mathcal L_{\widetilde{\text{CE}}}$ (point annotations) & $0.143$ & $0.192$ \\
                                $\quad\quad \text{w/ } \mathcal L^{euc}_\text{B} $  & $0.144$ & $0.201$ \\
                                $\quad\quad \text{w/ } \mathcal L^{geo}_\text{B} $  & $0.144$ & $0.205$ \\
                                $\quad\quad \text{w/ } \mathcal L^{int}_\text{B} $  & $0.146$ & $0.200$ \\
                                $\quad\quad \text{w/ } \mathcal L^{mbd}_\text{B} $  & $0.146$ & $0.202$ \\
                                \midrule
                                $\quad\quad \text{w/ CRF-loss}$ \cite{tang2018regularized}  & $0.289$ & $0.304$ \\
                                \bottomrule
                        \end{tabular}%
                        }
                \end{table}

        \section{Discussion}
        We have proposed the use of boundary loss together with intensity-aware distances to train CNNs for segmentation tasks under very weak supervision.
        Our results demonstrate the usefulness of this simple approach, with performance close to full supervision.
        
        Compared to using CRF-loss, which typically requires heavy tuning and is highly sensitive to parameter settings, our method achieves the same or better results without additional tuning or increased training time. 
        Our approach generalized better across multi-class segmentation tasks, 
        while training with CRF-loss performed adequately only on the POEM data. 
        The reason behind the increased performance on the POEM dataset may be due to a larger number of classes, and thus inherently more supervision. On the ACDC dataset, annotations of the background would potentially be needed to steer the CRF-loss training in the right direction. 

        In addition to carrying better across different tasks/datasets, our approach is more easily understood and potentially tuned, as it is based on distance maps that are visually interpretable and have certain expected behaviours depending on the type of data and annotations. 
 
        Based on our results, the proposed use of boundary loss with intensity-aware distances generally performs better than the original formulation with Euclidean distance. However, using (purely) intensity based distances can result in more fragmented segmentations, while Euclidean distance based penalty produces smoother and more spatially contained ones. Therefore, the HD95 metric typically favours the latter. Perhaps surprisingly, the training with purely space-based, Euclidean distance loss, does not collapse towards the annotation points. We attribute this to the homogeneity of intensities within classes, and the network harnessing the underlying image intensity information through input. It also shows the importance of the distance map range. 

        We show also that for 3D datasets like POEM, with more complex class coocurrences, the distance maps calculated in 3D directly should be preferred. However, this can (depending on the chosen distance) incur large preprocessing computational costs. 

        Our experiments showed usefulness of intensity-aware distances within boundary loss based training under weak supervision. Not many available works in the literature can deal with that little supervision. And most that do, require  extra information that is typically extracted/based on the underlying image, or knowledge about the present objects, and not outside annotations. In our case, using intensity-aware distance maps computed from point annotations joins both, the information inherent to the image, and the outside annotation information.
        However, there are many small adaptations that can be explored for further improvements. For example, to avoid the potentially degrading effects of the boundary variation in computing the distance maps, the intensity average over the border (or entire annotation) could be taken as the intensity to compute the distance map from. This could potentially stabilize training and prevent bleed-out in MBD. 

        In addition, for MBD specifically, the effects of smoothing and  grayscale dilation prior to distance map computation could be investigated, as well as various curve-smoothing-based heuristics for early stopping (to remove the need for fully annotated validation datasets). 

        While we focused here on intensity-aware distances that account for the underlying intensities in a direct way, texture-type distances (e.g. \cite{shen_93_texturedist}), using more than single pixel intensity values, could potentially also be beneficial. Moreover, applying CRFs as a postprocessing step, or adaptively controlling  the spatial vs. intensity component contributions (within the distance definition) throughout training, remain to be explored.

        \section{Conclusion}
        In this paper, a novel approach of using intensity-aware distance with boundary loss for weakly supervised segmentation is presented. Despite its simplicity, it performs on-par or better than training with CRF-loss. We achieve results close to full supervision without additional tuning or an increase in training time. 
        Being directly interpretable and easily applied across datasets, our approach provides a promising alternative to the CRF-loss training and methods derived from it.

        \section*{Acknowledgements}
        E.B. is partially funded by the Centre for interdisciplinary mathematics (CIM), Uppsala University. H.K. and MdB are funded by the Dutch Research Council (NWO), VI.C.182.042.
        
\bibliographystyle{abbrv}
        \bibliography{main}
\end{document}

%% file: acdccurves.tex
\begin{figure}[h!]
                        \centering
                        \pgfplotslegendfromname{common}\\
                        \begin{tikzpicture}
                                \begin{axis}[
                                        name=acdcdsc,
                                        width=0.62\textwidth,
                                        height=0.35\textheight,
                                        at={(0,0)},
                                        ymin=0,
                                        ymax=1,
                                        xmin=0,
                                        xmax=199,
                                        ymajorgrids=true,
                                        ylabel={DSC},
                                        every axis y label/.style={
                                                at={(ticklabel cs:0.5)},rotate=90,anchor=near ticklabel,
                                        },
                                        every axis x label/.style={
                                                at={(ticklabel cs:0.5)},anchor=near ticklabel,
                                        },
                                        xlabel=Epoch,
                                        xtick={0,25,50,75,100,125,150,175,199},
                                        xticklabels={0,25,50,75,100,125,150,175,200},
                                        ytick={0,0.1,...,1},
                                        yticklabels={0.0,0.1,0.2,0.3,0.4,0.5,0.6,0.7,0.8,0.9,1.0},
                                        legend pos=outer north east,
                                        legend cell align=left,
                                        tick label style={font=\scriptsize},
                                        label style={font=\scriptsize},
                                        legend style={draw=none},
                                        legend style={line width=0.35mm},
                                        title style={font=\scriptsize},
                                        legend to name=common,
                                        legend columns=3,
                                        title={ACDC, validation DSC (3D)}
                                ]
                                        \addplot[color=Maroon, thick, dashed] table [x=epoch, y=val_Dice3D, col sep=comma, mark=none] {results/acdc/ce/metrics.csv};
                                        \addlegendentry{$\mathcal L_\text{CE}$, full supervision}
                                        
                                        \addplot[color=MidnightBlue, semithick] table [x=epoch, y=val_Dice3D, col sep=comma, mark=none] {results/acdc/euc/metrics.csv};
                                        \addlegendentry{$\mathcal L_{\widetilde{\text{CE}}} + \mathcal L^{euc}_\text{B}$}

                                        \addplot[color=SkyBlue, semithick] table [x=epoch, y=val_Dice3D, col sep=comma, mark=none] {results/acdc/geo/metrics.csv};
                                        \addlegendentry{$\mathcal L_{\widetilde{\text{CE}}} + \mathcal L^{geo}_\text{B}$}

                                        \addplot[color=Orange, thick, dashed] table [x=epoch, y=val_Dice3D, col sep=comma, mark=none] {results/acdc/ce_weak/metrics.csv};
                                        \addlegendentry{$\mathcal L_{\widetilde{\text{CE}}}$, point supervision}
                                        
                                        \addplot[color=OliveGreen, semithick] table [x=epoch, y=val_Dice3D, col sep=comma, mark=none] {results/acdc/int/metrics.csv};
                                        \addlegendentry{$\mathcal L_{\widetilde{\text{CE}}} + \mathcal L^{int}_\text{B}$}

                                        \addplot[color=SpringGreen, semithick] table [x=epoch, y=val_Dice3D, col sep=comma, mark=none] {results/acdc/mbd/metrics.csv};
                                        \addlegendentry{$\mathcal L_{\widetilde{\text{CE}}} + \mathcal L^{mbd}_\text{B}$}

                                        \addplot[color=Purple, semithick] table [x=epoch, y=val_Dice3D, col sep=comma, mark=none] {results/acdc/crf/metrics.csv};
                                        \addlegendentry{$\mathcal L_{\widetilde{\text{CE}}} + \text{CRF-loss}$}
                                        

                                \end{axis}
                        \end{tikzpicture}
                        \caption{Curve evolution of the average (over foreground classes) validation (3D) Dice scores during training, for the ACDC dataset in the multi-label segmentation training. } 
                        \label{fig:acdc_curve}
                \end{figure}

%% file: tableACDC.tex
\begin{table*}
                        \centering
                        \caption{Mean $\uparrow$DSC and $\downarrow$HD95 values 
                        over five independent runs, calculated on 3D volumes of the ACDC test set.  
                        Labels RV, Myo and LV represent the right ventricle, myocardium and left ventricle classes respectively.
                        Boxplots for one run showing the distributions over subjects are available in Figures  \ref{fig:boxplot_acdc_dsc} and \ref{fig:boxplot_acdc_hd95}.}
                        \label{tab:acdcresults}
                        \begin{tabular}{lccc|c}
                                \toprule
                                Method  
                                & \textsc{RV} & \textsc{Myo} & \textsc{LV} & \textsc{All} \\
                                \midrule
                                $\mathcal L_\text{CE}$ (fully supervised) & 
                                $\uparrow 0.7986\downarrow2.911$ & $\uparrow 0.8111\downarrow1.336$ & $\uparrow 0.8923\downarrow 2.774$ & $\uparrow 0.8748\downarrow1.754$ \\
                                \midrule
                                $\mathcal L_{\widetilde{\text{CE}}}$ (point annotations) & $\uparrow0.0991\downarrow91.645$ & $\uparrow0.0689 \downarrow82.347$ & $\uparrow0.2060 \downarrow 88.119$ & $\uparrow 0.2137\downarrow65.528$  \\
                                $\quad\quad \text{w/ } \mathcal L^{euc}_\text{B} $  & $\uparrow0.6214 \downarrow6.611$ & $\uparrow0.6709\downarrow3.976$ & $\uparrow 0.8112 \downarrow 5.69$ & $\uparrow0.7742\downarrow4.069$  \\
                                $\quad\quad \text{w/ } \mathcal L^{geo}_\text{B} $  & $\uparrow0.637\downarrow8.106$ & $\uparrow 0.679\downarrow5.186$ & $\uparrow0.82 \downarrow5.529$ & $\uparrow0.782\downarrow4.705$  \\
                                $\quad\quad \text{w/ } \mathcal L^{int}_\text{B} $  & $\uparrow0.639\downarrow9.991$ & $\uparrow0.6729\downarrow 5.435$ & $\uparrow 0.829 \downarrow4.282$ & $\uparrow0.7834\downarrow4.926$ \\
                                 $\quad\quad \text{w/ } \mathcal L^{mbd}_\text{B} $  & $\uparrow0.6553 \downarrow8.968 $ & $\uparrow0.6956\downarrow5.943$ & $\uparrow0.8297\downarrow6.179$ & $\uparrow0.7936\downarrow5.272$  \\
                                \midrule
                                $\quad\quad \text{w/ CRF-loss}$ \cite{tang2018regularized}  & $\uparrow0.2660\downarrow63.467$ & $\uparrow0.5385\downarrow27.492$ & $\uparrow0.8189\downarrow16.165$ & $\uparrow0.4558\downarrow26.781$  \\
                                 \bottomrule
                        \end{tabular}
                \end{table*}

%% file: examplesACDC.tex
\begin{figure*}
            \centering
        
      \begingroup
        \setlength{\tabcolsep}{1pt}
        
        \begin{tabular}{cccc} 
        \multicolumn{4}{c}{\includegraphics[width=0.05\textwidth, angle=270]{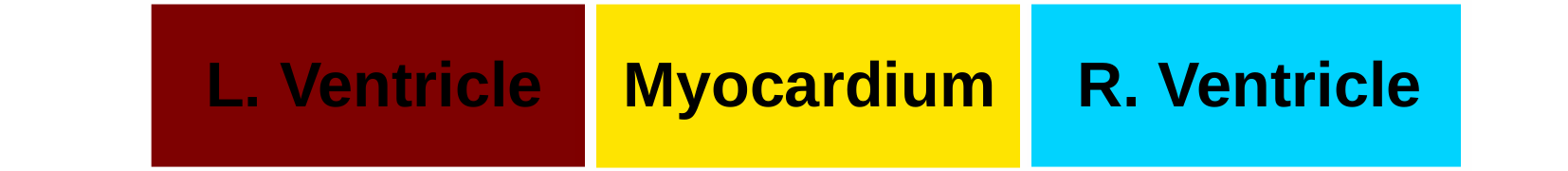}} \vspace{0.2cm}\\

                                \includegraphics[width=0.18\textwidth]{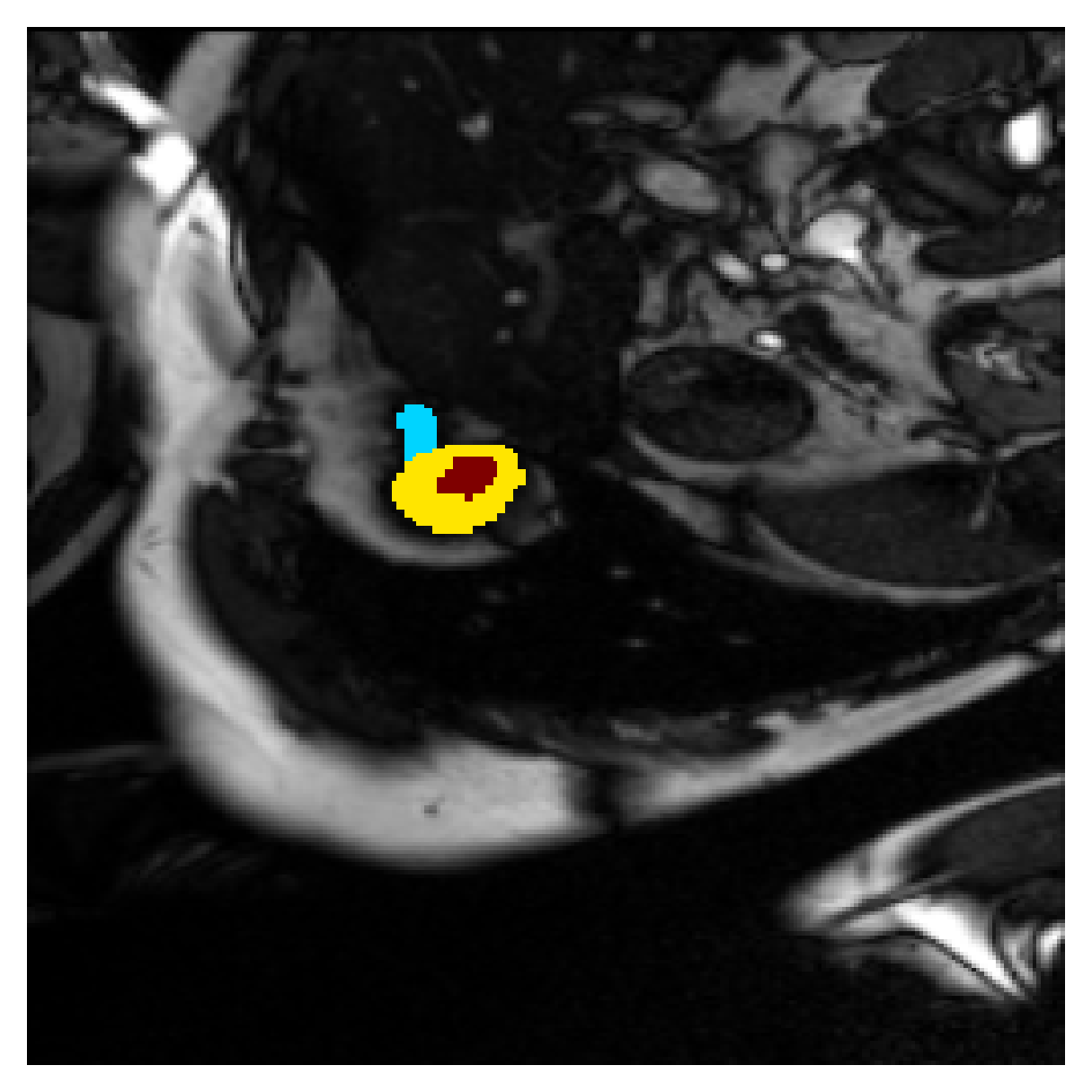} &
                                \includegraphics[width=0.18\textwidth]{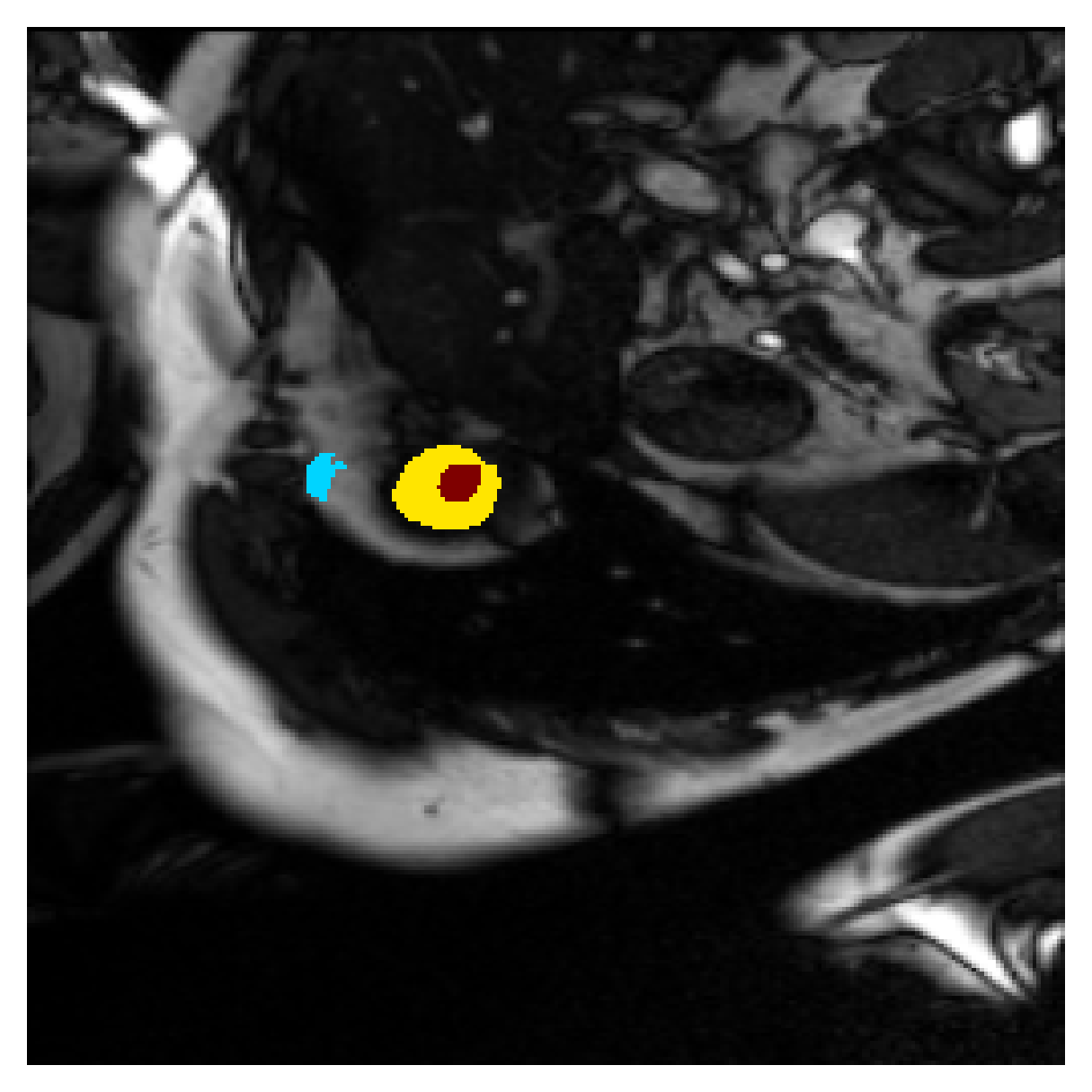} &
\includegraphics[width=0.18\textwidth]{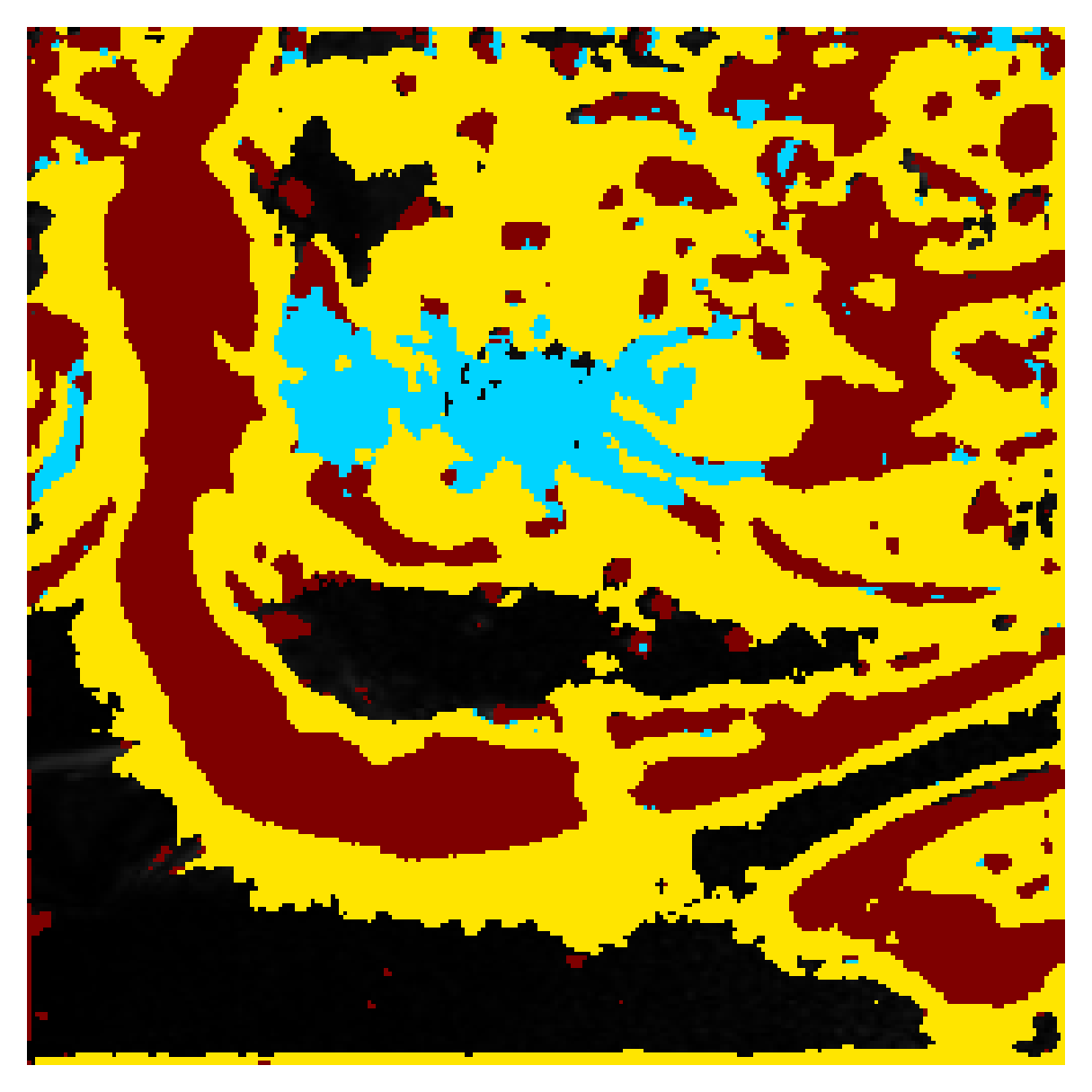} &
\includegraphics[width=0.18\textwidth]{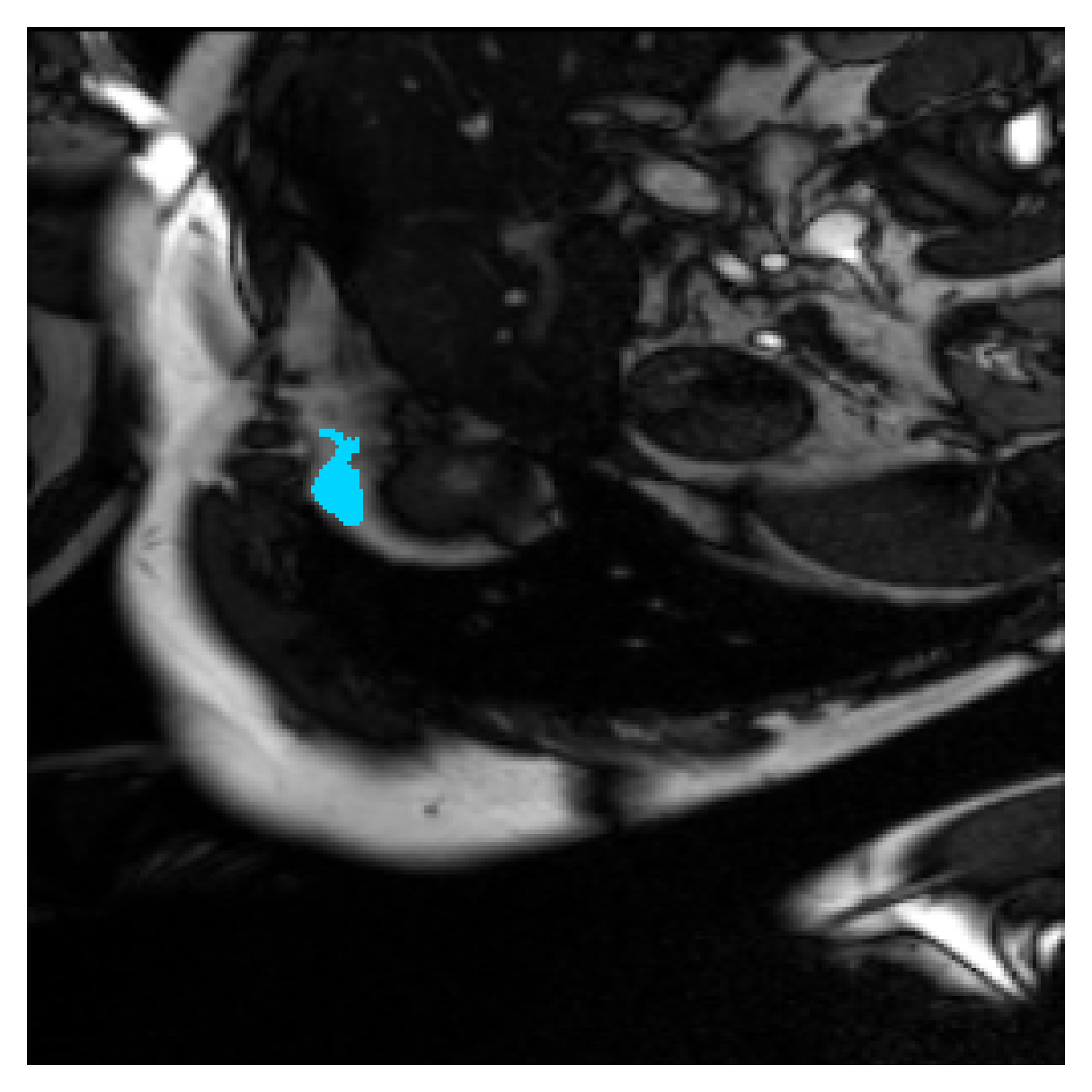} \\

                                \includegraphics[width=0.18\textwidth]{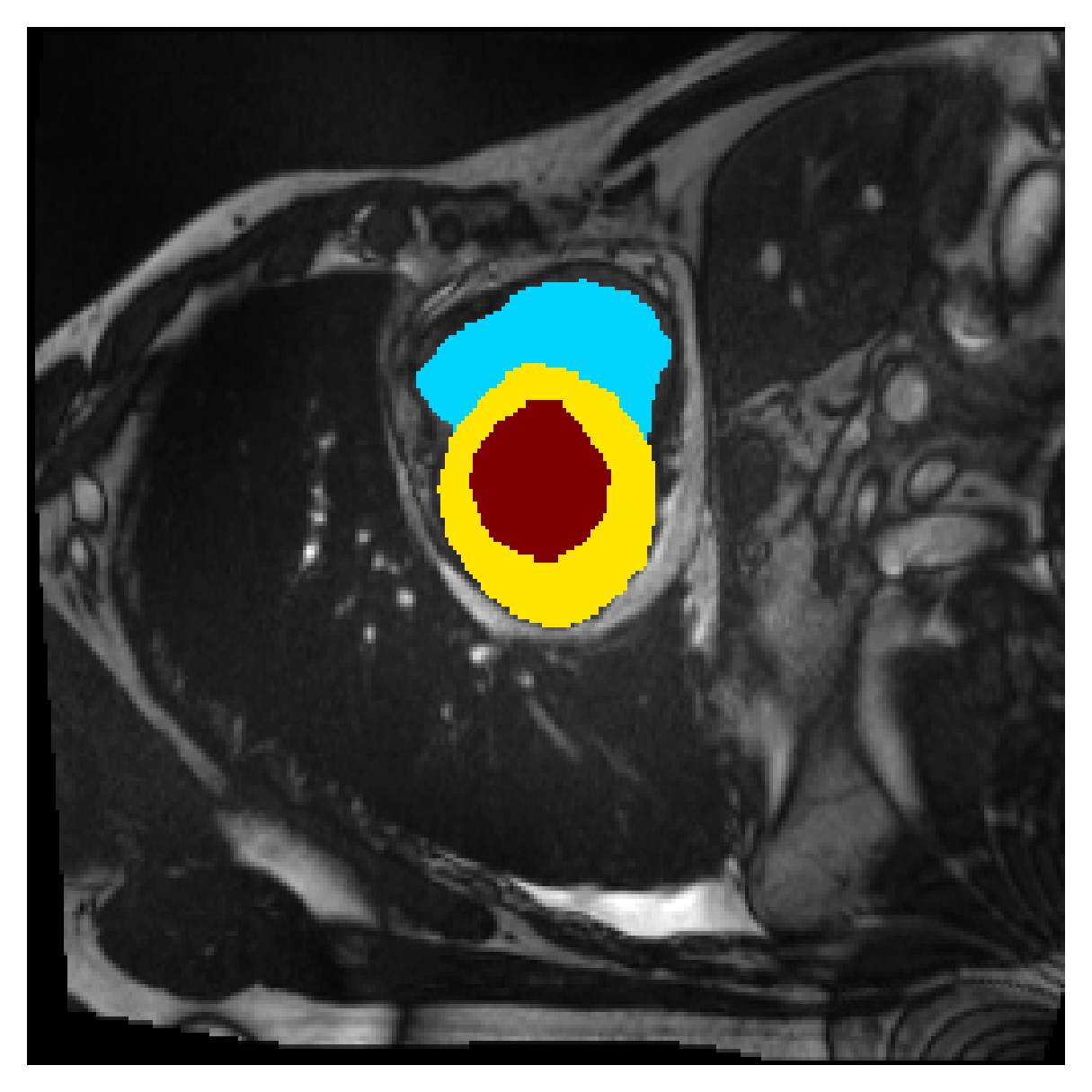} &
 \includegraphics[width=0.18\textwidth]{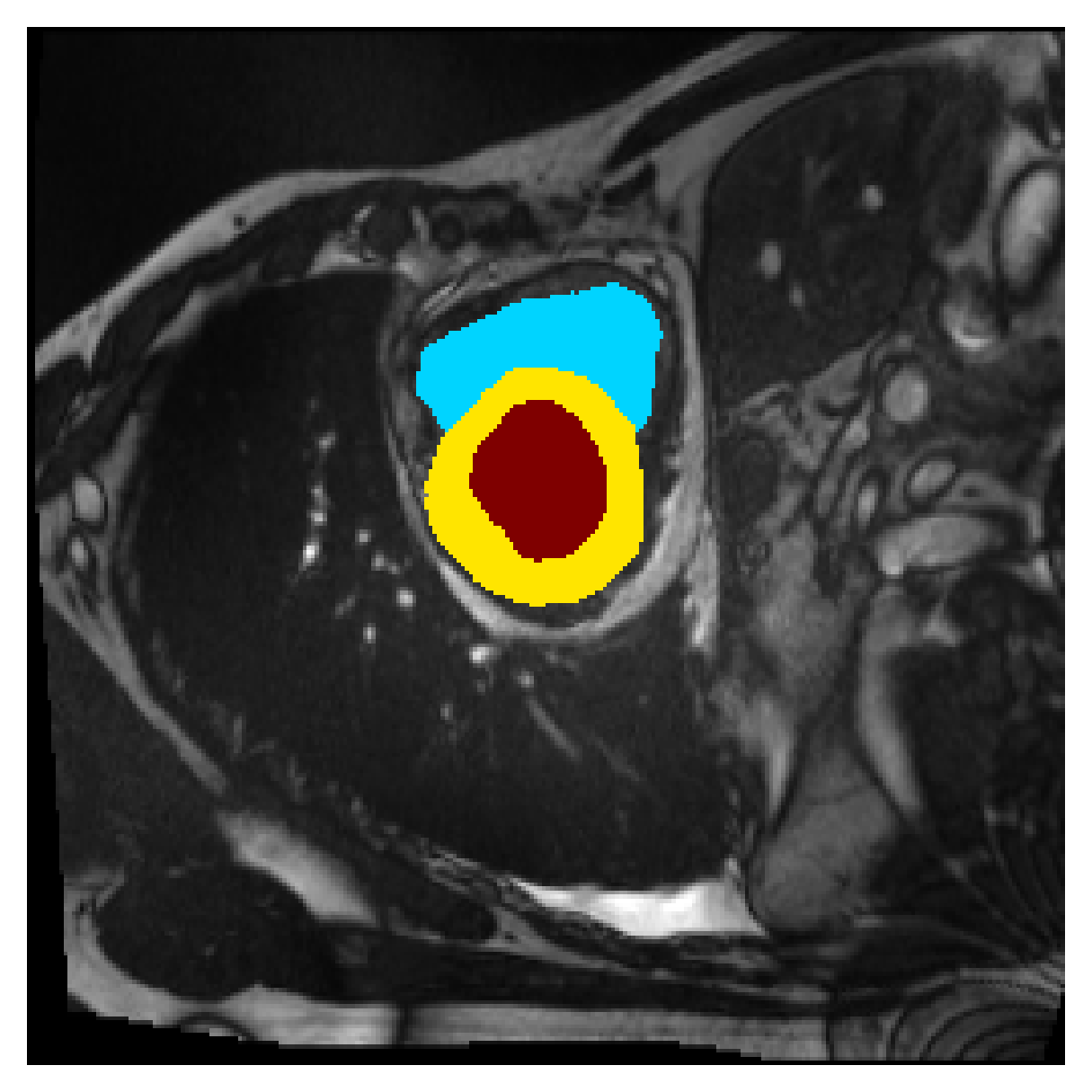} &
 \includegraphics[width=0.18\textwidth]{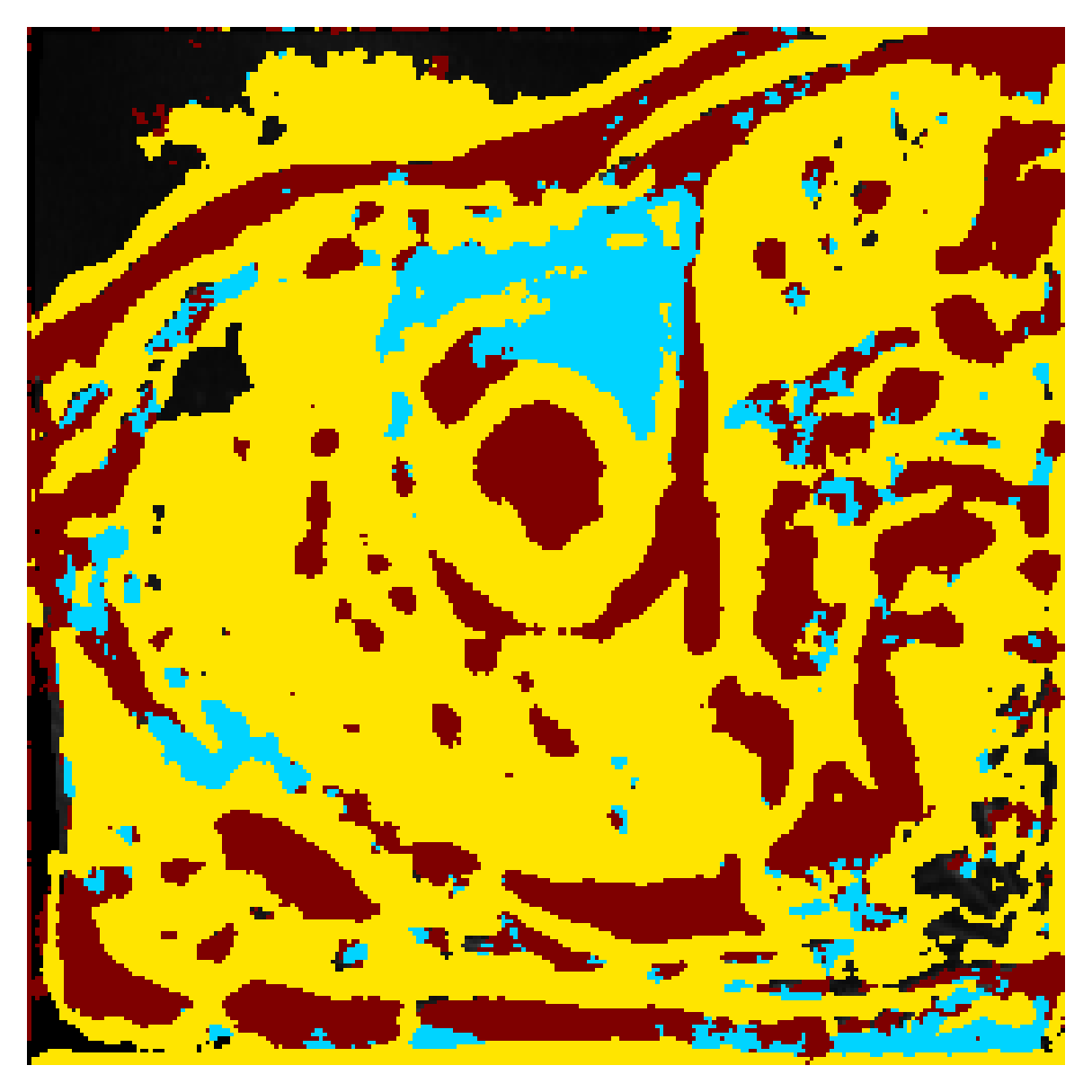}&
 \includegraphics[width=0.18\textwidth]{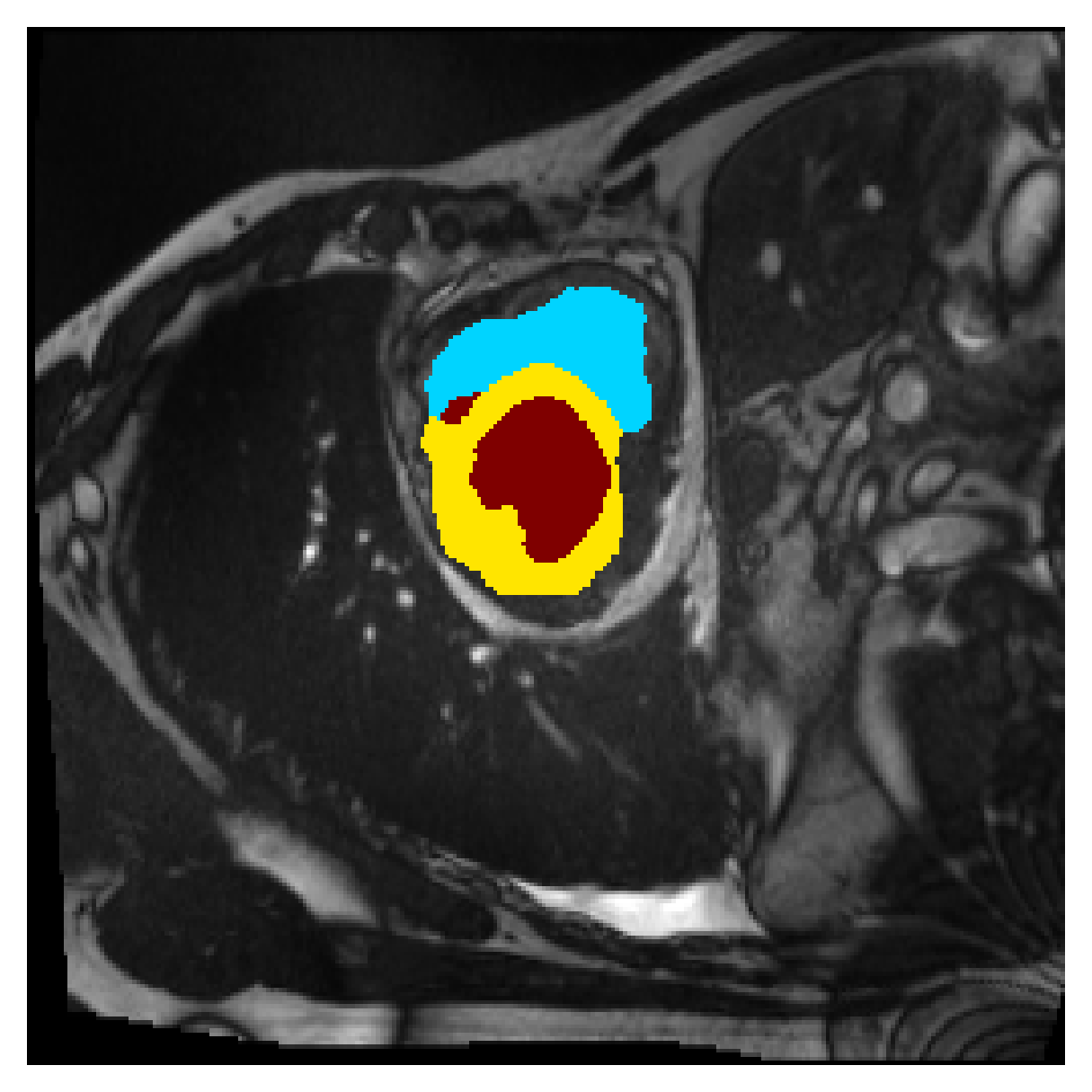}  \\

                                \includegraphics[width=0.18\textwidth]{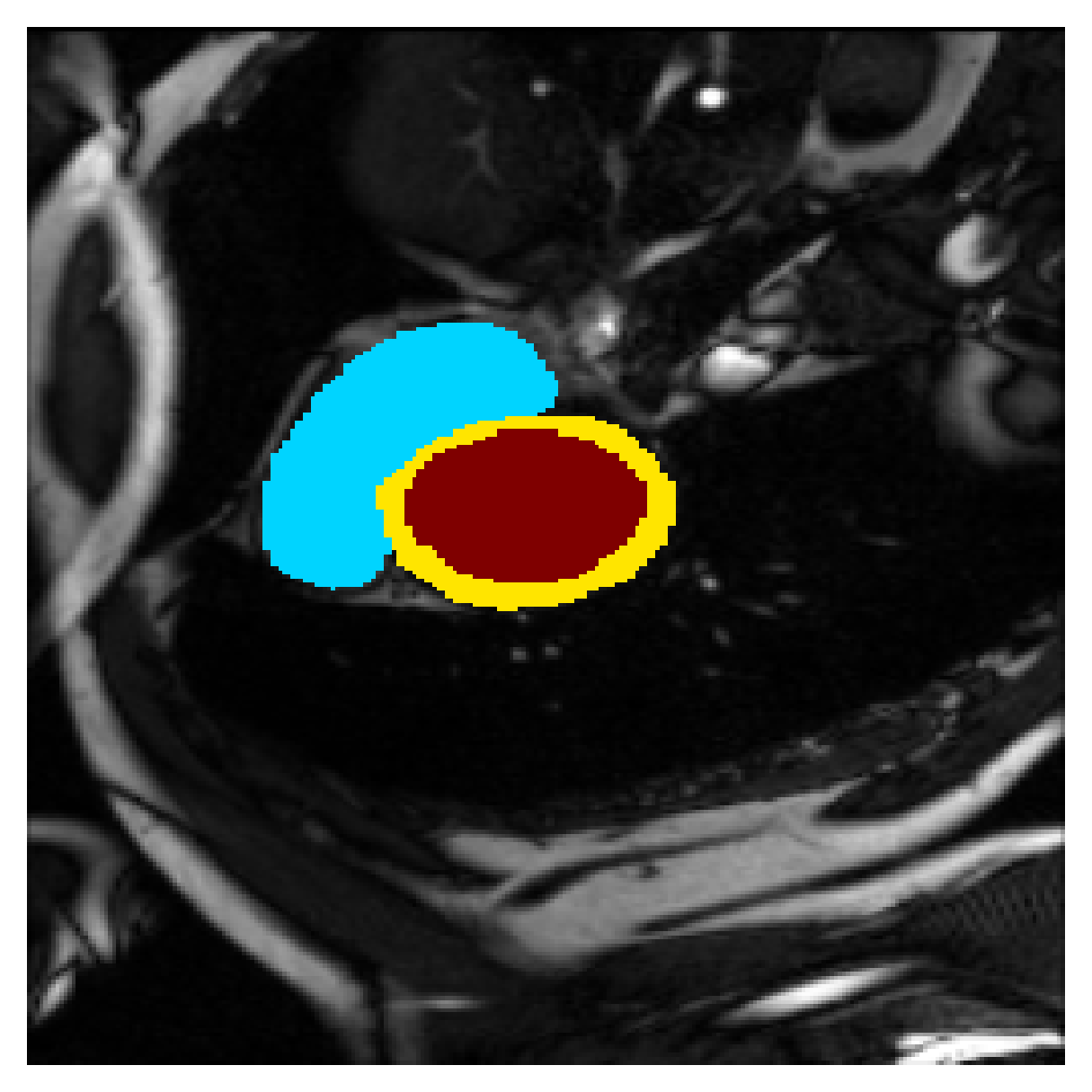} &
\includegraphics[width=0.18\textwidth]{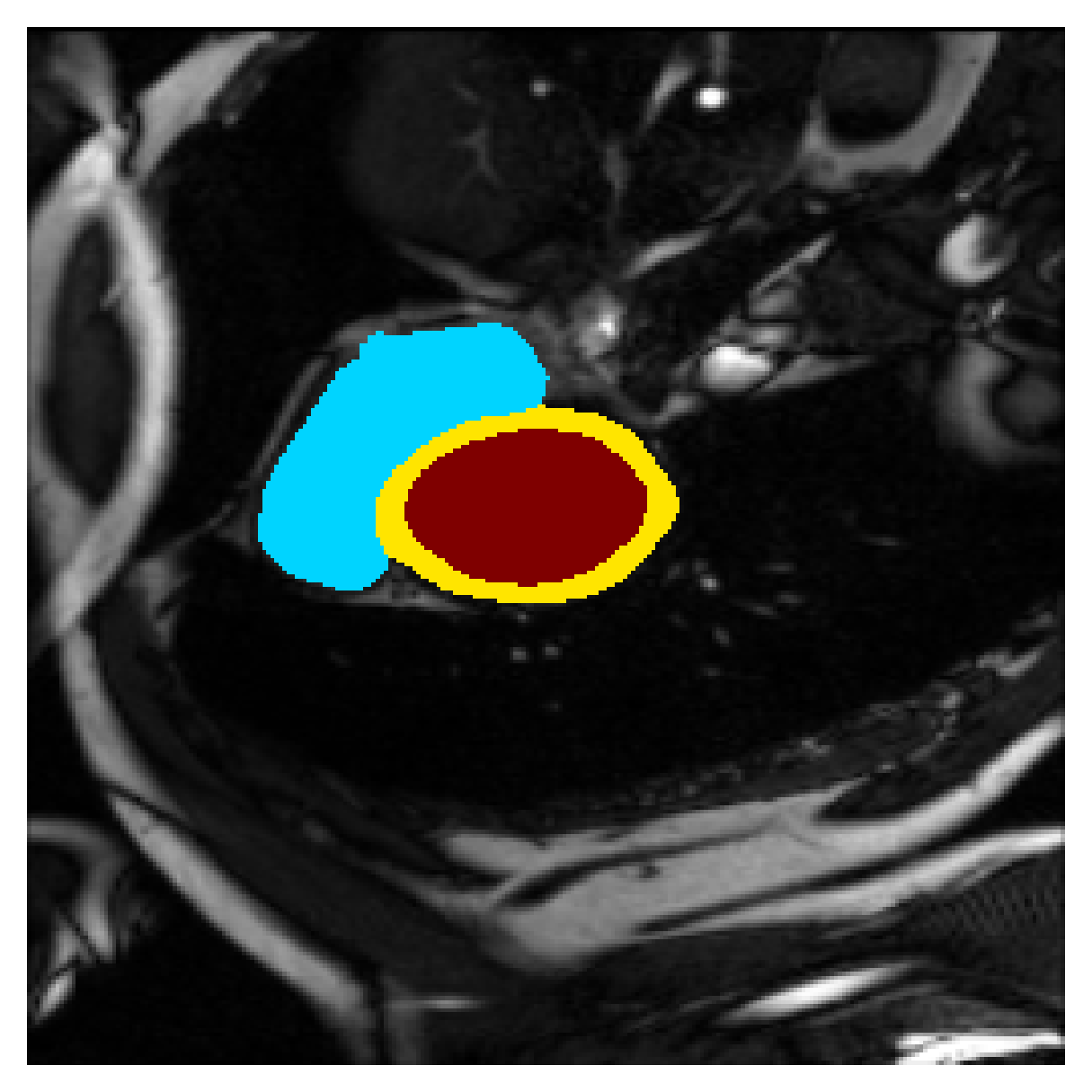} &
\includegraphics[width=0.18\textwidth]{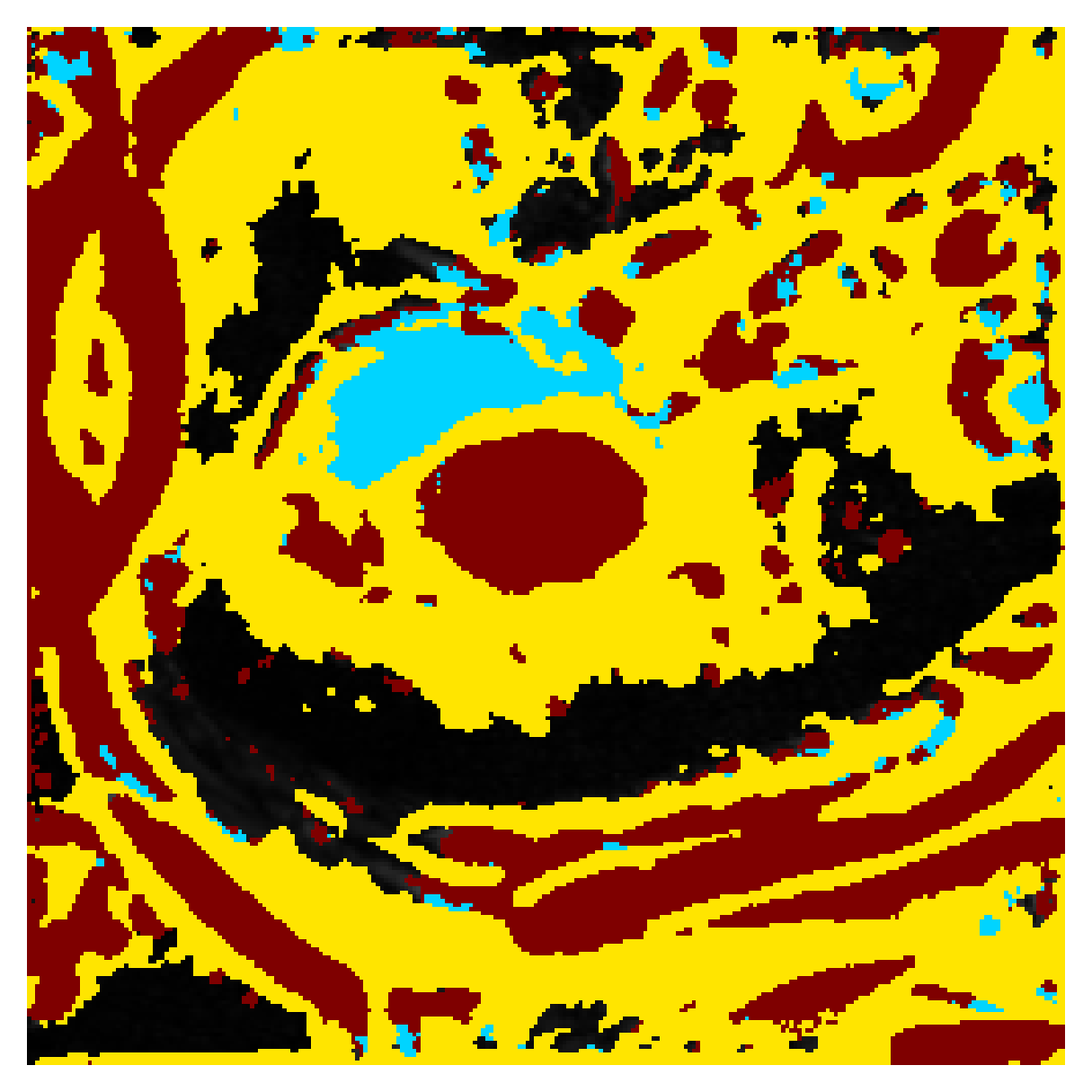} &
\includegraphics[width=0.18\textwidth]{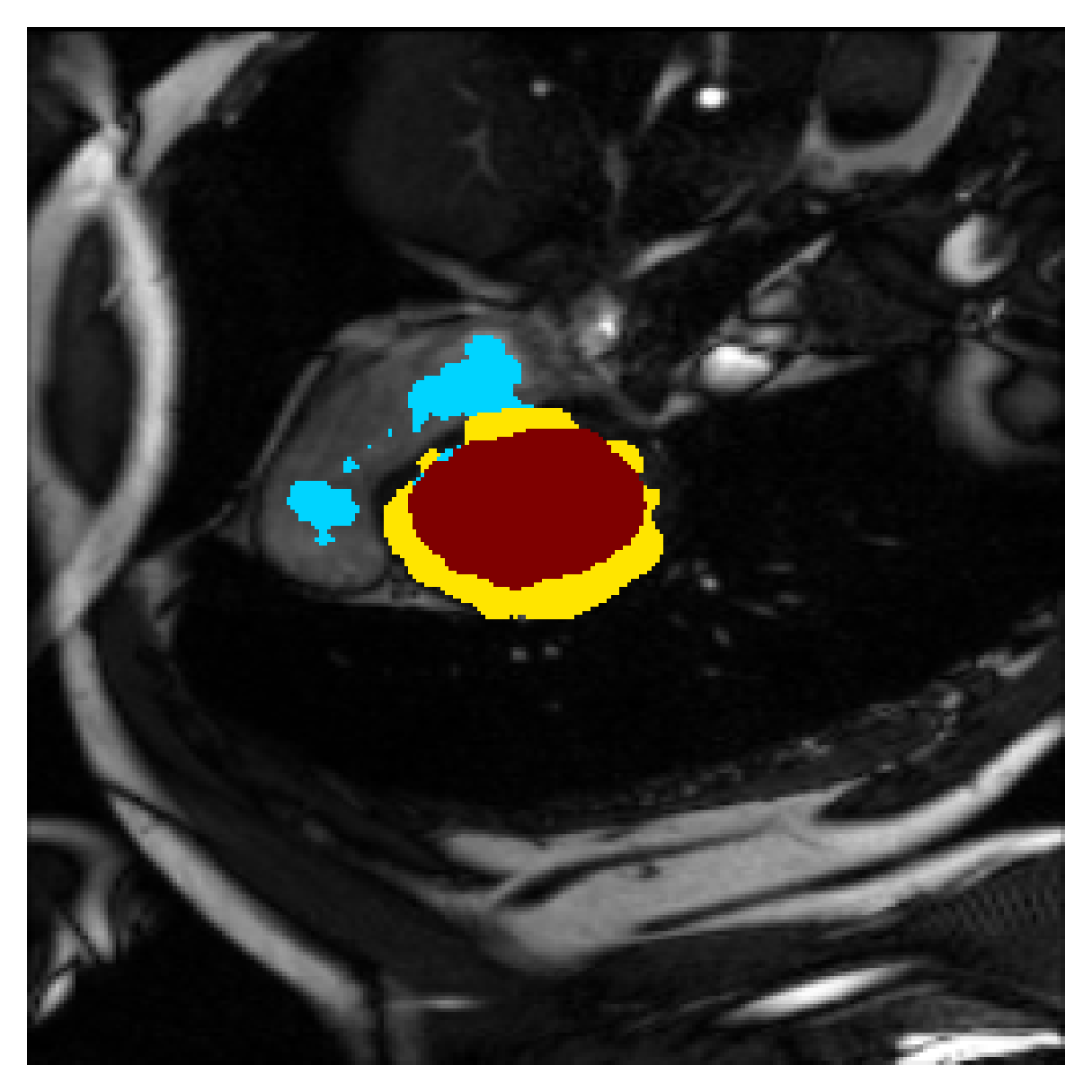} \\

(a) Ground truth  & (b) $\mathcal L_{\text{CE}}$, full  &
                                (c) $\mathcal L_{\widetilde{\text{CE}}}$, weak &
                                (d) $\mathcal L_{\widetilde{\text{CE}}} + \mathcal{L}_B^{euc}$ \\
                                (full) & supervision &supervision & 

\vspace{0.5cm} \\

                                \includegraphics[width=0.18\textwidth]{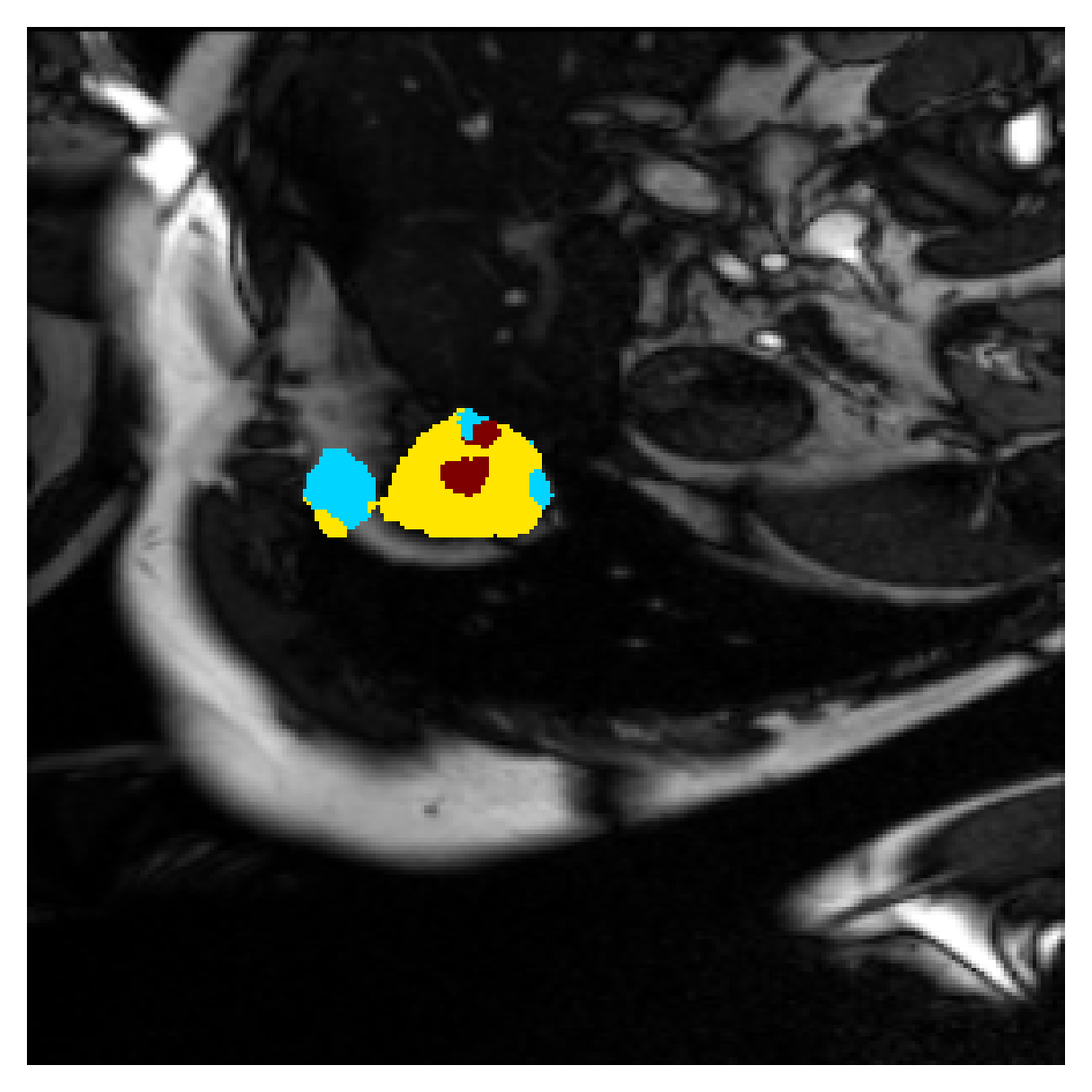} &
\includegraphics[width=0.18\textwidth]{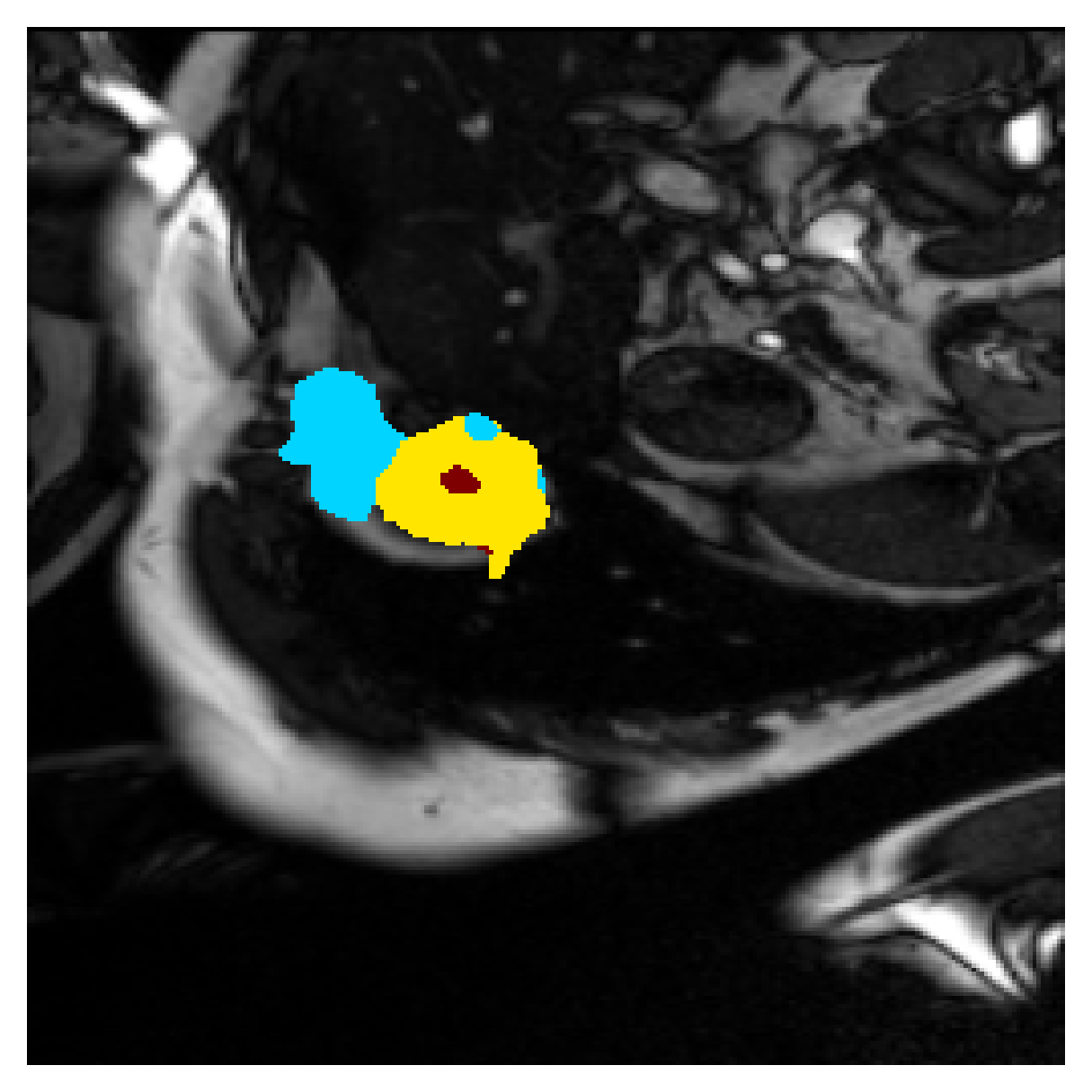} &
\includegraphics[width=0.18\textwidth]{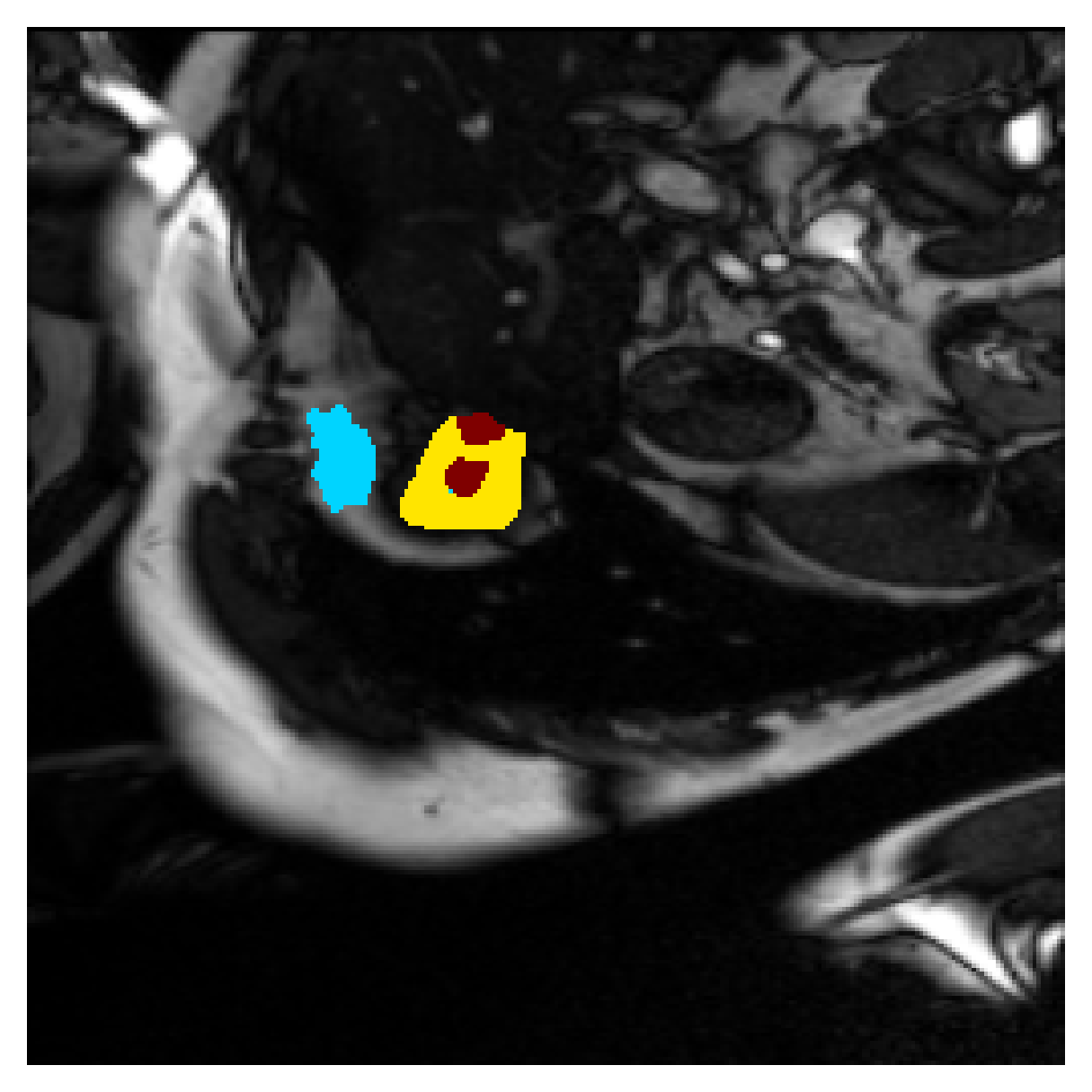} &
\includegraphics[width=0.18\textwidth]{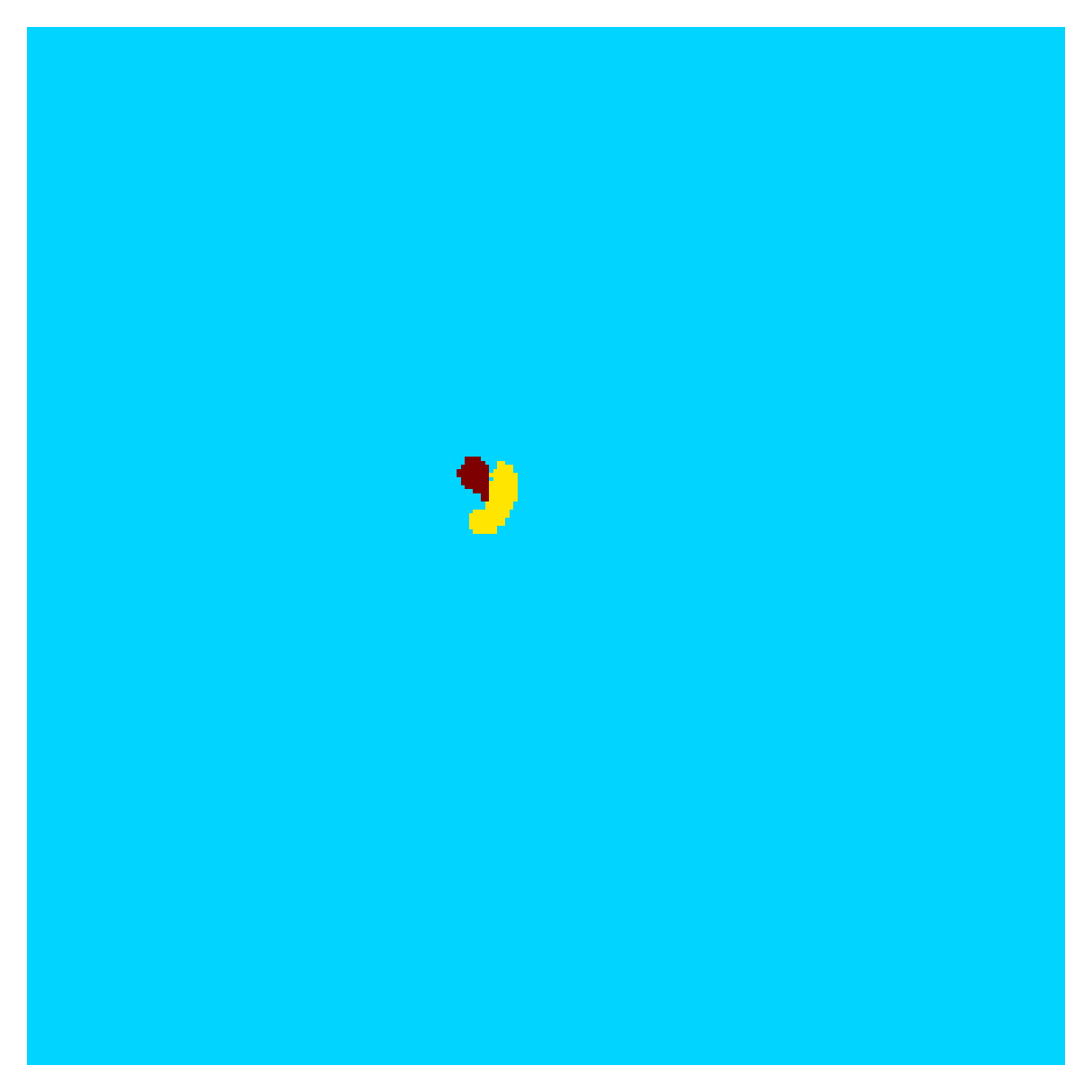} \\

                                \includegraphics[width=0.18\textwidth]{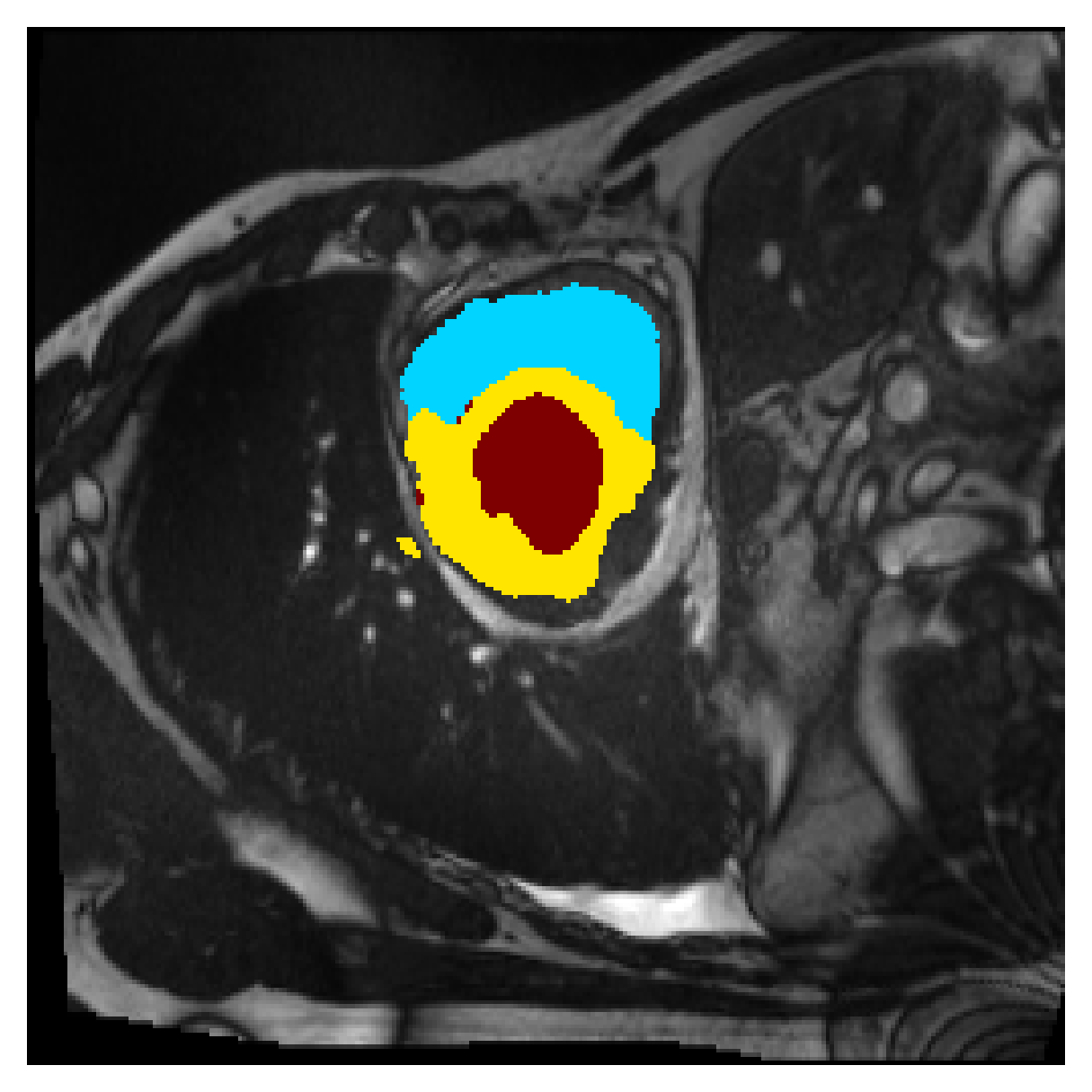}&
\includegraphics[width=0.18\textwidth]{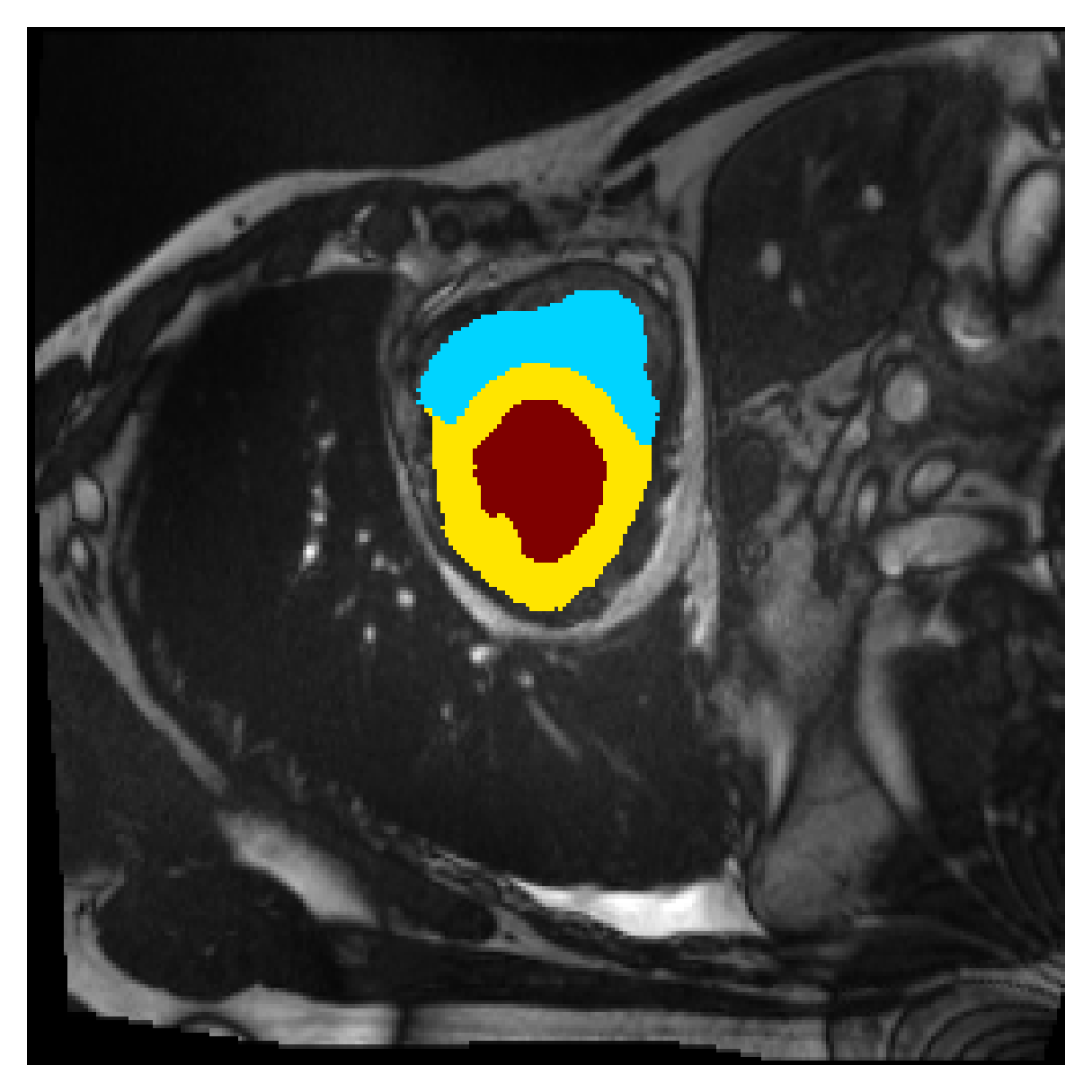} &
\includegraphics[width=0.18\textwidth]{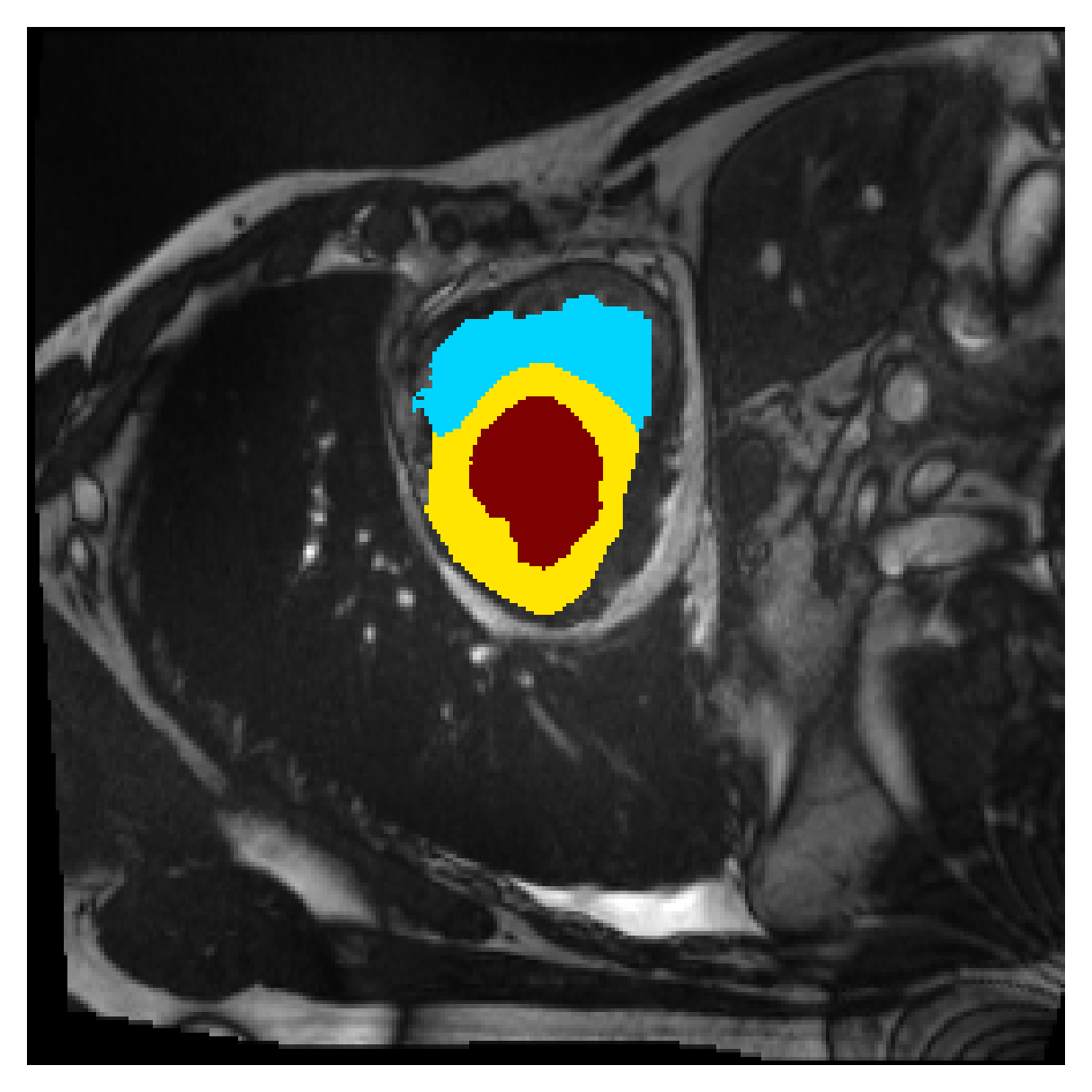} &
\includegraphics[width=0.18\textwidth]{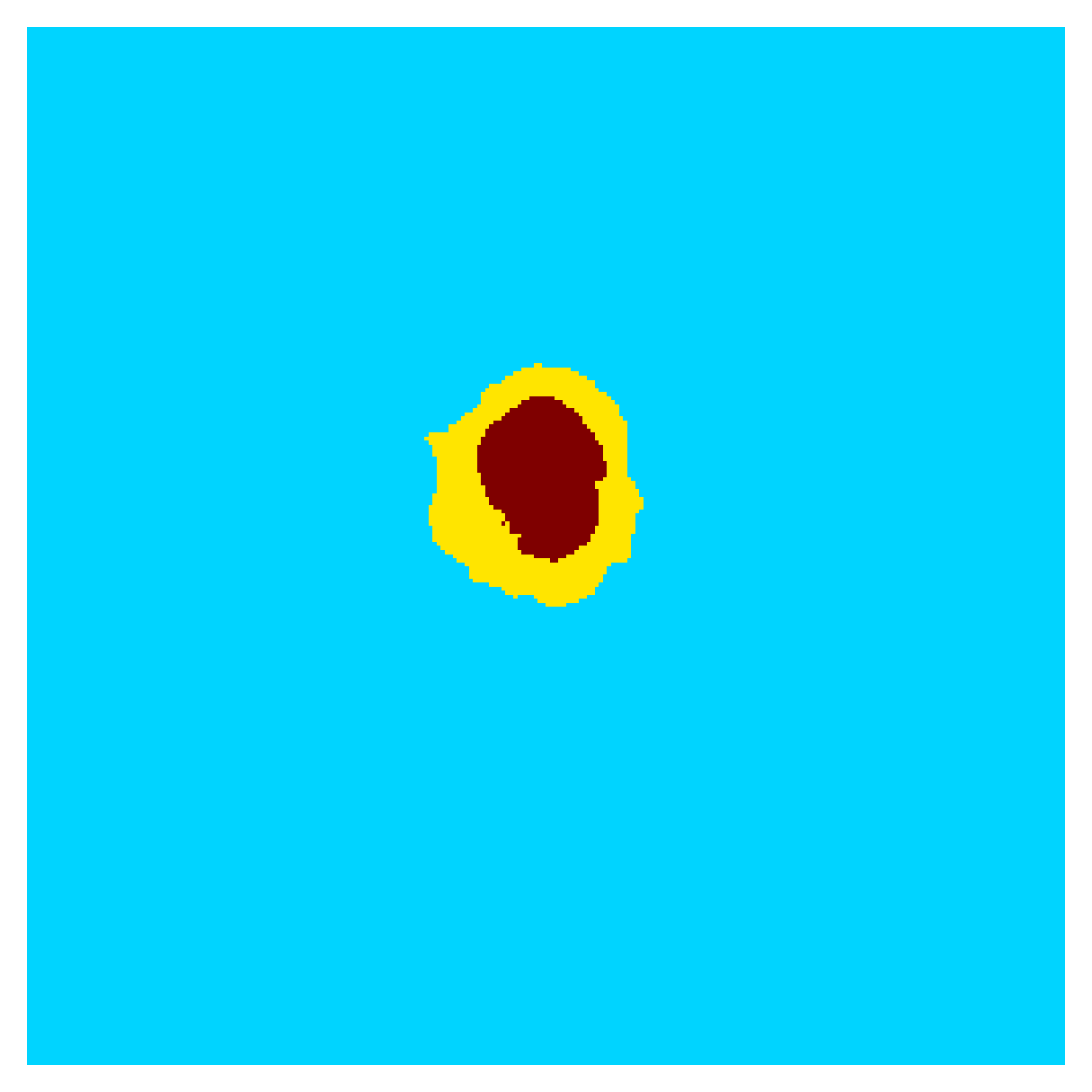} \\

                                \includegraphics[width=0.18\textwidth]{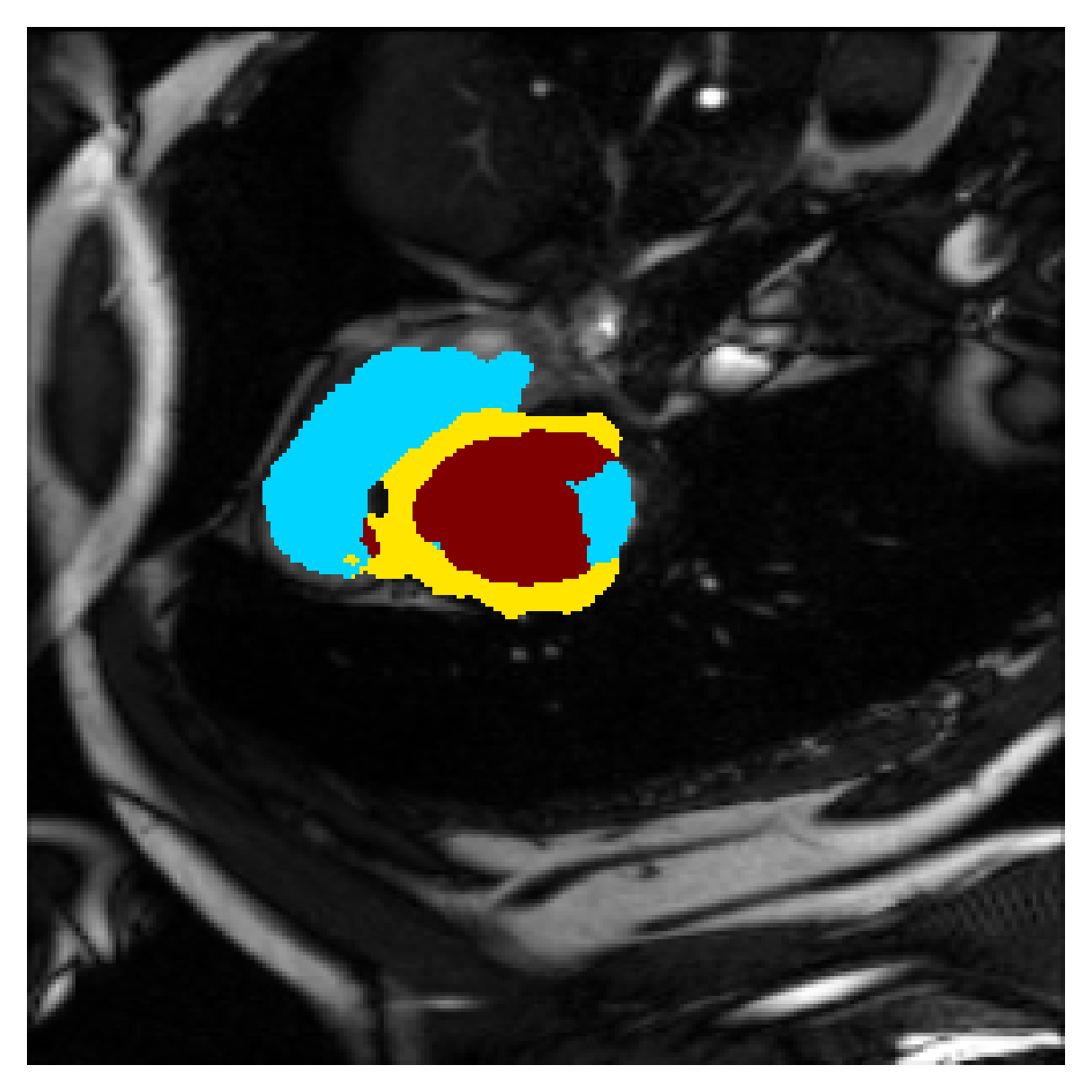} &
\includegraphics[width=0.18\textwidth]{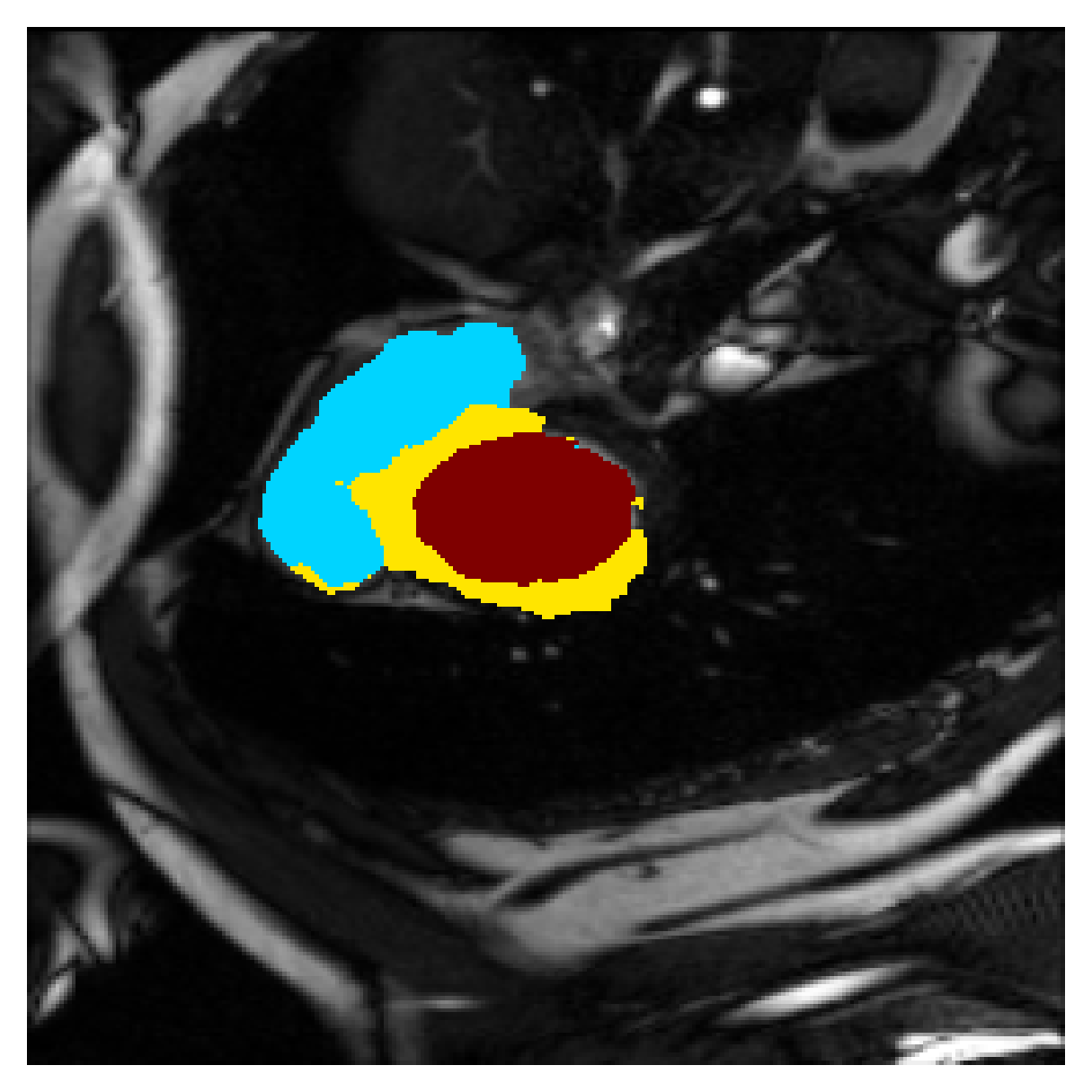} &
\includegraphics[width=0.18\textwidth]{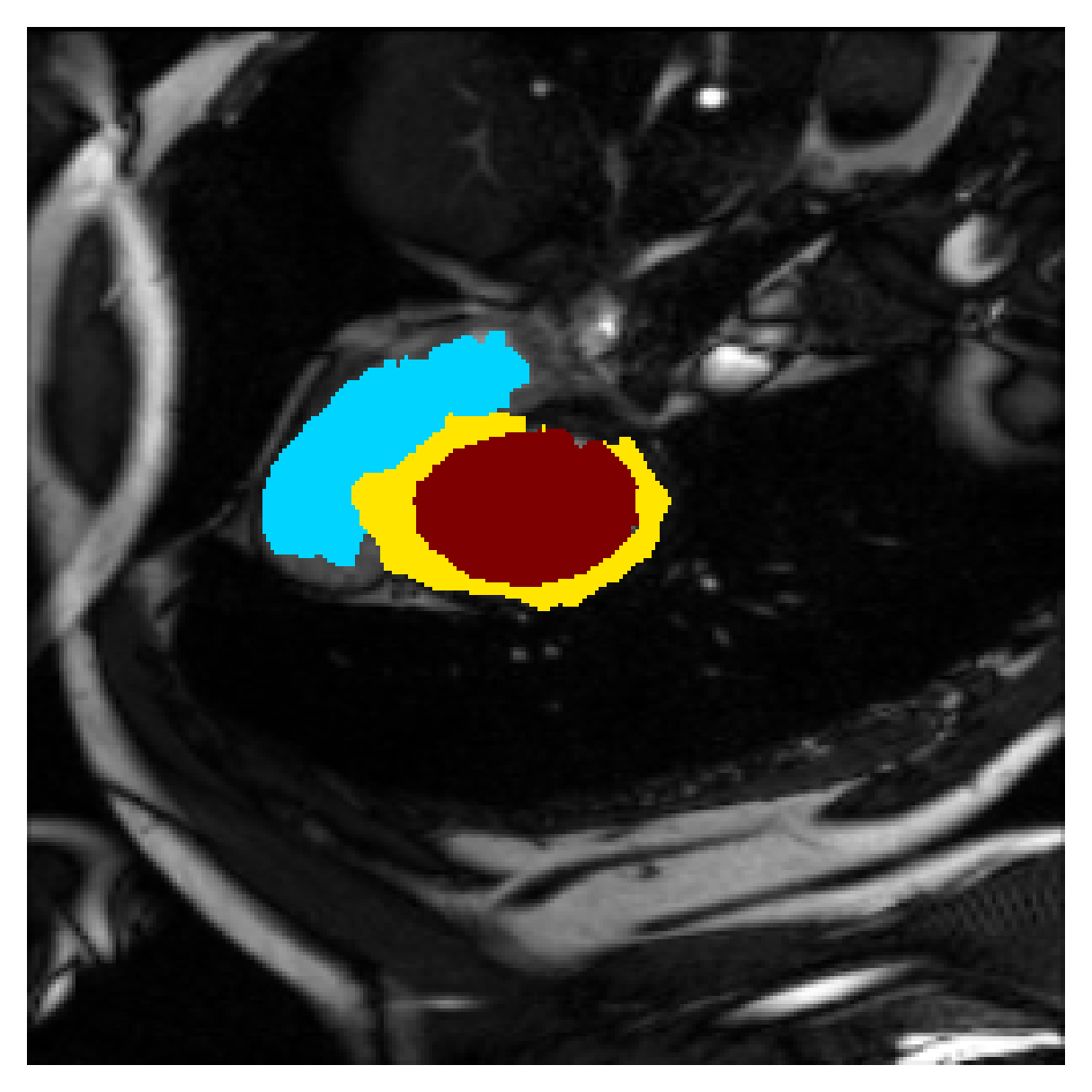} &
\includegraphics[width=0.18\textwidth]{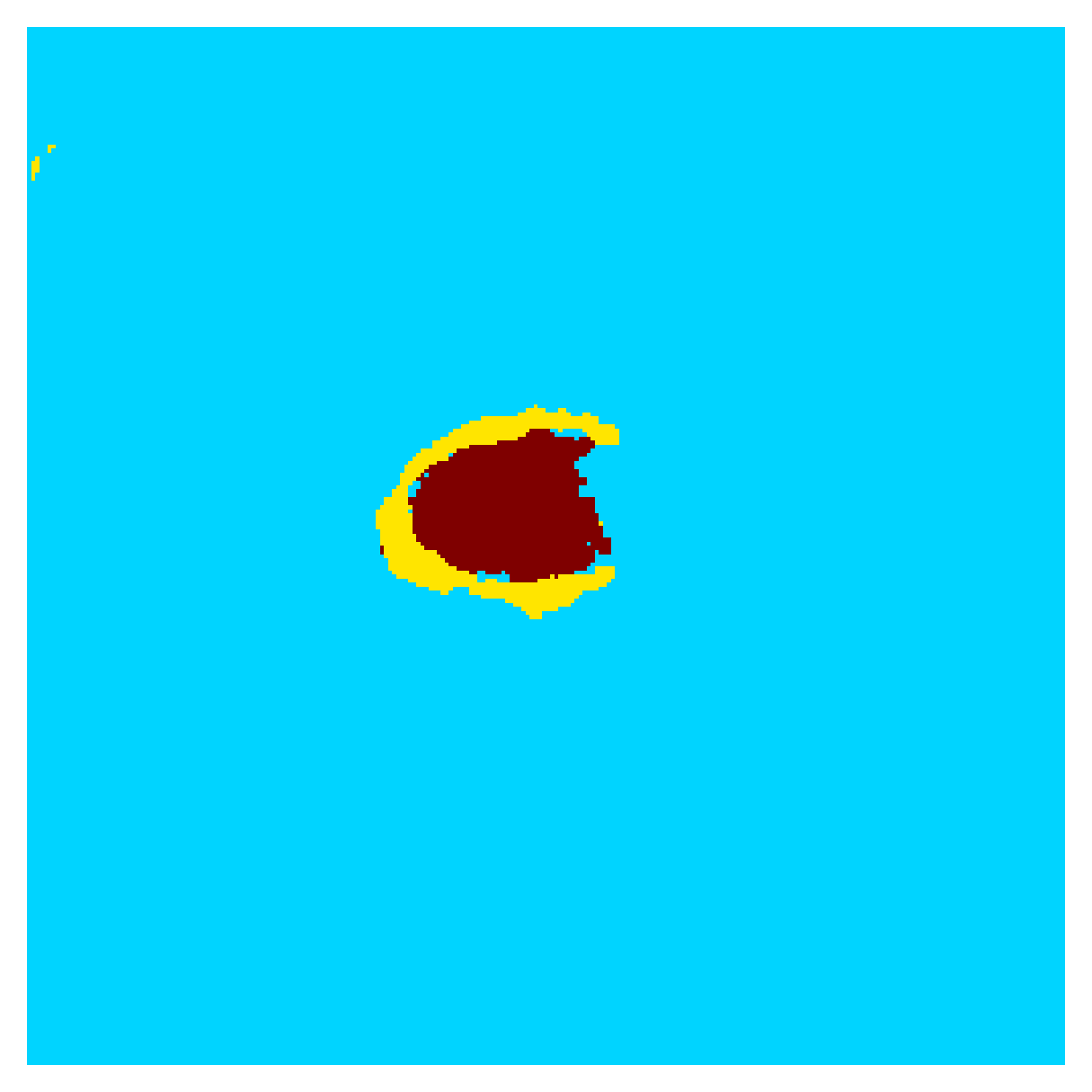} \\

 (e) $\mathcal L_{\widetilde{\text{CE}}} + \mathcal{L}_B^{geo}$ &
                                (f) $\mathcal L_{\widetilde{\text{CE}}} + \mathcal{L}_B^{int}$ & (g) $\mathcal L_{\widetilde{\text{CE}}} + \mathcal{L}_B^{mbd}$ & (h) $\mathcal L_{\widetilde{\text{CE}}}+$CRF-loss 
              \end{tabular}
              \endgroup
              
                        \caption{Example segmentations  on the ACDC test set. }
                        \label{fig:ACDCout}
                \end{figure*}

%% file: tablePOEM.tex
\begin{table*}
        \centering
        \caption{Mean $\uparrow$DSC and $\downarrow$HD95 values 
        over three independent runs, calculated on the 3D volumes of the POEM test set. Labels BLD, KDR, LVR, PNC, SPL and
        KDL and for bladder, right kidney, liver, pancreas, spleen and left kidney classes respectively. 
        Boxplots for one run showing the distributions over subjects are available in Figures  \ref{fig:boxplot_poem_dsc} and \ref{fig:boxplot_poem_hd95}.}
        \label{tab:poemresults}
        \resizebox{\linewidth}{!}{
        \begin{tabular}{lcccccc|c}
                \toprule
                Method  
                & \textsc{BLD} & \textsc{KDR} & \textsc{LVR} & \textsc{PNC} & \textsc{SPL} & \textsc{KDL}  & \textsc{All} \\
                \midrule
                $\mathcal L_\text{CE}$ (fully supervised) & $\uparrow0.607\downarrow7.161$ & $\uparrow0.734\downarrow4.261$ & $\uparrow0.895\downarrow4.992$ & $\uparrow0.327\downarrow10.504$ & $\uparrow0.588\downarrow10.138$ & $\uparrow0.656\downarrow3.630$ & $\uparrow0.687\downarrow5.812$  \\
                \midrule
                $\mathcal L_{\widetilde{\text{CE}}}$ (point annotations) & $\uparrow0.004\downarrow106.589$ & $\uparrow0.015\downarrow99.163$ & $\uparrow0.169\downarrow71.601$ & $\uparrow0.005\downarrow102.475$ & $\uparrow0.027\downarrow112.903$ & $\uparrow0.024\downarrow102.385$ & $\uparrow0.035\downarrow85.017$  \\
                
                $\quad\quad \text{w/ } \mathcal L^{euc}_\text{B} $ (in 2D) & $\uparrow0.482\downarrow9.576$ & $\uparrow0.689\downarrow6.952$ & $\uparrow0.087\downarrow25.049$ & $\uparrow0.436\downarrow6.832$ & $\uparrow0.530\downarrow7.673$ & $\uparrow0.664\downarrow6.538$ & $\uparrow0.555\downarrow8.946$   \\
                $\quad\quad \text{w/ } \mathcal L^{euc}_\text{B} $ (in3D) & $\uparrow0.471\downarrow5.469$ & $\uparrow0.722\downarrow4.028$ & $\uparrow0.363\downarrow13.576$ & $\uparrow0.412\downarrow5.417$ & $\uparrow0.622\downarrow4.864$ & $\uparrow0.663\downarrow3.094$ & $\uparrow0.608\downarrow5.207$ \\

                $\quad\quad \text{w/ } \mathcal L^{geo}_\text{B} $ (in 2D)  & $\uparrow0.354\downarrow21.926$ & $\uparrow0.558\downarrow12.403$ & $\uparrow0.078\downarrow33.166$ & $\uparrow0.326\downarrow13.904$ & $\uparrow0.323\downarrow26.795$ & $\uparrow0.492\downarrow8.188$ & $\uparrow0.447\downarrow16.626$  \\
                $\quad\quad \text{w/ } \mathcal L^{geo}_\text{B} $ (in 3D) & $\uparrow0.475\downarrow6.857$ & $\uparrow0.715\downarrow4.948$ & $\uparrow0.391\downarrow14.342$ & $\uparrow0.415\downarrow6.212$ & $\uparrow0.673\downarrow8.500$ & $\uparrow0.684\downarrow2.878$ & $\uparrow0.622\downarrow6.248$ \\

                $\quad\quad \text{w/ } \mathcal L^{int}_\text{B} $ (in 2D)  & $\uparrow0.256\downarrow57.424$ & $\uparrow0.571\downarrow15.279$ & $\uparrow0.052\downarrow25.584$ & $\uparrow0.322\downarrow19.557$ & $\uparrow0.330\downarrow68.651$ & $\uparrow0.409\downarrow23.349$ & $\uparrow0.420\downarrow29.978$\\
                $\quad\quad \text{w/ } \mathcal L^{int}_\text{B} $ (in 3D) & $\uparrow0.483\downarrow5.596$ & $\uparrow0.670\downarrow8.650$ & $\uparrow0.611\downarrow16.394$ & $\uparrow0.408\downarrow6.234$ & $\uparrow0.618\downarrow13.020$ & $\uparrow0.630\downarrow5.846$ & $\uparrow0.631\downarrow7.963$ \\

                 $\quad\quad \text{w/ } \mathcal L^{mbd}_\text{B} $ (in 2D) & $\uparrow0.468\downarrow7.716$ & $\uparrow0.634\downarrow16.368$ & $\uparrow0.218\downarrow29.664$ & $\uparrow0.386\downarrow11.425$ & $\uparrow0.530\downarrow20.234$ & $\uparrow0.598\downarrow7.755$ & $\uparrow0.547\downarrow13.309$  \\
                 $\quad\quad \text{w/ } \mathcal L^{mbd}_\text{B} $ (in 3D) & $\uparrow0.466\downarrow7.696$ & $\uparrow0.647\downarrow4.415$ & $\uparrow0.360\downarrow22.900$ & $\uparrow0.332\downarrow8.855$ & $\uparrow0.496\downarrow12.700$ & $\uparrow0.574\downarrow4.648$ & $\uparrow0.553\downarrow8.745$ \\
  
                \midrule
                $\quad\quad \text{w/ CRF-loss}$ \cite{tang2018regularized}  & $\uparrow0.396\downarrow8.835$ & $\uparrow0.685\downarrow6.413$ & $\uparrow0.758\downarrow19.622$ & $\uparrow0.448\downarrow5.738$ & $\uparrow0.695\downarrow8.316$ & $\uparrow0.661\downarrow5.657$ & $\uparrow0.663\downarrow7.797$  \\
                 \bottomrule
        \end{tabular}
        }
\end{table*}



  
  

%% file: poemcurves.tex
\begin{figure}[h!]
        \centering
        \pgfplotslegendfromname{common}\\
        \begin{tikzpicture}
                \begin{axis}[
                        name=acdcdsc,
                        width=0.62\textwidth,
                        height=0.35\textheight,
                        at={(0,0)},
                        ymin=0,
                        ymax=1,
                        xmin=0,
                        xmax=199,
                        ymajorgrids=true,
                        ylabel={DSC},
                        every axis y label/.style={
                                at={(ticklabel cs:0.5)},rotate=90,anchor=near ticklabel,
                        },
                        every axis x label/.style={
                                at={(ticklabel cs:0.5)},anchor=near ticklabel,
                        },
                        xlabel=Epoch,
                        xtick={0,25,50,75,100,125,150,175,199},
                        xticklabels={0,25,50,75,100,125,150,175,200},
                        ytick={0,0.1,...,1},
                        yticklabels={0.0,0.1,0.2,0.3,0.4,0.5,0.6,0.7,0.8,0.9,1.0},
                        legend pos=outer north east,
                        legend cell align=left,
                        tick label style={font=\scriptsize},
                        label style={font=\scriptsize},
                        legend style={draw=none},
                        legend style={line width=0.35mm},
                        title style={font=\scriptsize},
                        legend to name=common,
                        legend columns=3,
                        title={POEM, validation DSC (2D)}
                ]
                        \addplot[color=Maroon, thick, dashed] table [x=epoch, y=val_DiceFG, col sep=comma, mark=none] {results/ce_metrics.csv};
                        \addlegendentry{$\mathcal L_\text{CE}$, full supervision}
                        
                        \addplot[color=MidnightBlue, semithick] table [x=epoch, y=val_DiceFG, col sep=comma, mark=none] {results/euc_metrics.csv};
                        \addlegendentry{$\mathcal L_{\widetilde{\text{CE}}} + \mathcal L^{euc}_\text{B}$}

                        \addplot[color=SkyBlue, semithick] table [x=epoch, y=val_DiceFG, col sep=comma, mark=none] {results/geo_metrics.csv};
                        \addlegendentry{$\mathcal L_{\widetilde{\text{CE}}} + \mathcal L^{geo}_\text{B}$}

                        \addplot[color=Orange, thick, dashed] table [x=epoch, y=val_DiceFG, col sep=comma, mark=none] {results/cew_metrics.csv};
                        \addlegendentry{$\mathcal L_{\widetilde{\text{CE}}}$, point supervision}
                        
                        \addplot[color=OliveGreen, semithick] table [x=epoch, y=val_DiceFG, col sep=comma, mark=none] {results/int_metrics.csv};
                        \addlegendentry{$\mathcal L_{\widetilde{\text{CE}}} + \mathcal L^{int}_\text{B}$}

                        \addplot[color=SpringGreen, semithick] table [x=epoch, y=val_DiceFG, col sep=comma, mark=none] {results/mbd_metrics.csv};
                        \addlegendentry{$\mathcal L_{\widetilde{\text{CE}}} + \mathcal L^{mbd}_\text{B}$}

                        \addplot[color=Purple, semithick] table [x=epoch, y=val_DiceFG, col sep=comma, mark=none] {results/crf_metrics.csv};
                        \addlegendentry{$\mathcal L_{\widetilde{\text{CE}}} + \text{CRF-loss}$}
                        

                \end{axis}
        \end{tikzpicture}
        \caption{Curve evolution of the average (over foreground classes) validation batch (2D) Dice scores during training, on the POEM dataset in multi-label segmentation training. The boundary losses $\mathcal L_B^{\cdot}$ use distance maps calculated on 3D volumes.} 
        \label{fig:poem_curve}
\end{figure}

%% file: poemcurves3d.tex
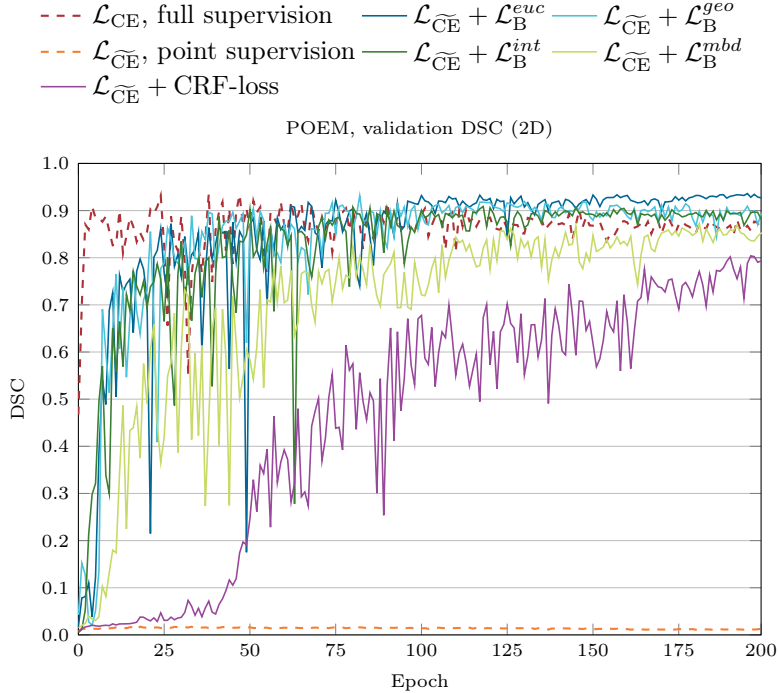
\begin{figure}[h!]
        \centering
        \pgfplotslegendfromname{common}\\
        \begin{tikzpicture}
                \begin{axis}[
                        name=acdcdsc,
                        width=0.62\textwidth,
                        height=0.35\textheight,
                        at={(0,0)},
                        ymin=0,
                        ymax=1,
                        xmin=0,
                        xmax=199,
                        ymajorgrids=true,
                        ylabel={DSC},
                        every axis y label/.style={
                                at={(ticklabel cs:0.5)},rotate=90,anchor=near ticklabel,
                        },
                        every axis x label/.style={
                                at={(ticklabel cs:0.5)},anchor=near ticklabel,
                        },
                        xlabel=Epoch,
                        xtick={0,25,50,75,100,125,150,175,199},
                        xticklabels={0,25,50,75,100,125,150,175,200},
                        ytick={0,0.1,...,1},
                        yticklabels={0.0,0.1,0.2,0.3,0.4,0.5,0.6,0.7,0.8,0.9,1.0},
                        legend pos=outer north east,
                        legend cell align=left,
                        tick label style={font=\scriptsize},
                        label style={font=\scriptsize},
                        legend style={draw=none},
                        legend style={line width=0.35mm},
                        title style={font=\scriptsize},
                        legend to name=common,
                        legend columns=3,
                        title={POEM, validation DSC (2D)}
                ]
                        \addplot[color=Maroon, thick, dashed] table [x=epoch, y=val_DiceFG, col sep=comma, mark=none] {results/ce_metrics.csv};
                        \addlegendentry{$\mathcal L_\text{CE}$, full supervision}
                        
                        \addplot[color=MidnightBlue, semithick] table [x=epoch, y=val_DiceFG, col sep=comma, mark=none] {results/euc3d_metrics.csv};
                        \addlegendentry{$\mathcal L_{\widetilde{\text{CE}}} + \mathcal L^{euc}_\text{B}$}

                        \addplot[color=SkyBlue, semithick] table [x=epoch, y=val_DiceFG, col sep=comma, mark=none] {results/geo3d_metrics.csv};
                        \addlegendentry{$\mathcal L_{\widetilde{\text{CE}}} + \mathcal L^{geo}_\text{B}$}

                        \addplot[color=Orange, thick, dashed] table [x=epoch, y=val_DiceFG, col sep=comma, mark=none] {results/cew_metrics.csv};
                        \addlegendentry{$\mathcal L_{\widetilde{\text{CE}}}$, point supervision}
                        
                        \addplot[color=OliveGreen, semithick] table [x=epoch, y=val_DiceFG, col sep=comma, mark=none] {results/int3d_metrics.csv};
                        \addlegendentry{$\mathcal L_{\widetilde{\text{CE}}} + \mathcal L^{int}_\text{B}$}

                        \addplot[color=SpringGreen, semithick] table [x=epoch, y=val_DiceFG, col sep=comma, mark=none] {results/mbd3d_metrics.csv};
                        \addlegendentry{$\mathcal L_{\widetilde{\text{CE}}} + \mathcal L^{mbd}_\text{B}$}

                        \addplot[color=Purple, semithick] table [x=epoch, y=val_DiceFG, col sep=comma, mark=none] {results/crf_metrics.csv};
                        \addlegendentry{$\mathcal L_{\widetilde{\text{CE}}} + \text{CRF-loss}$}
                        

                \end{axis}
        \end{tikzpicture}
        \caption{Curve evolution of the average (over foreground classes) validation batch (2D) Dice scores during training, on the POEM dataset in multi-label segmentation training. The boundary losses $\mathcal{L}_B^{\cdot}$ use distance maps calculated on 3D volumes. The curves for training with $\mathcal{L}_{\text{CE}}$, $\mathcal{L}_{\widetilde{\text{CE}}}$ and CRF-loss are plotted again for easier comparison.} 
        \label{fig:poem_curve3d}
\end{figure}

%% file: examplesPOEM.tex
\begin{sidewaysfigure}
                        \centering
                        \begin{subfigure}[b]{0.11\textwidth}
                                \centering
                                \includegraphics[width=\textwidth]{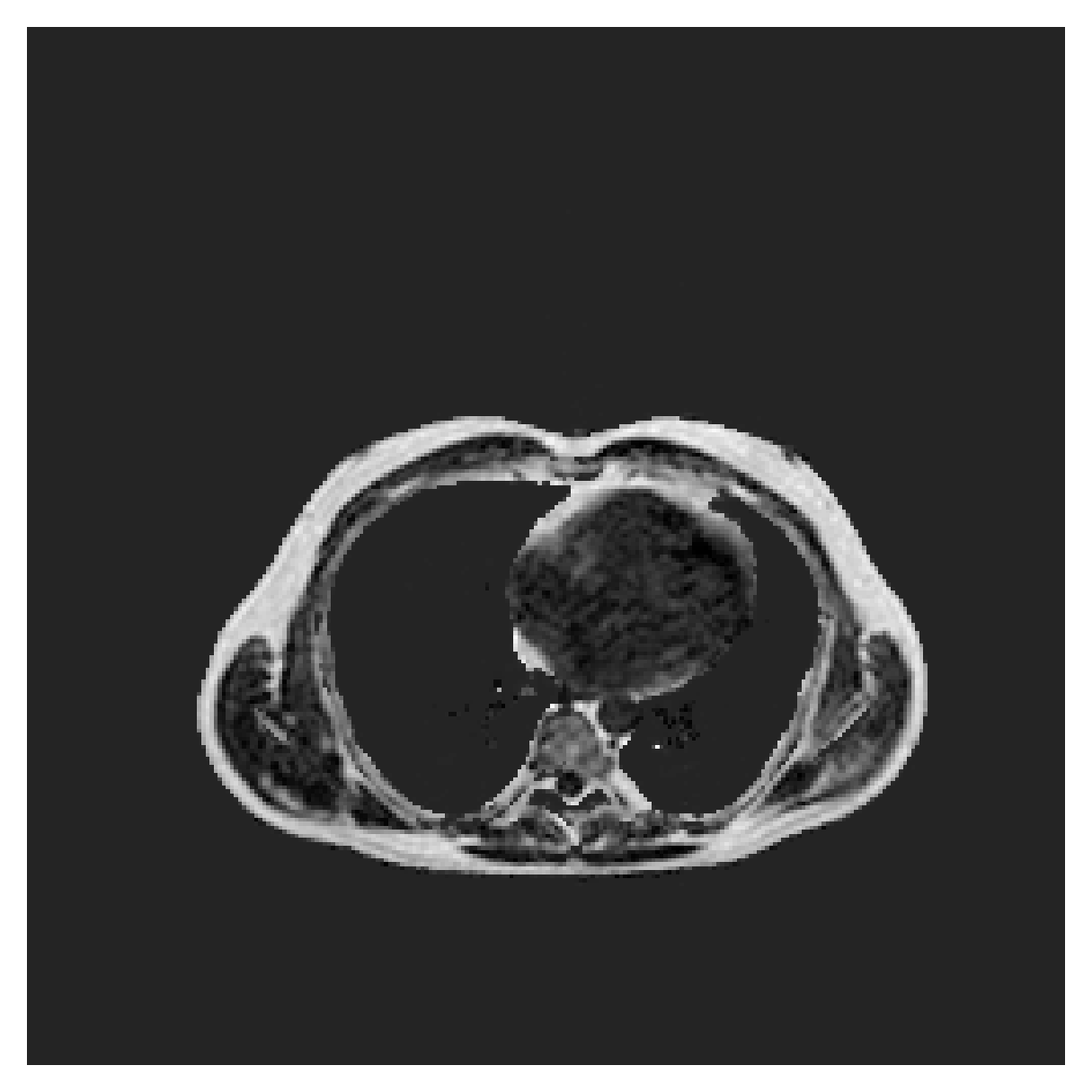}\\
                                \includegraphics[width=\textwidth]{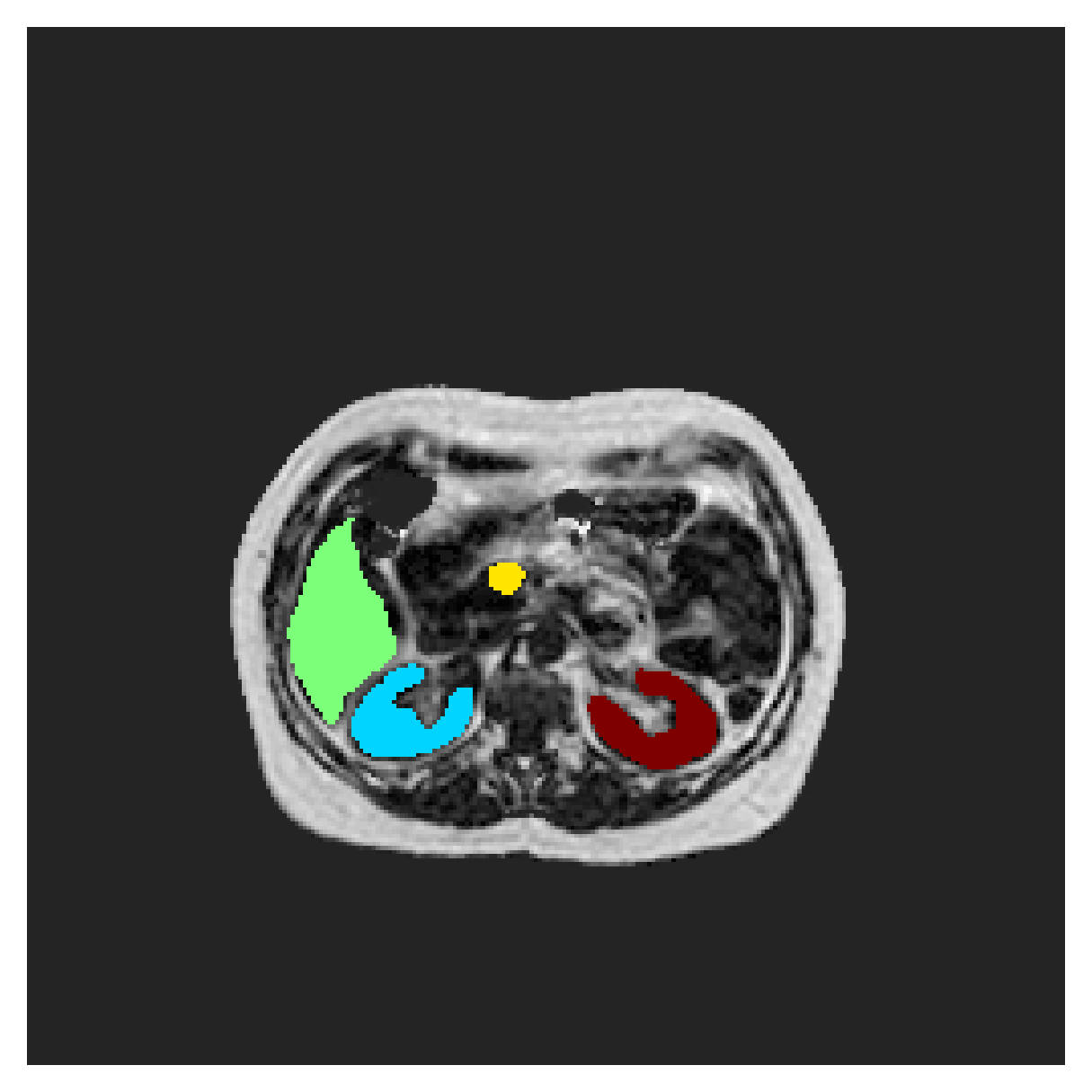}\\
                                \includegraphics[width=\textwidth]{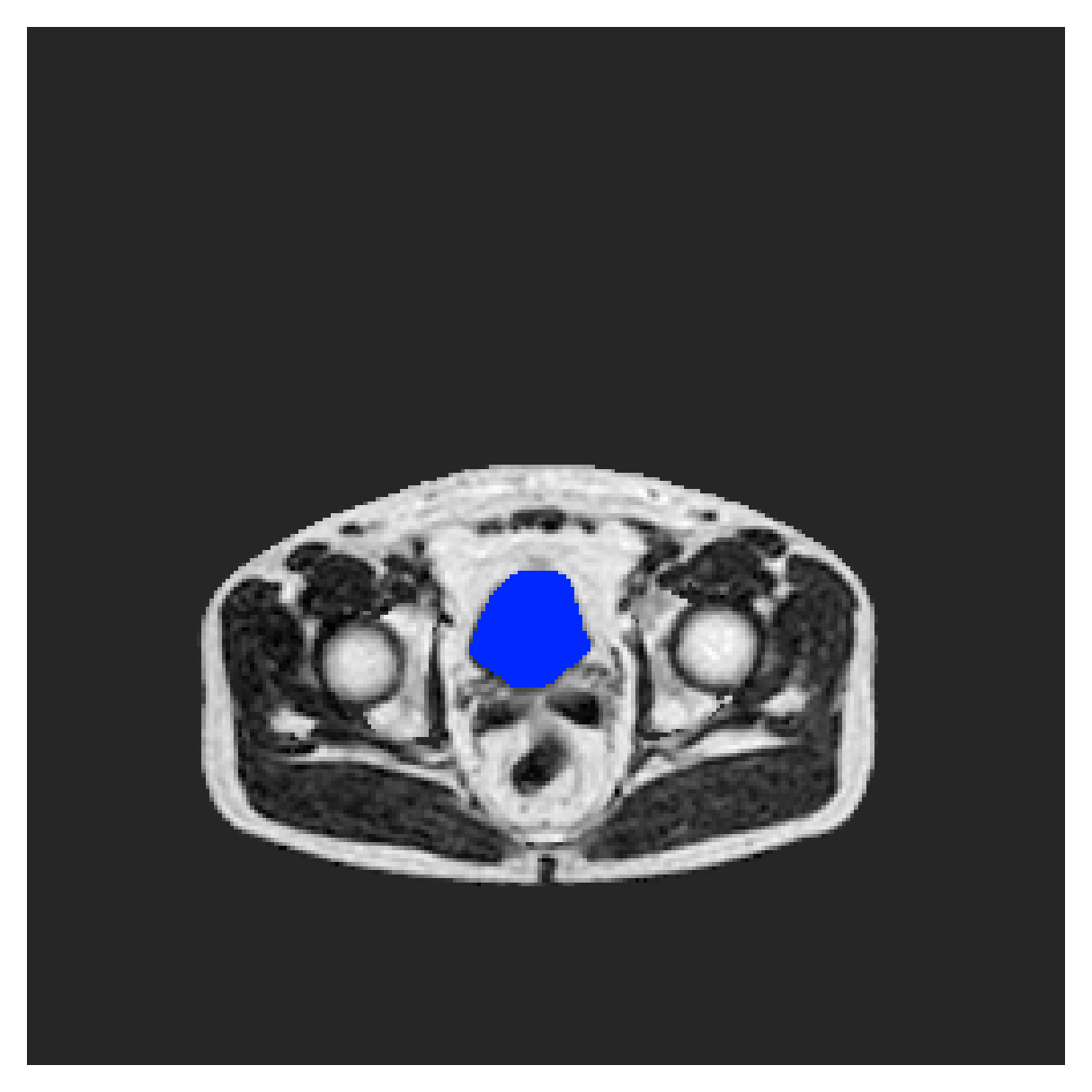}\\
                                \includegraphics[width=\textwidth]{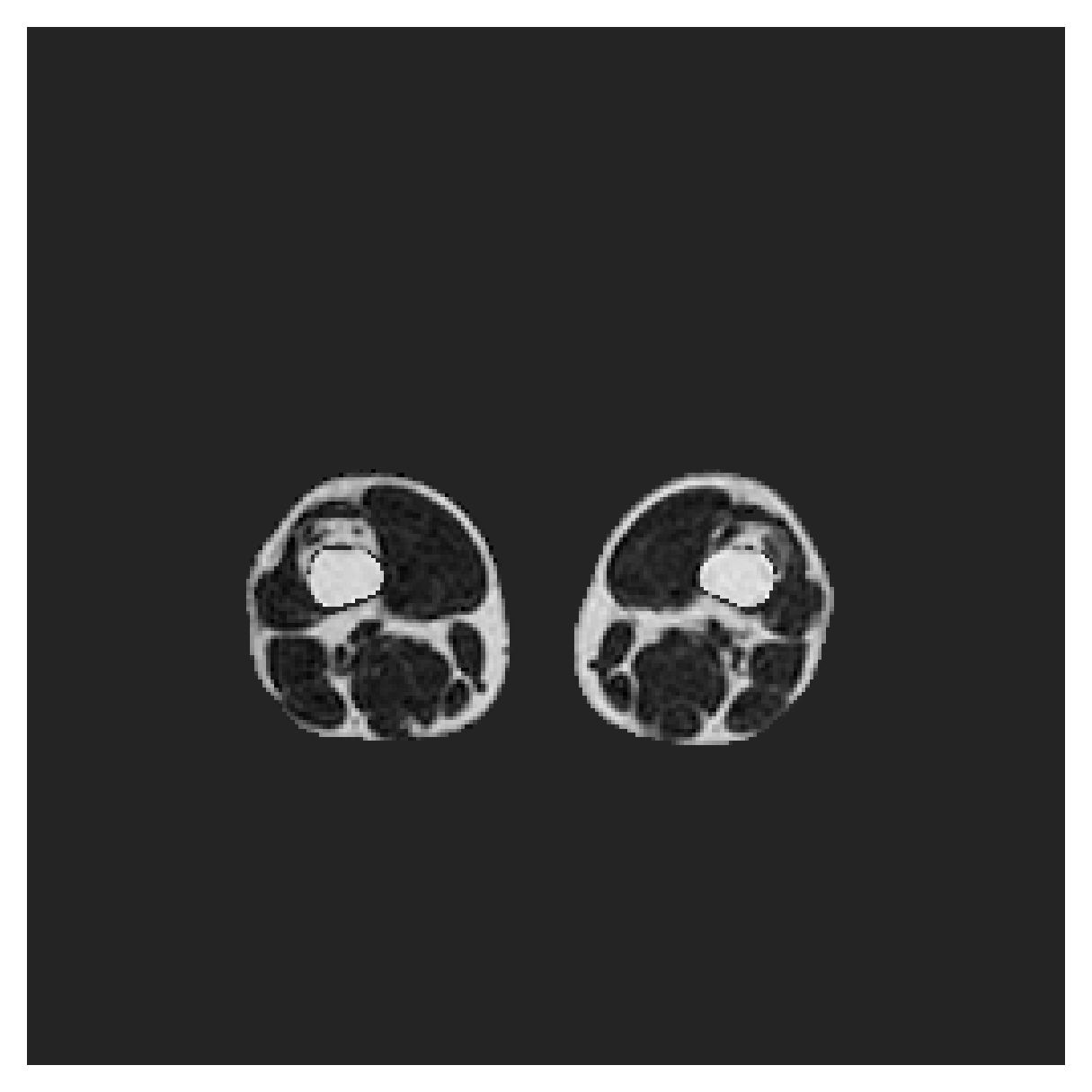}\\
                                \includegraphics[width=\textwidth]{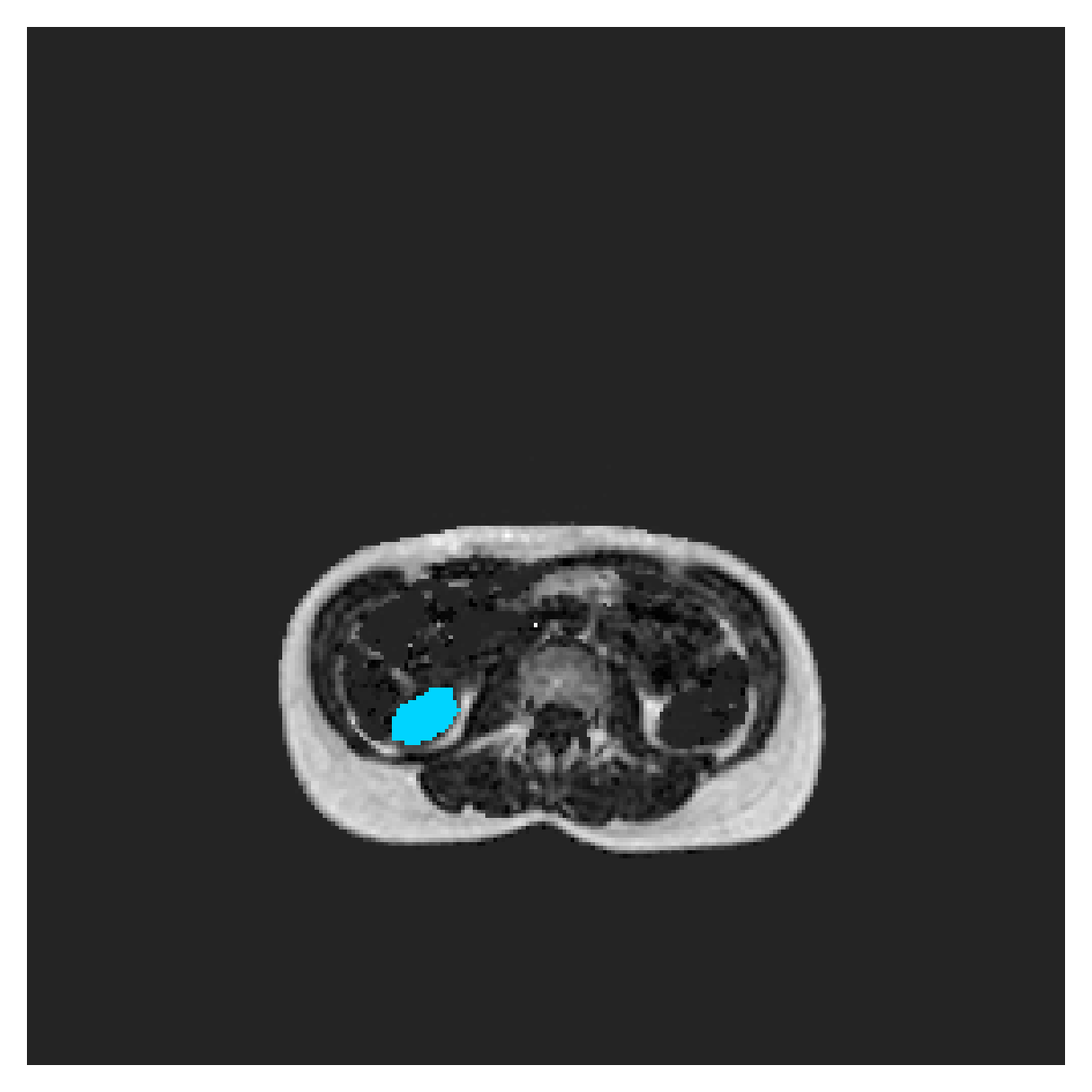}
                                \caption{Ground truth\\ $\qquad$}
                                \label{subfig:gt1b}
                        \end{subfigure}
                        \begin{subfigure}[b]{0.11\textwidth}
                                \centering
                                \includegraphics[width=\textwidth]{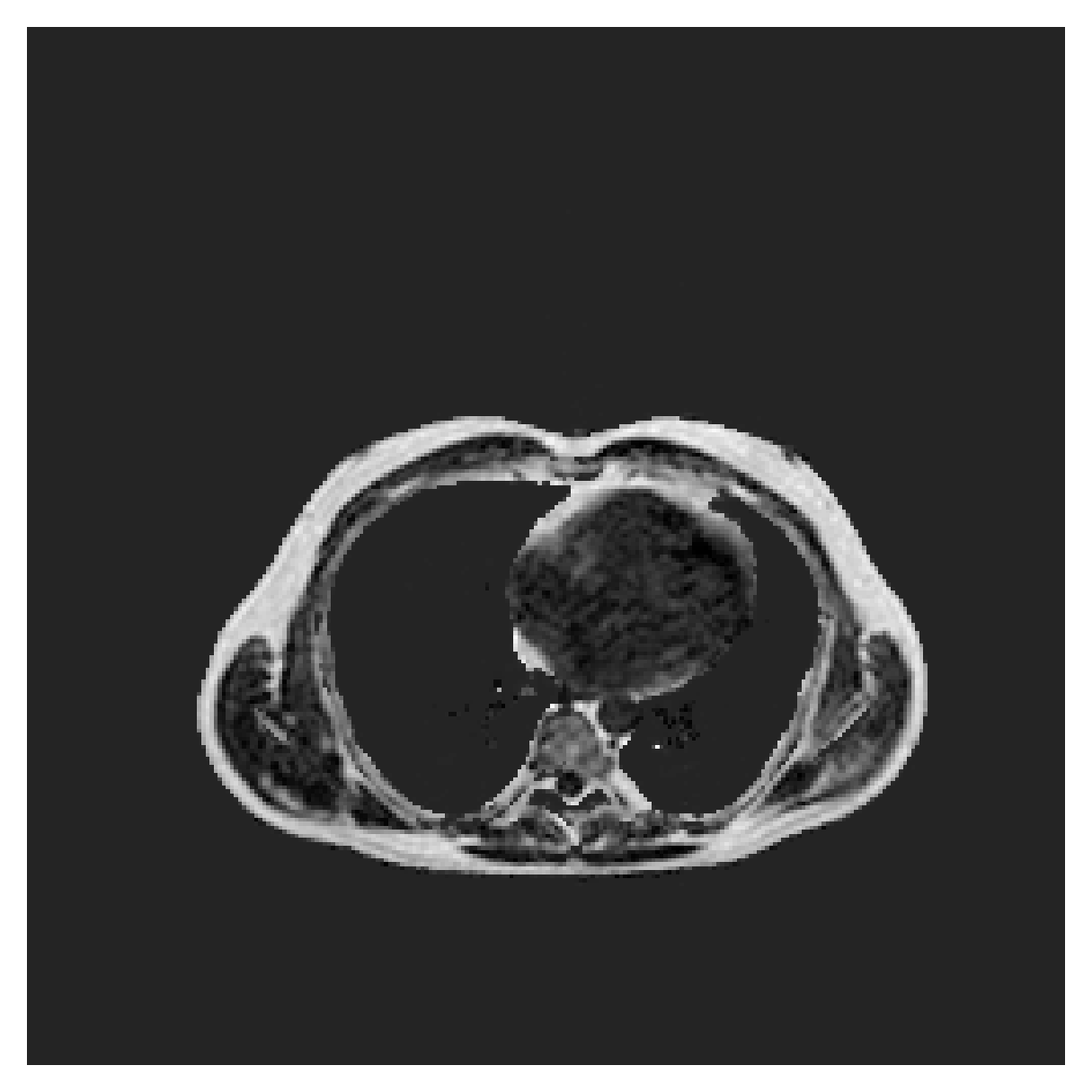}\\
                                \includegraphics[width=\textwidth]{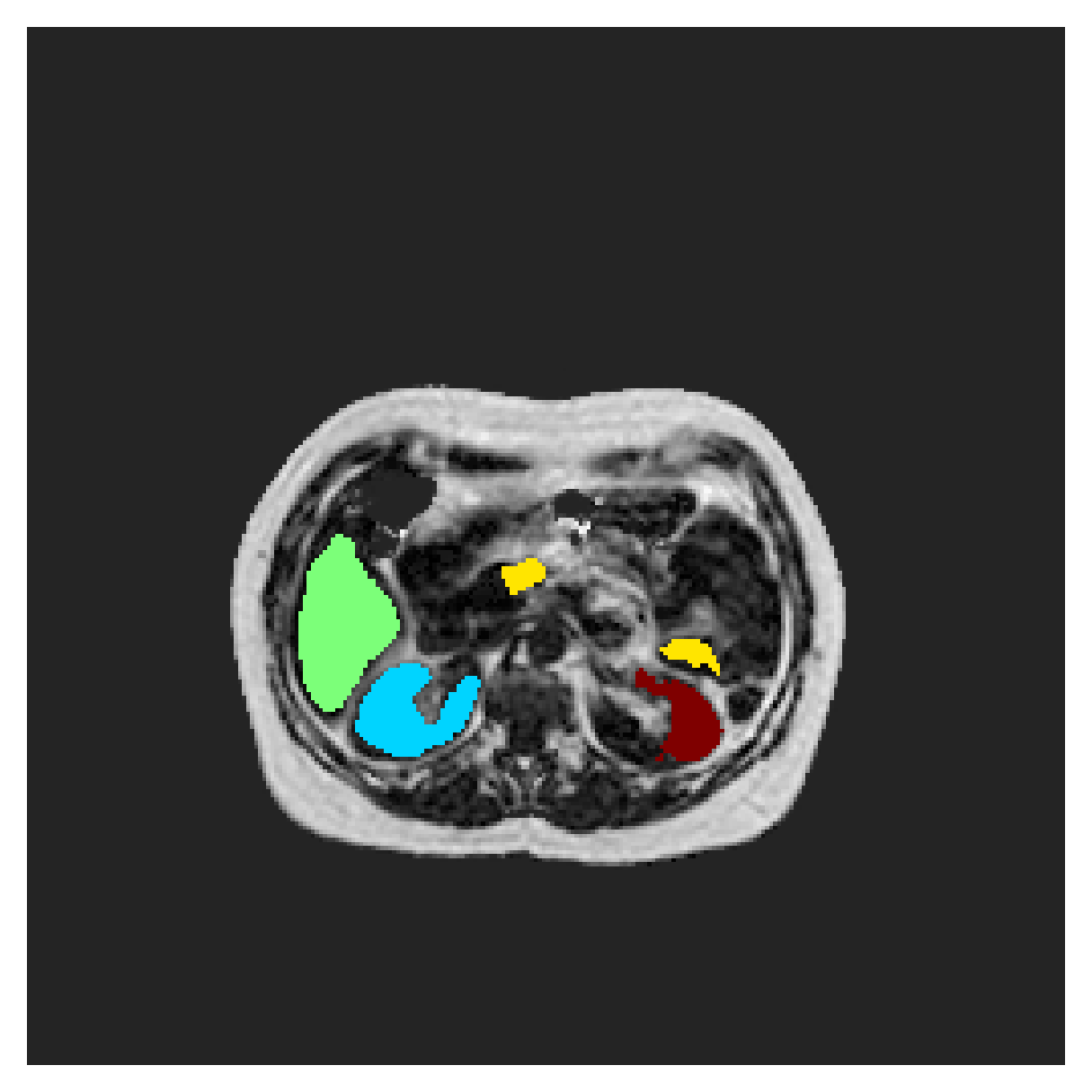}\\
                                \includegraphics[width=\textwidth]{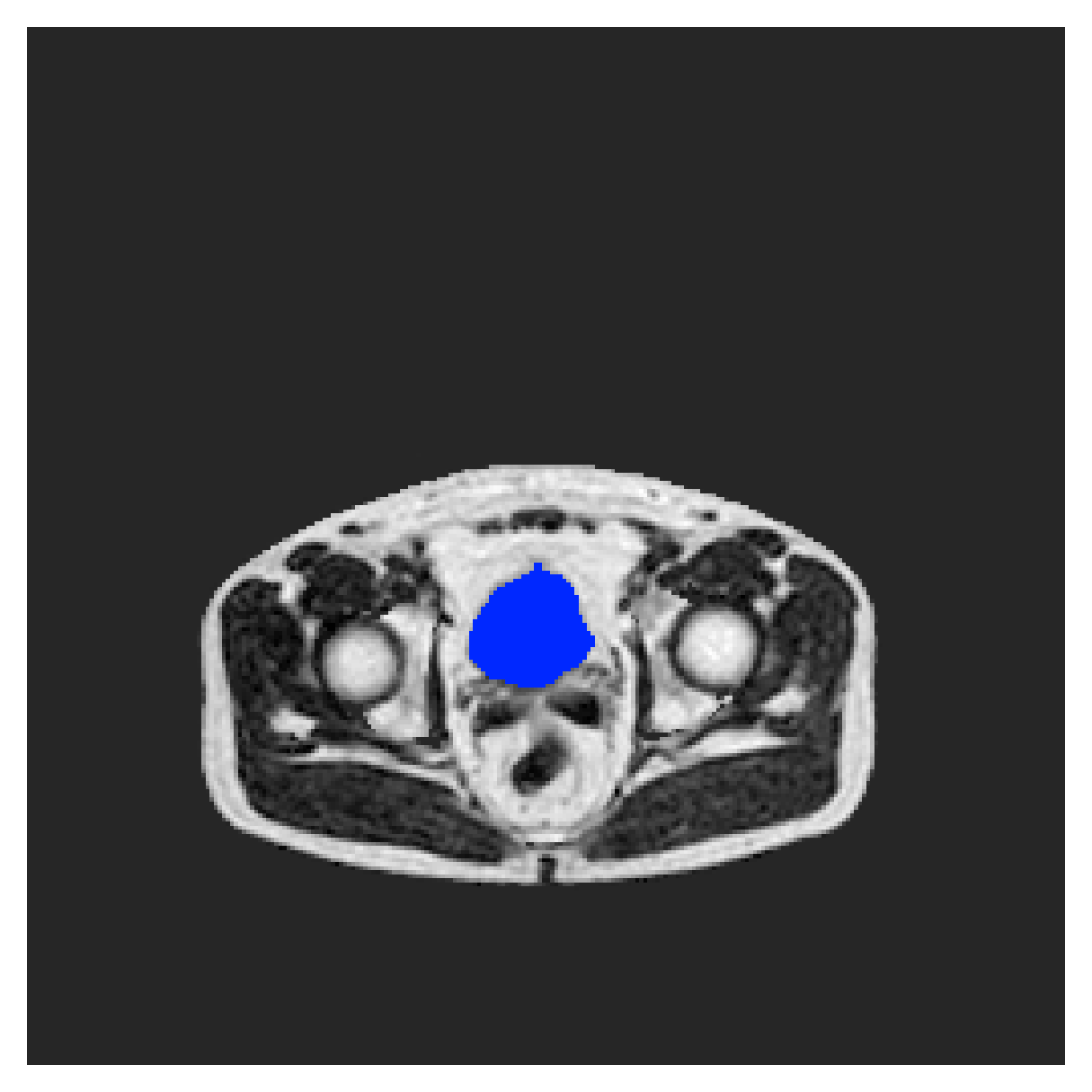}\\
                                \includegraphics[width=\textwidth]{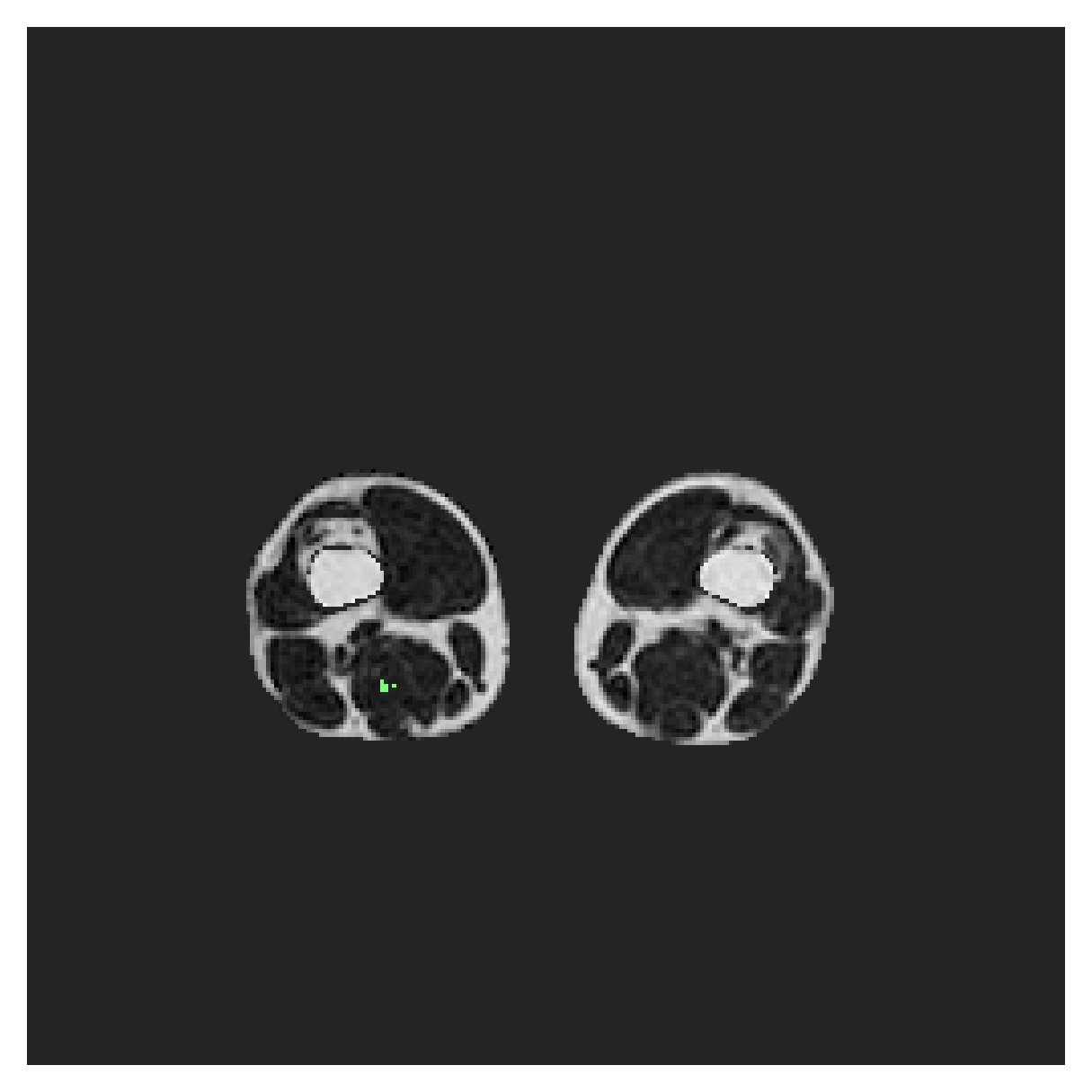}\\
                                \includegraphics[width=\textwidth]{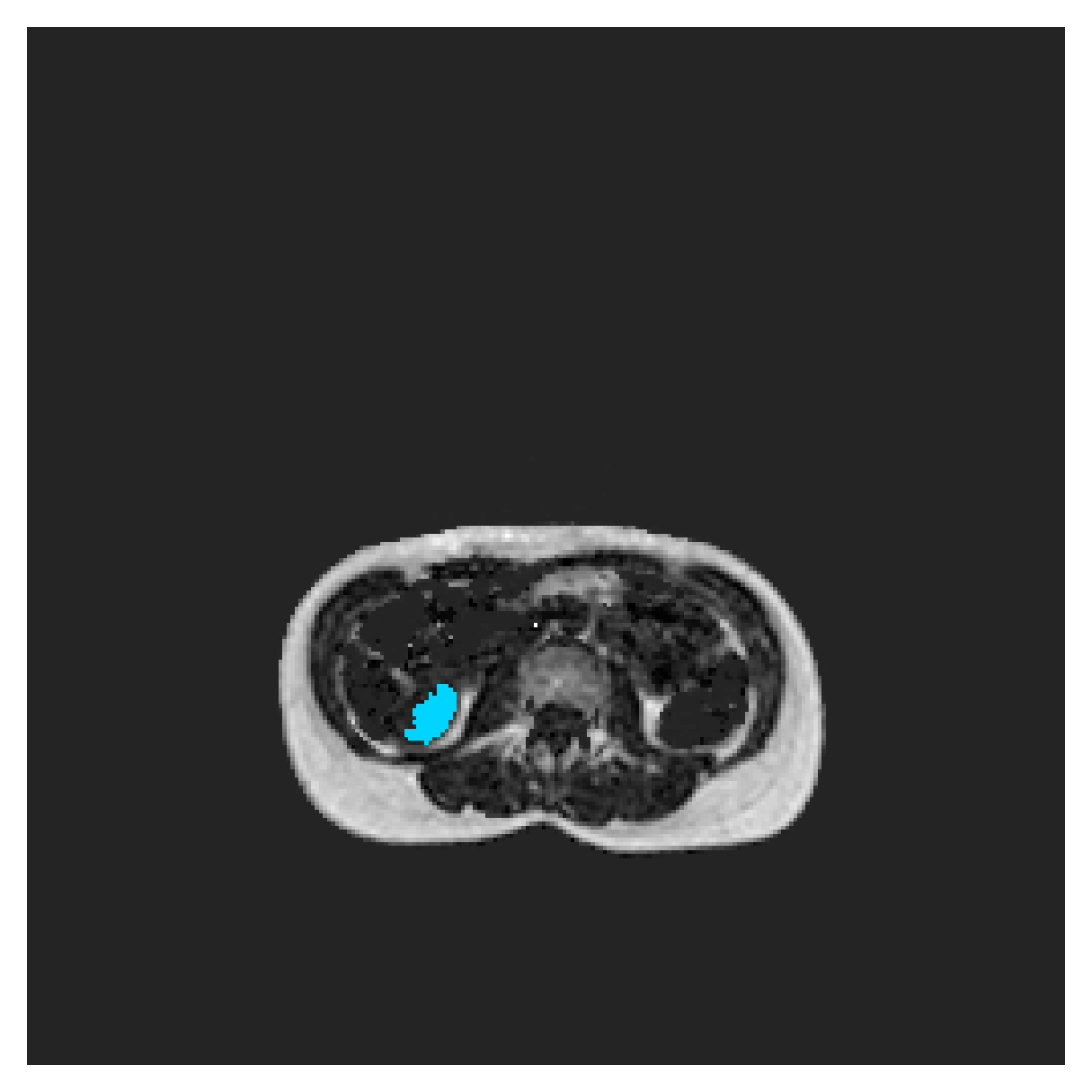}
                                \caption{$\mathcal L_{\text{CE}}$, full supervision} 
                                \label{subfig:ce1b}
                        \end{subfigure}
                        \begin{subfigure}[b]{0.11\textwidth}
                                \centering
                                \includegraphics[width=\textwidth]{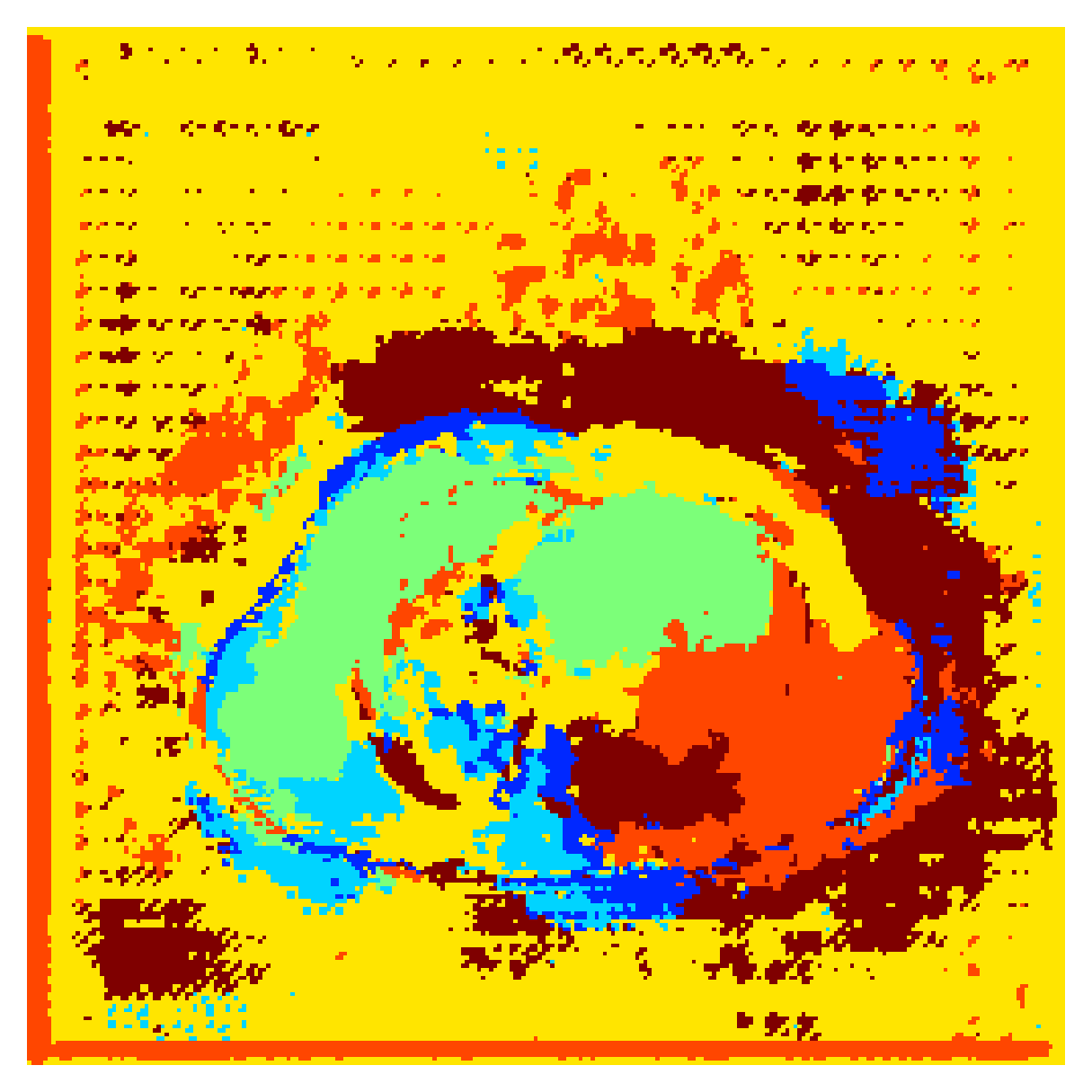}\\
                                \includegraphics[width=\textwidth]{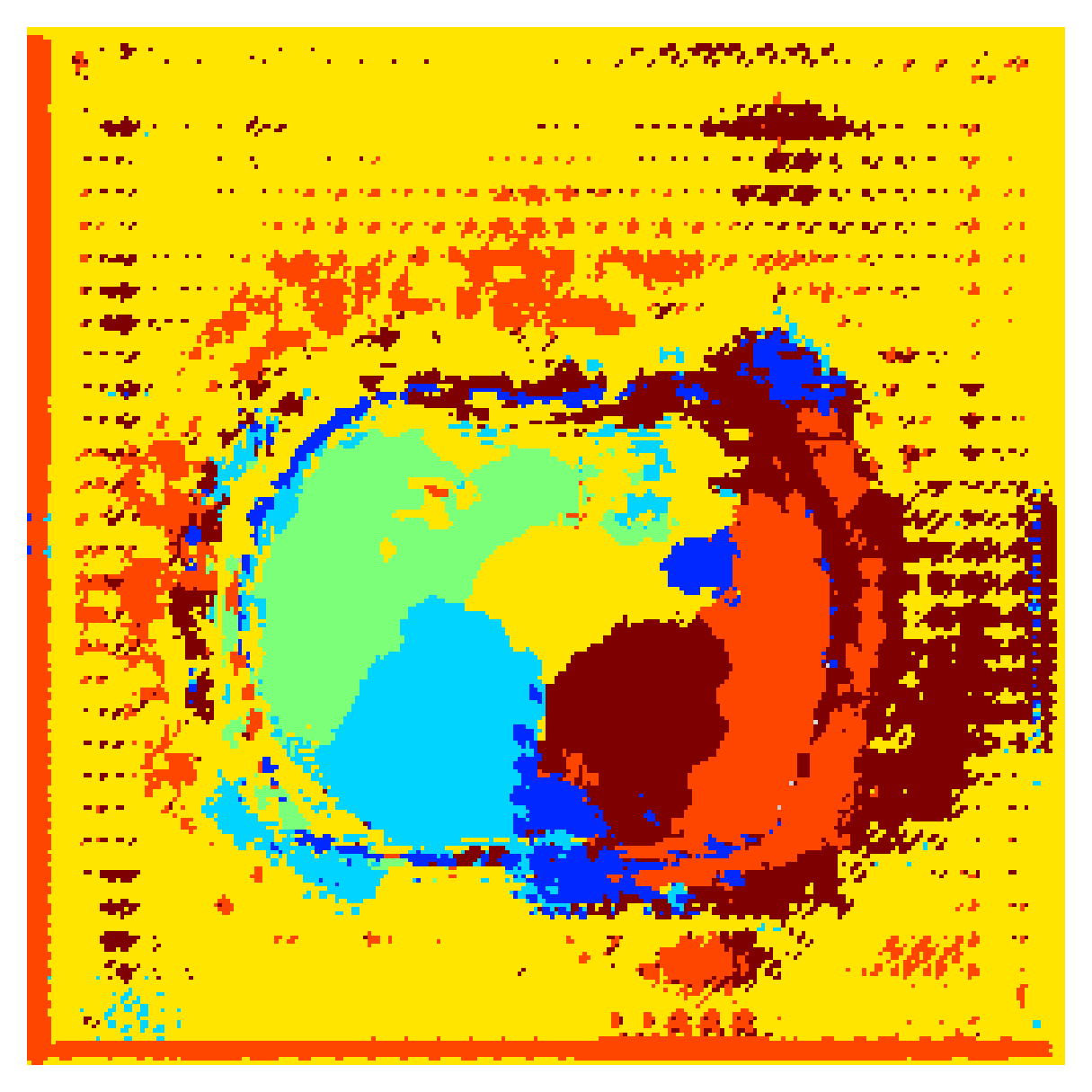}\\
                                \includegraphics[width=\textwidth]{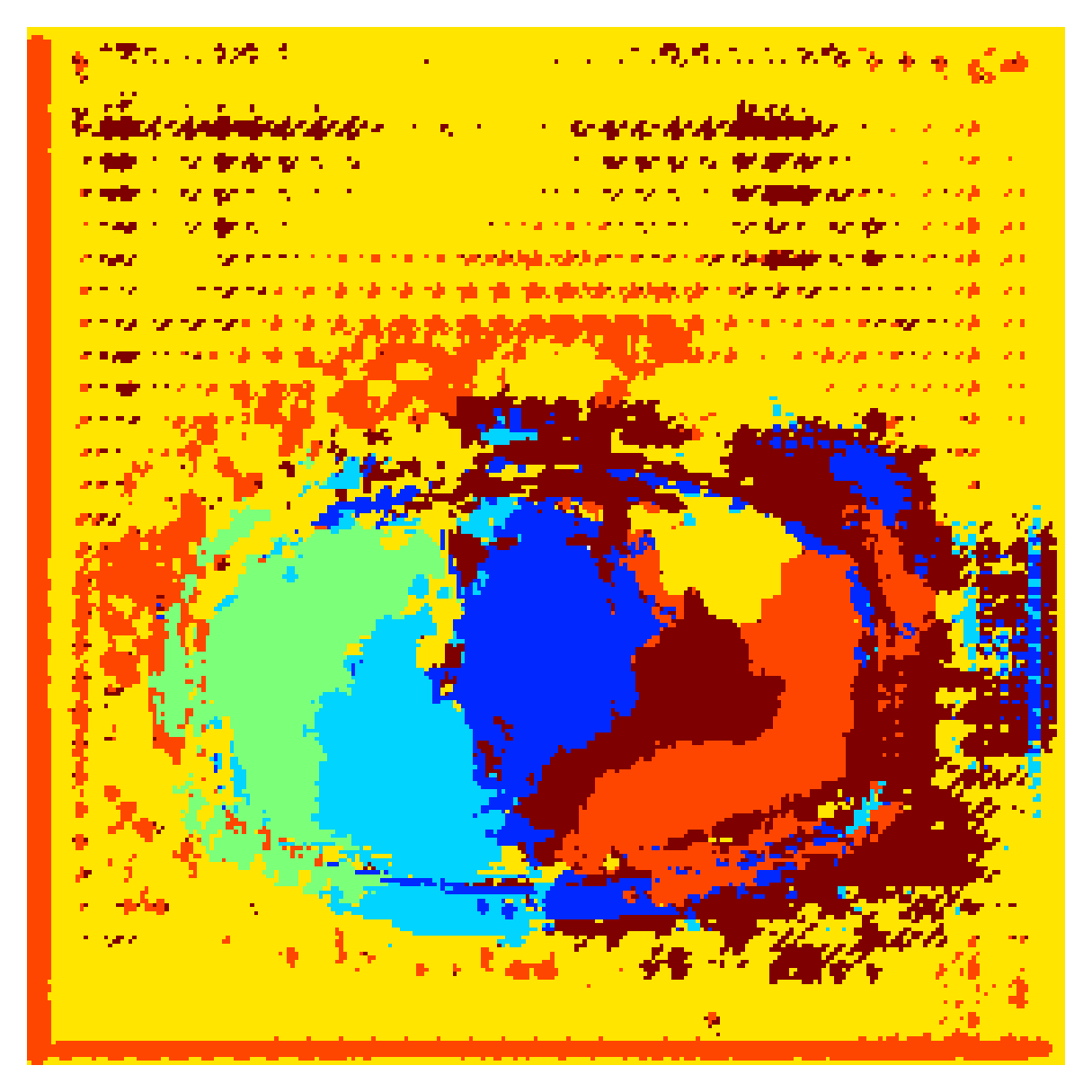}\\
                                \includegraphics[width=\textwidth]{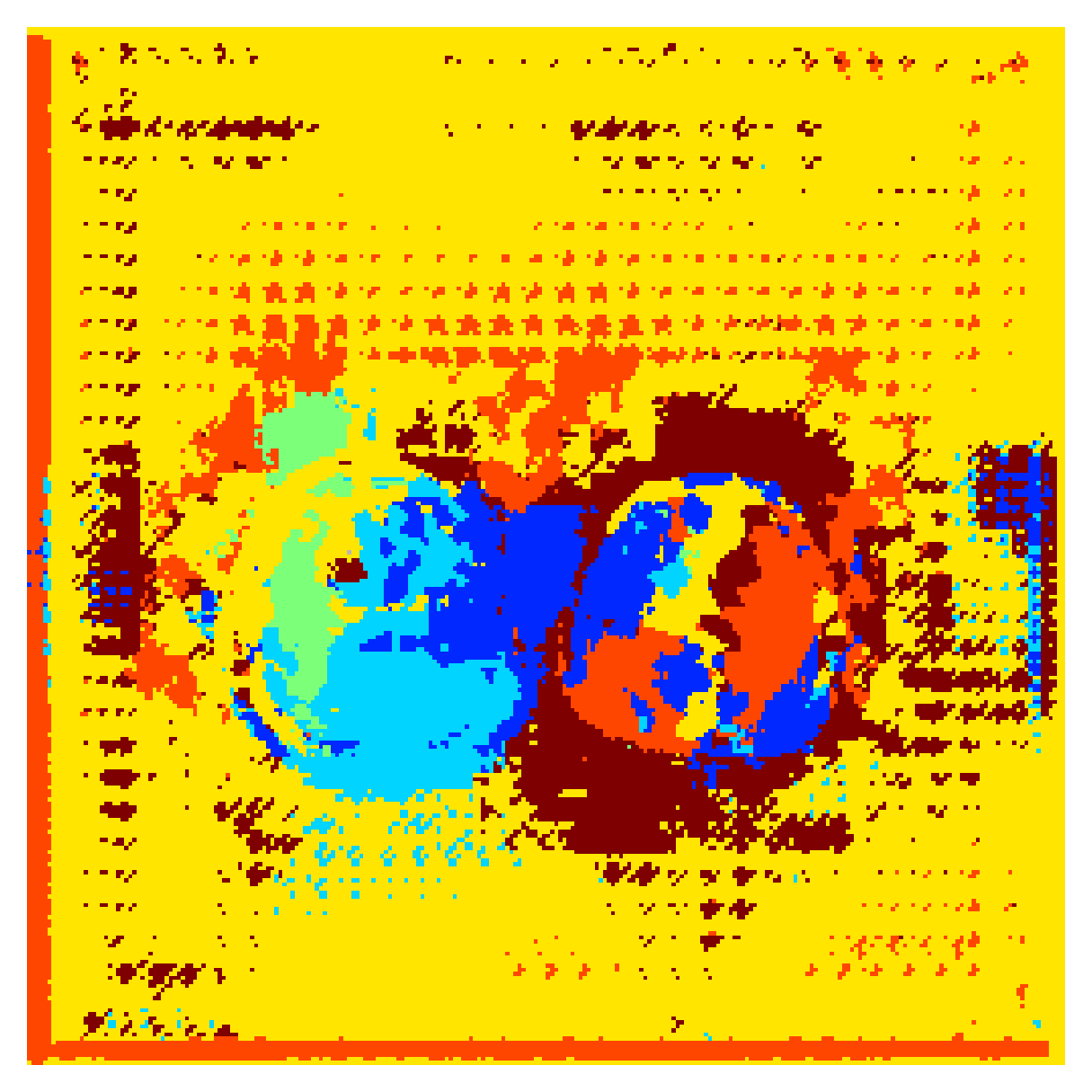}\\
                                \includegraphics[width=\textwidth]{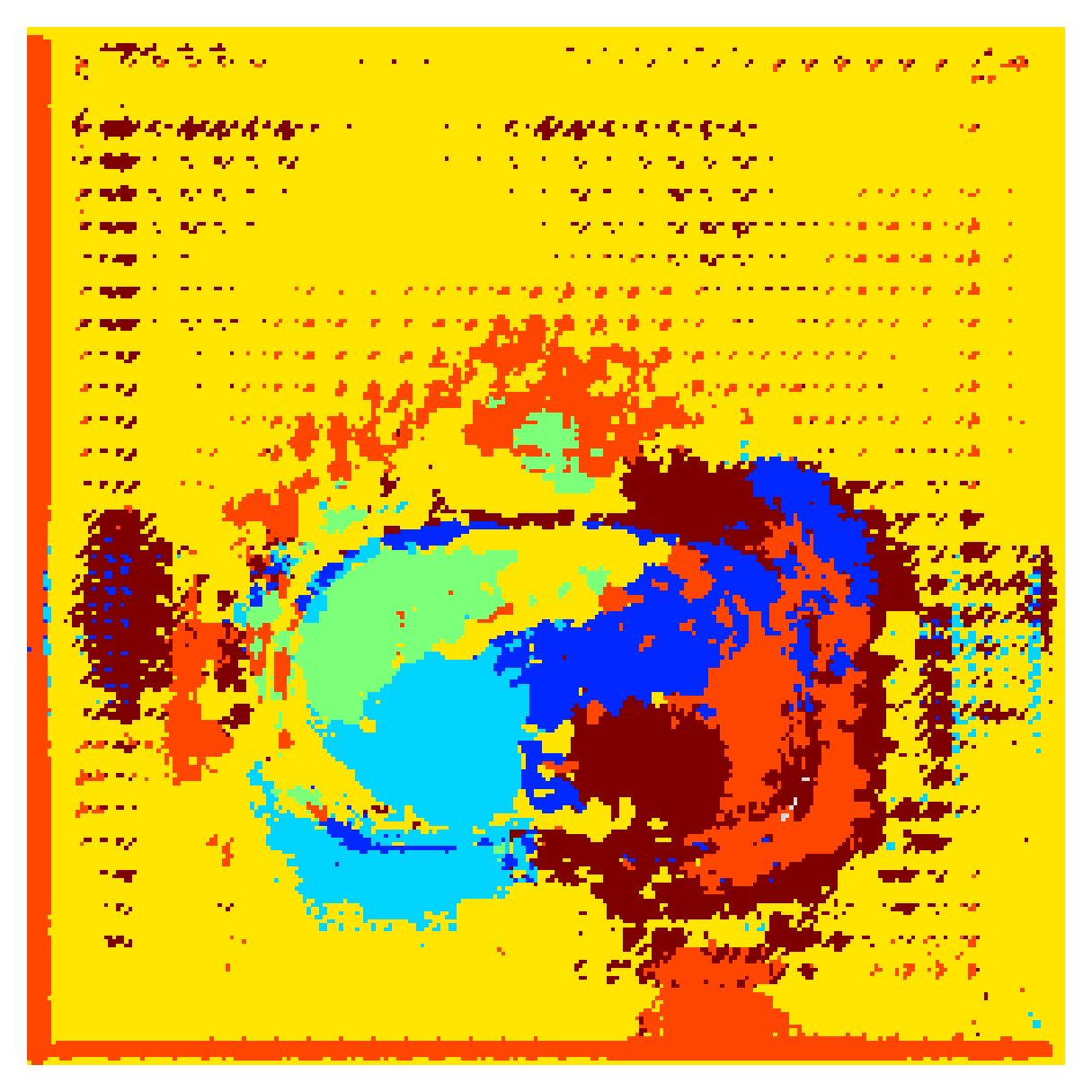}
                                \caption{$\mathcal L_{\widetilde{\text{CE}}}$, weak supervision}
                                \label{subfig:cew1b}
                        \end{subfigure}
                        \begin{subfigure}[b]{0.11\textwidth}
                                \centering
                                \includegraphics[width=\textwidth]{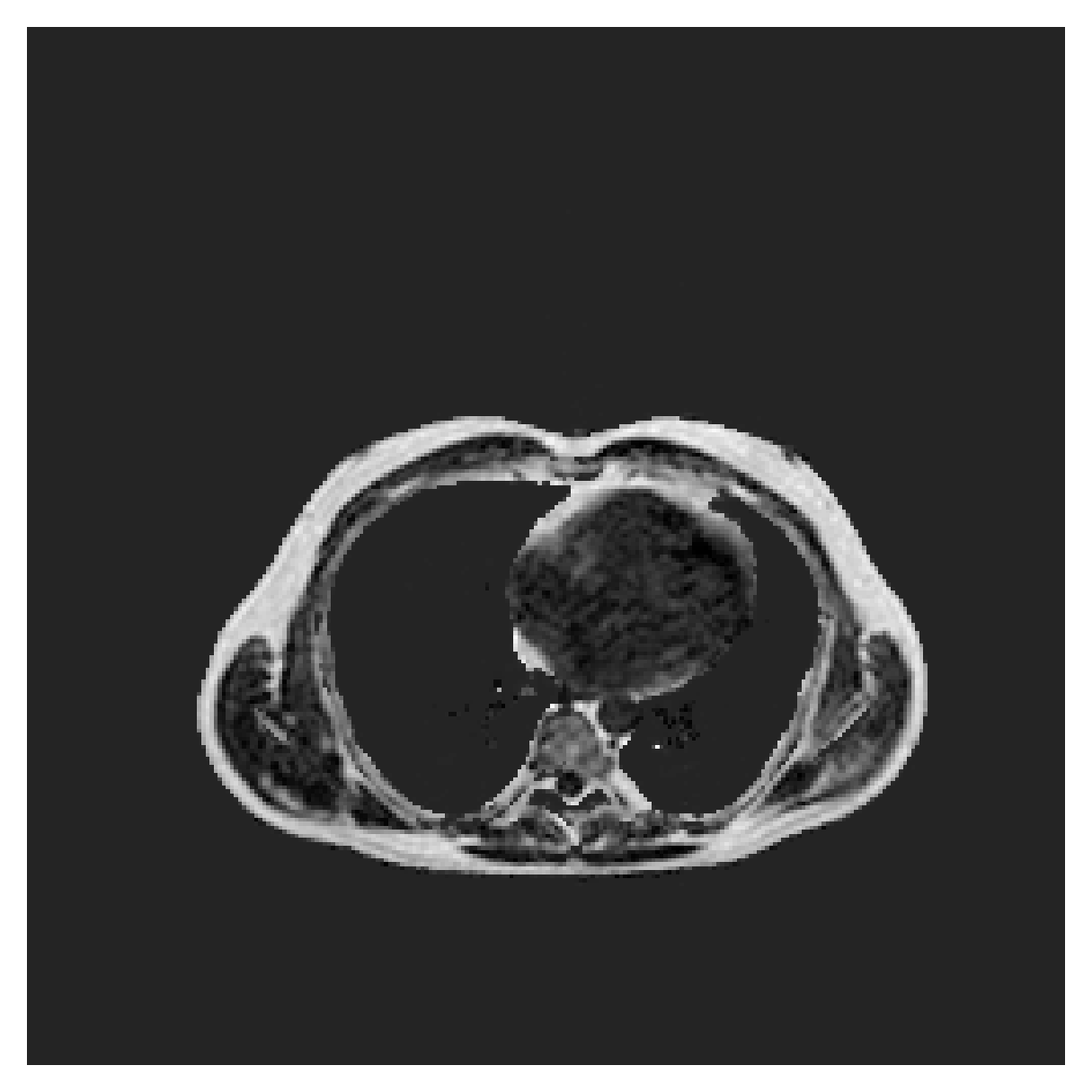}\\
                                \includegraphics[width=\textwidth]{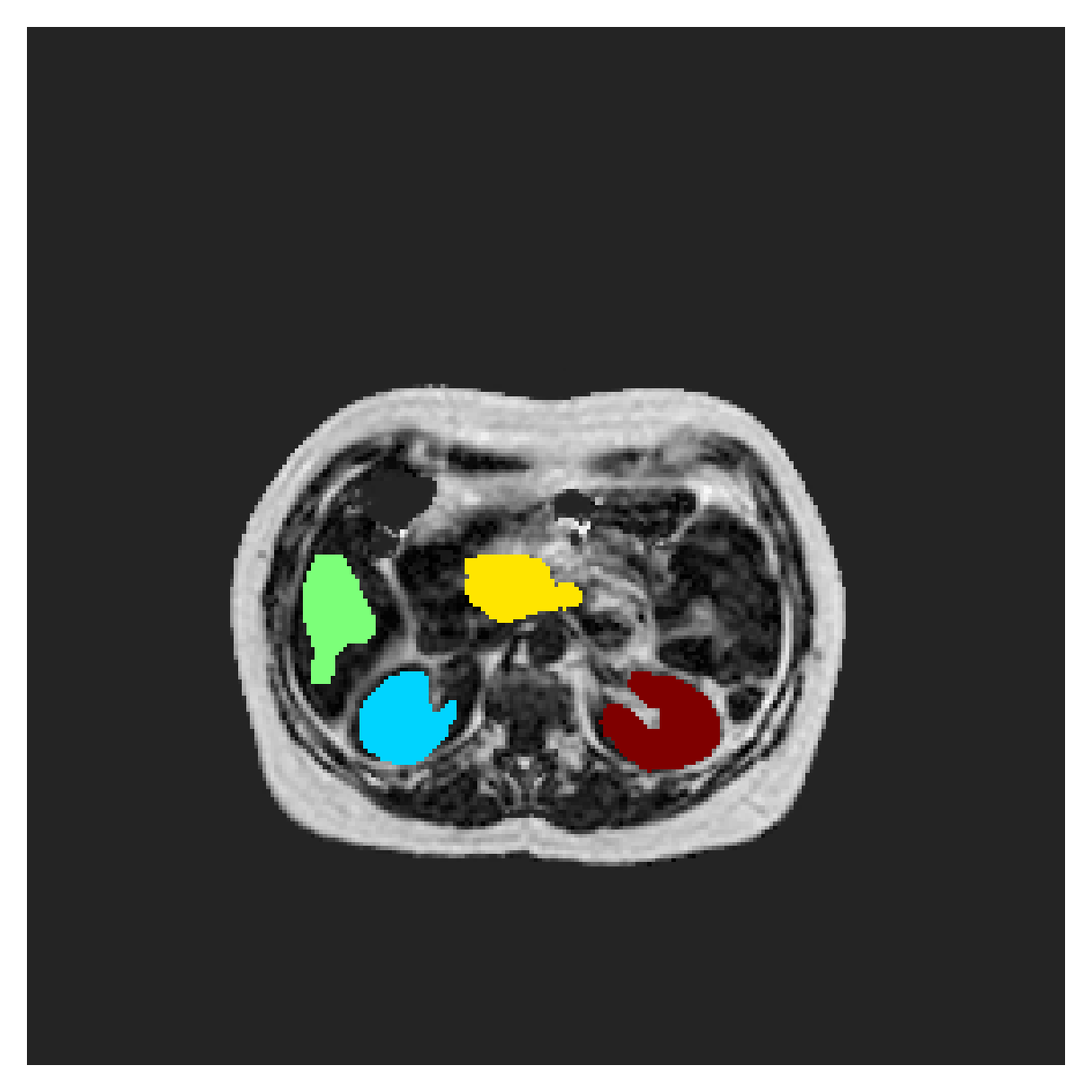}\\
                                \includegraphics[width=\textwidth]{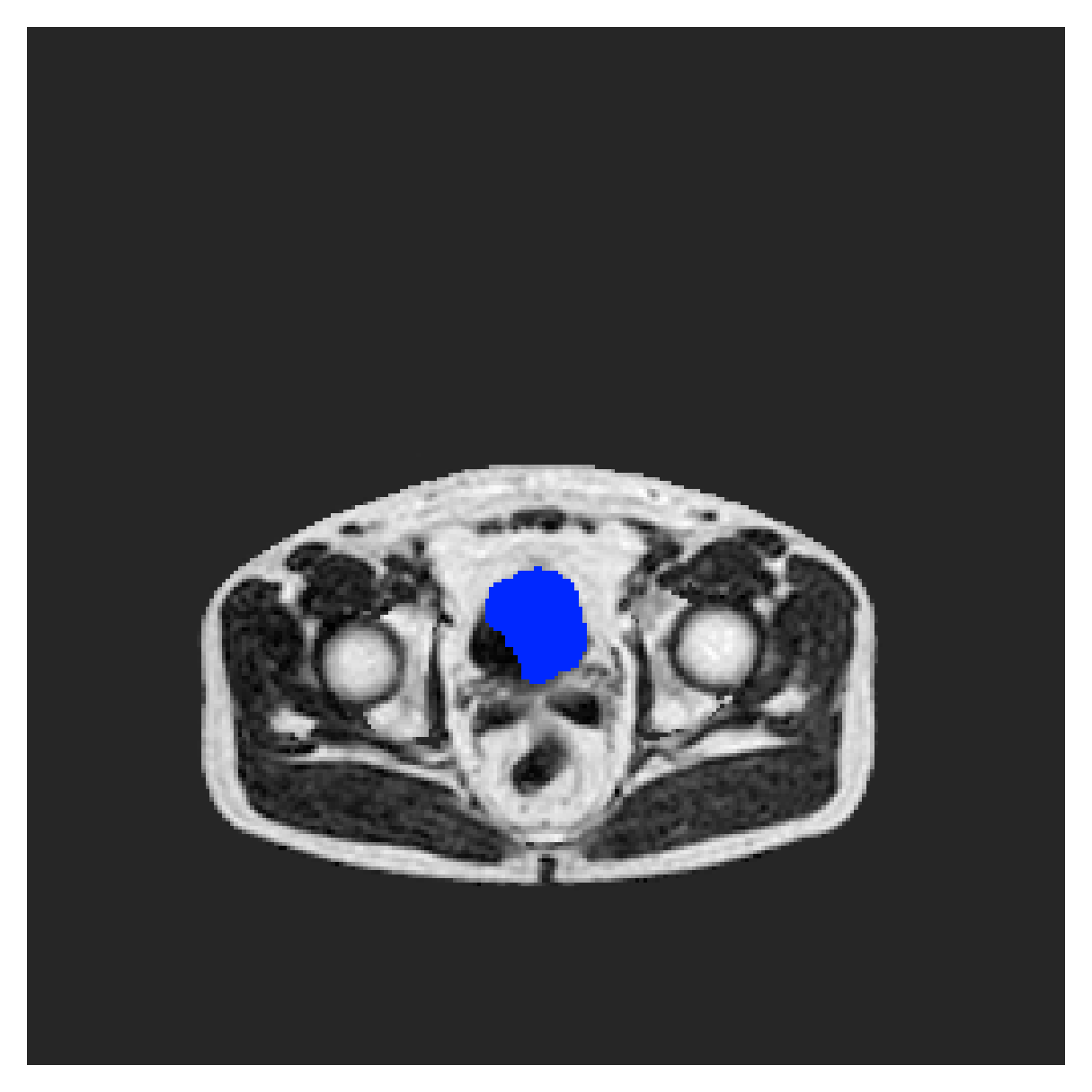}\\
                                \includegraphics[width=\textwidth]{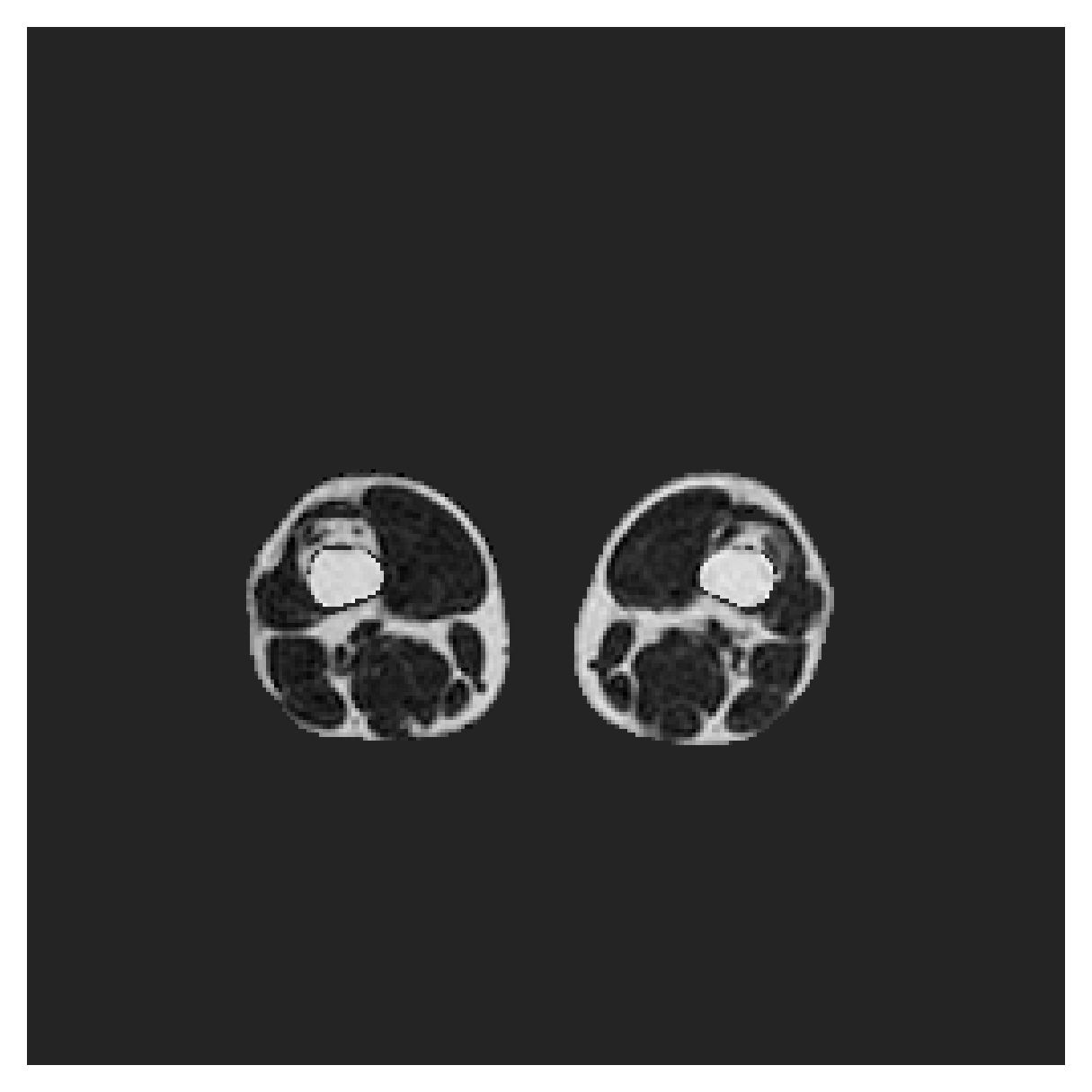}\\
                                \includegraphics[width=\textwidth]{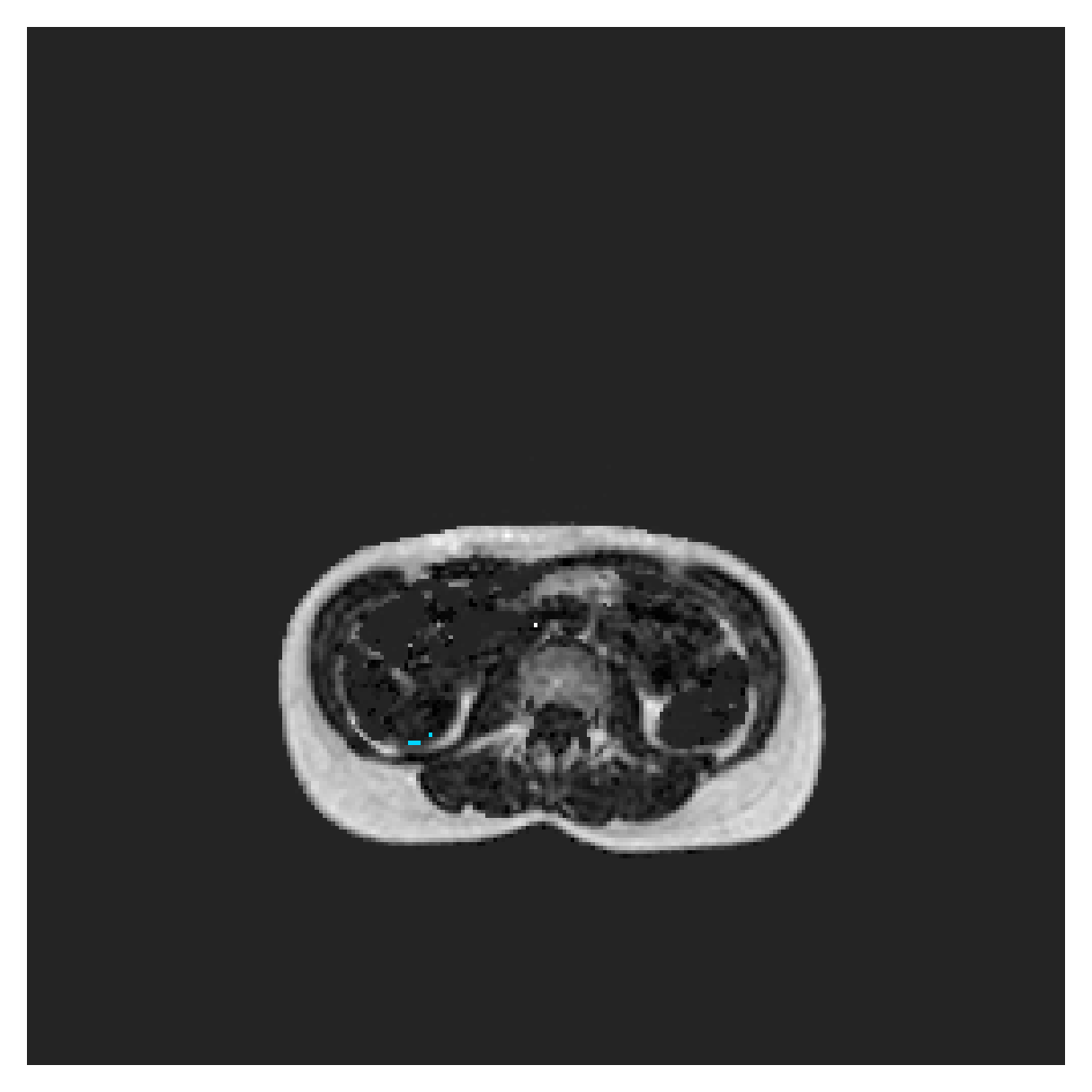}
                                \caption{$\mathcal L_{\widetilde{\text{CE}}} + \mathcal{L}_B^{euc}$\\ $\qquad$}
                                \label{subfig:euc1b}
                        \end{subfigure}
                        \begin{subfigure}[b]{0.11\textwidth}
                                \centering
                                \includegraphics[width=\textwidth]{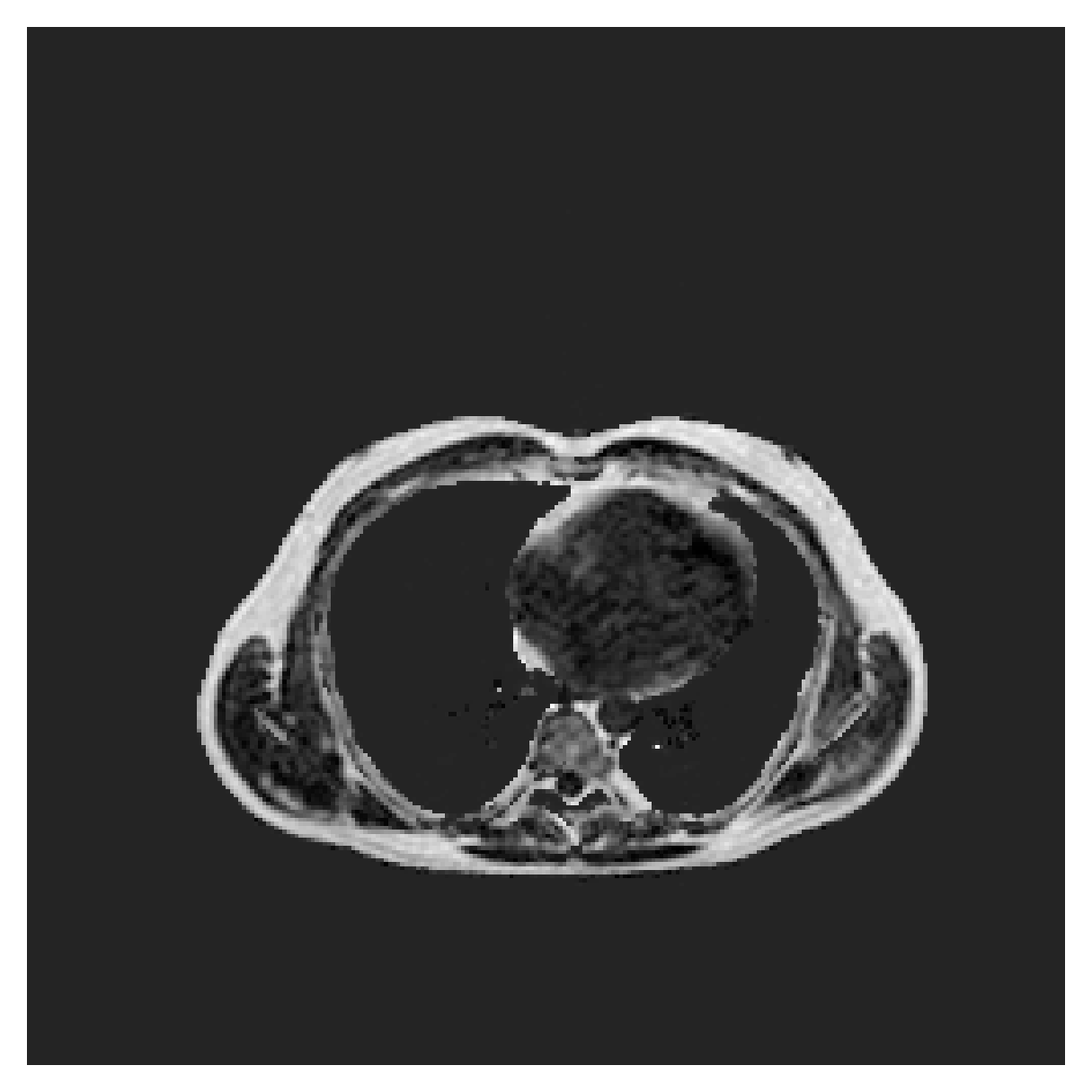}\\ 
                                \includegraphics[width=\textwidth]{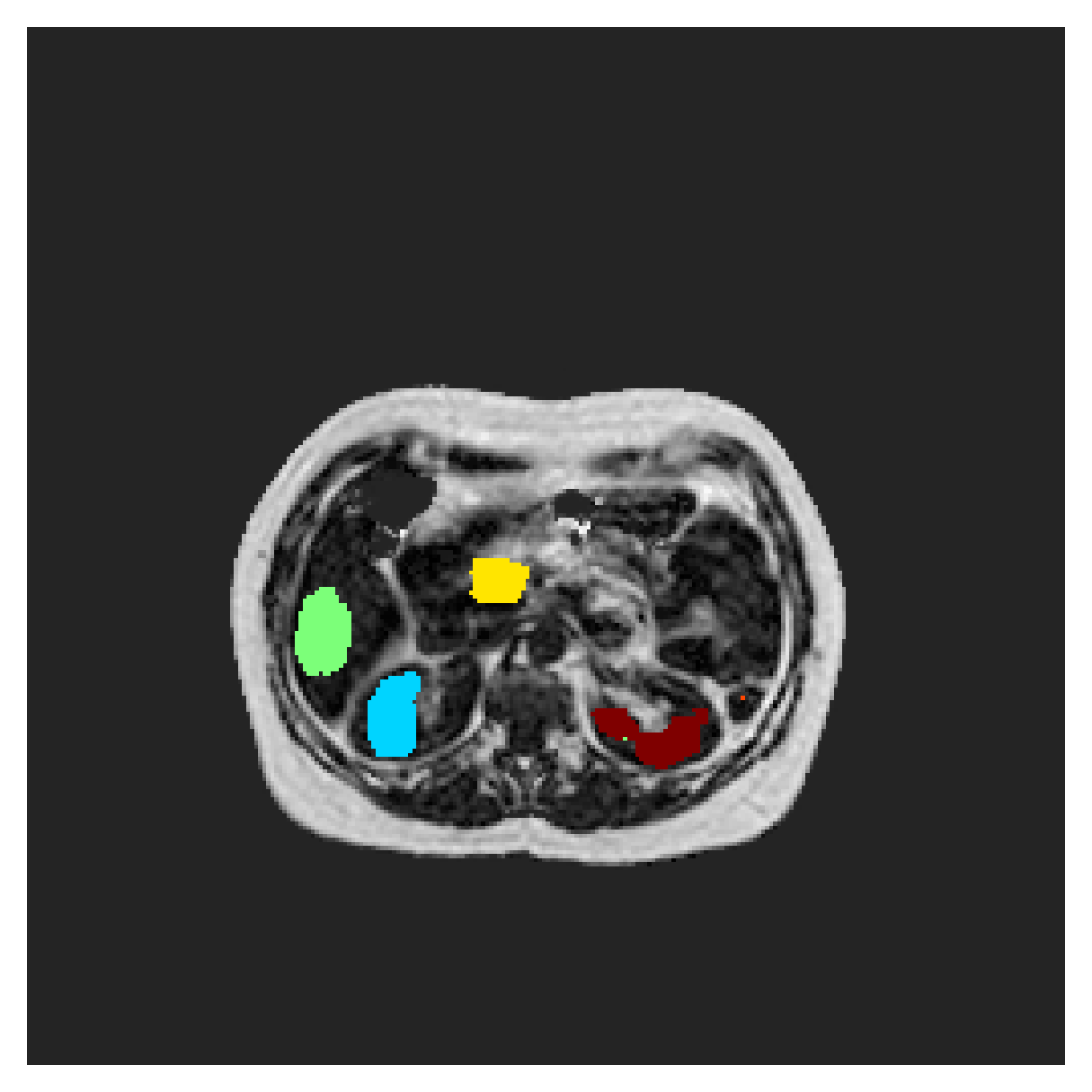}\\
                                \includegraphics[width=\textwidth]{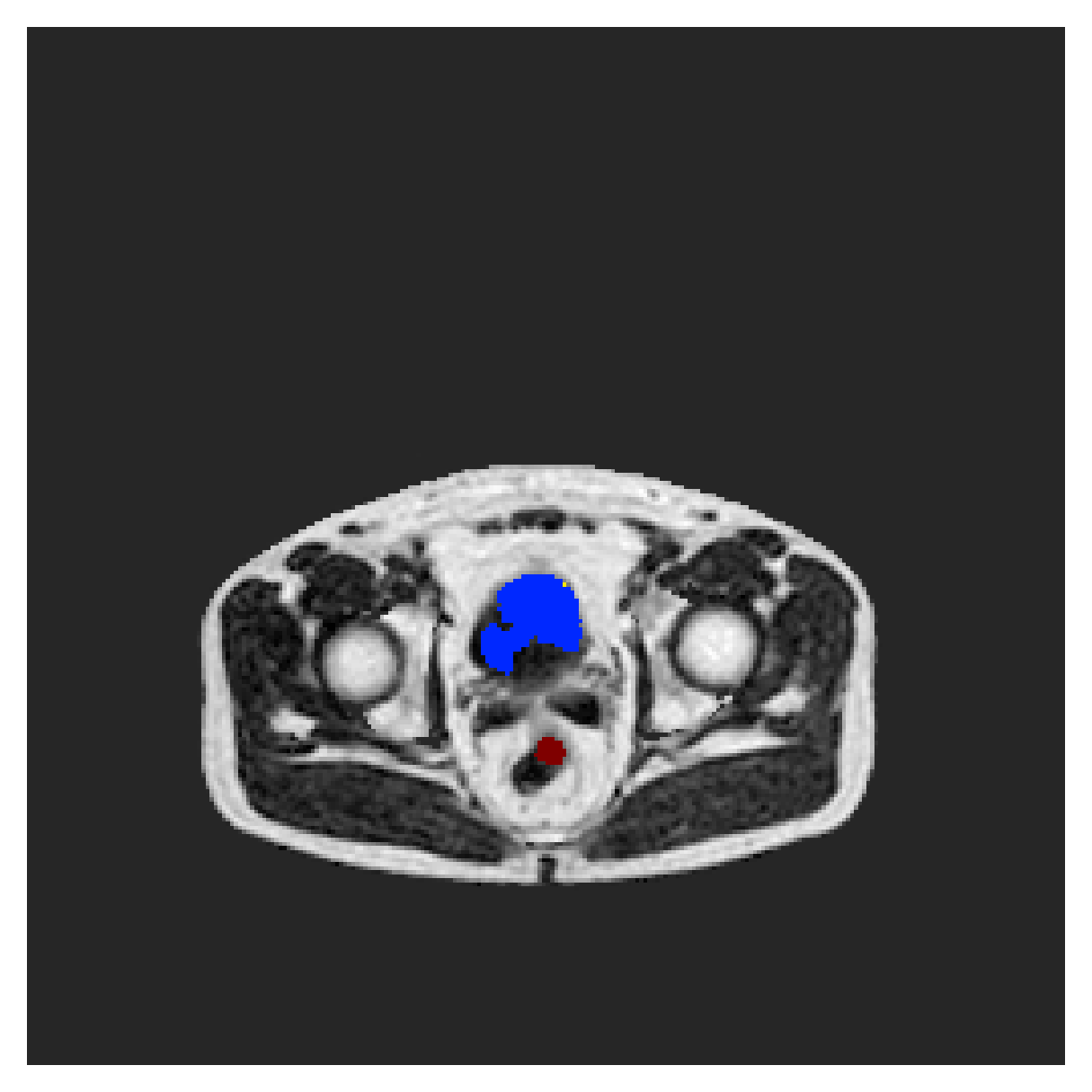}\\
                                \includegraphics[width=\textwidth]{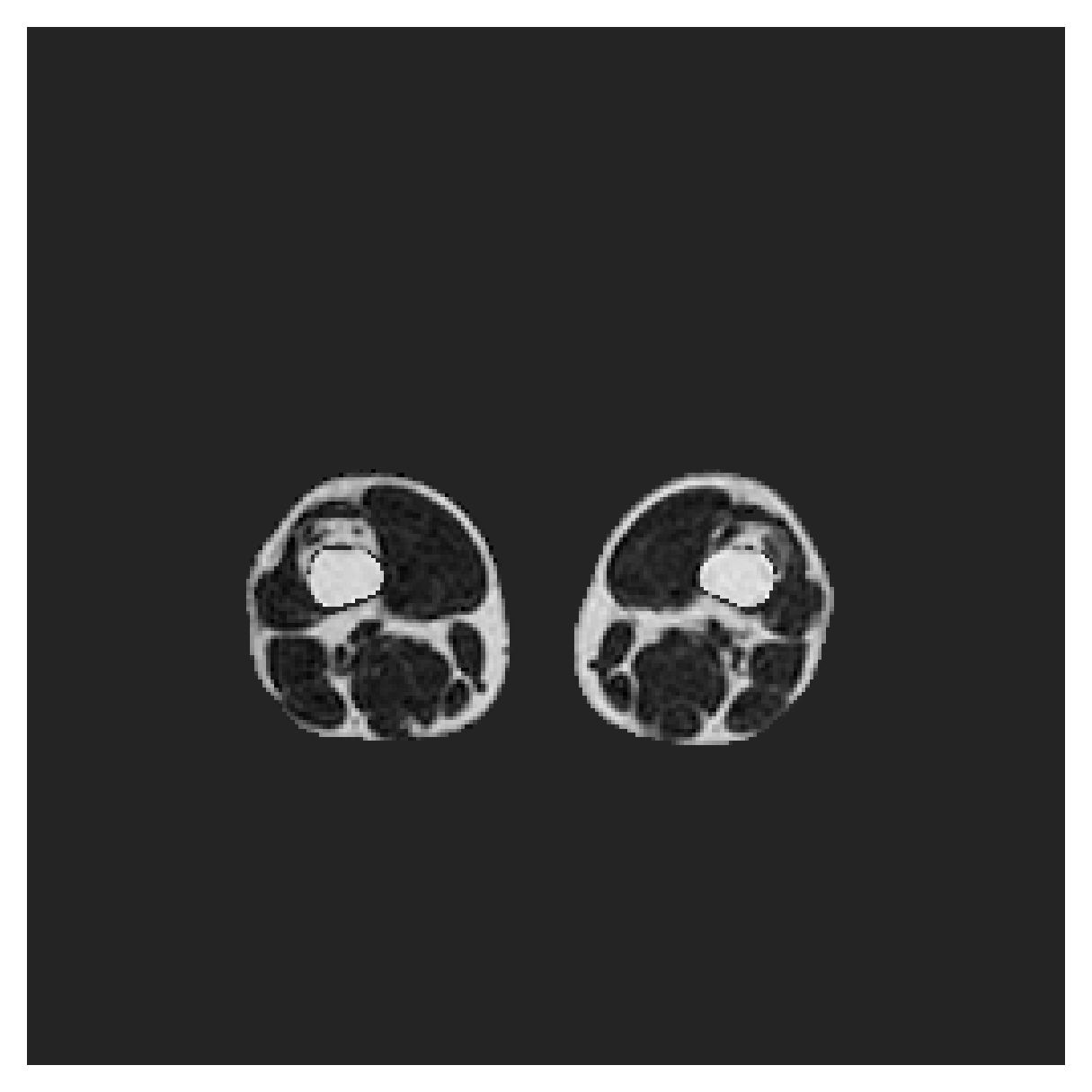}\\
                                \includegraphics[width=\textwidth]{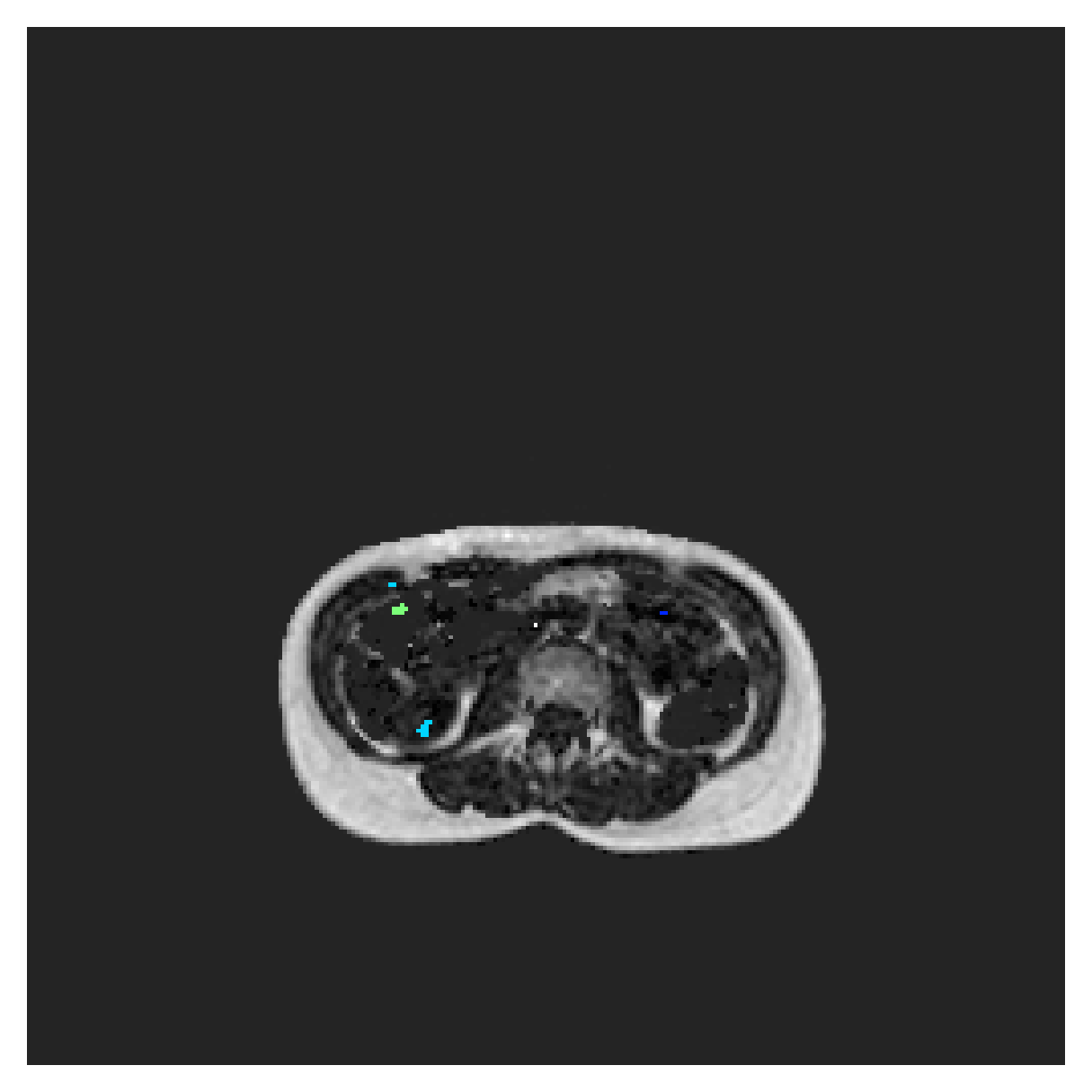}
                                \caption{$\mathcal L_{\widetilde{\text{CE}}} + \mathcal{L}_B^{geo}$\\ $\qquad$}
                                \label{subfig:geo1b}
                        \end{subfigure}
                        \begin{subfigure}[b]{0.11\textwidth}
                                \centering
                                \includegraphics[width=\textwidth]{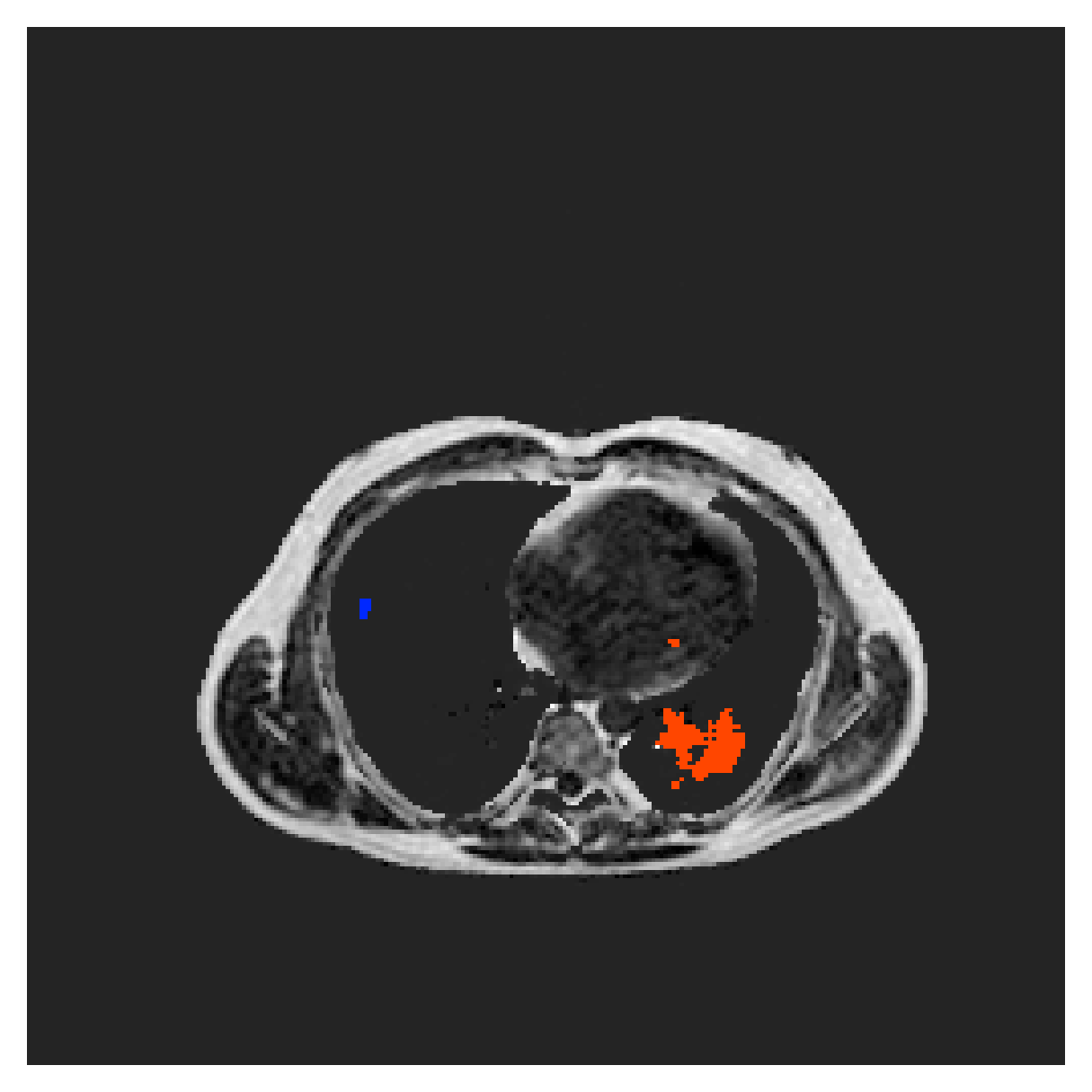}\\
                                \includegraphics[width=\textwidth]{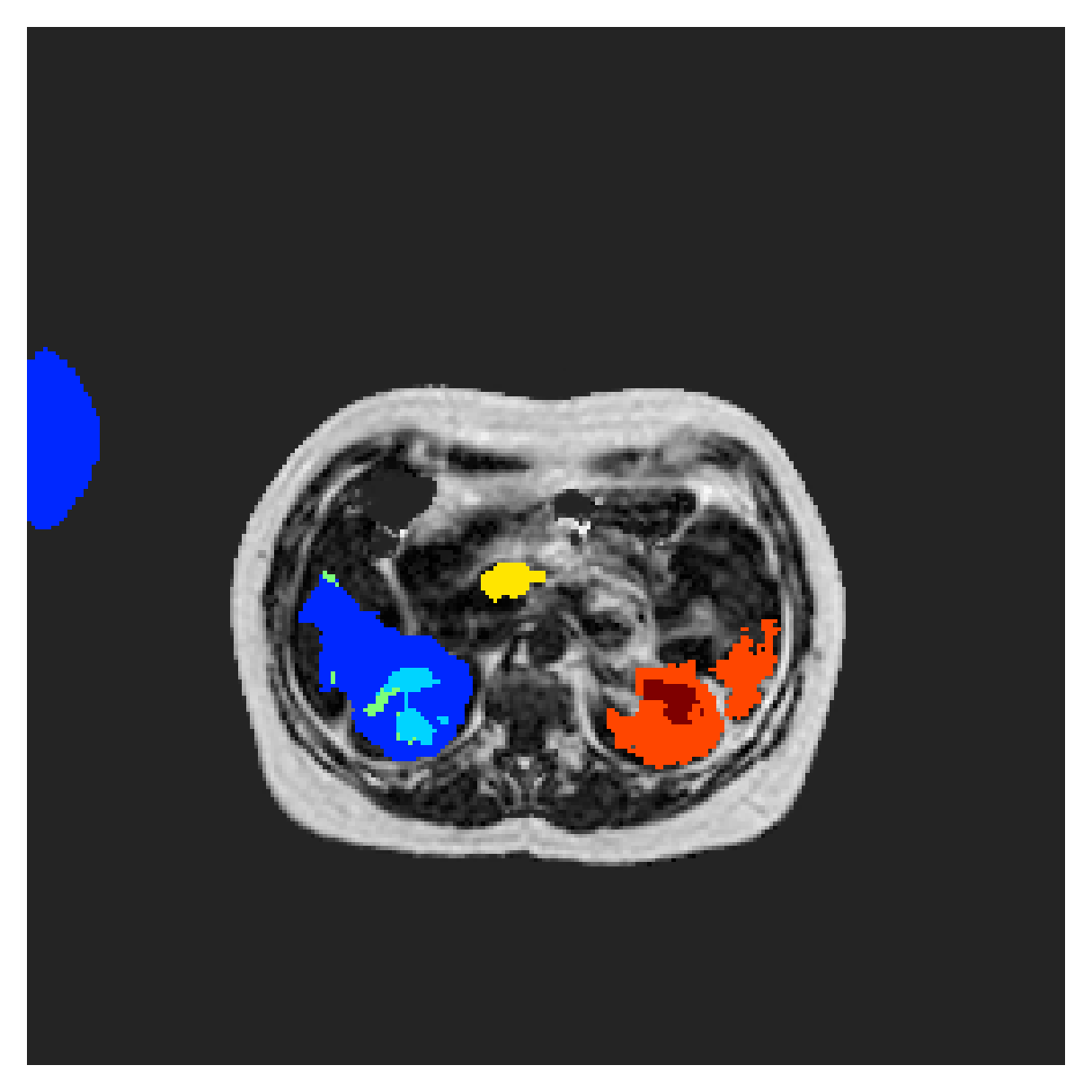}\\
                                \includegraphics[width=\textwidth]{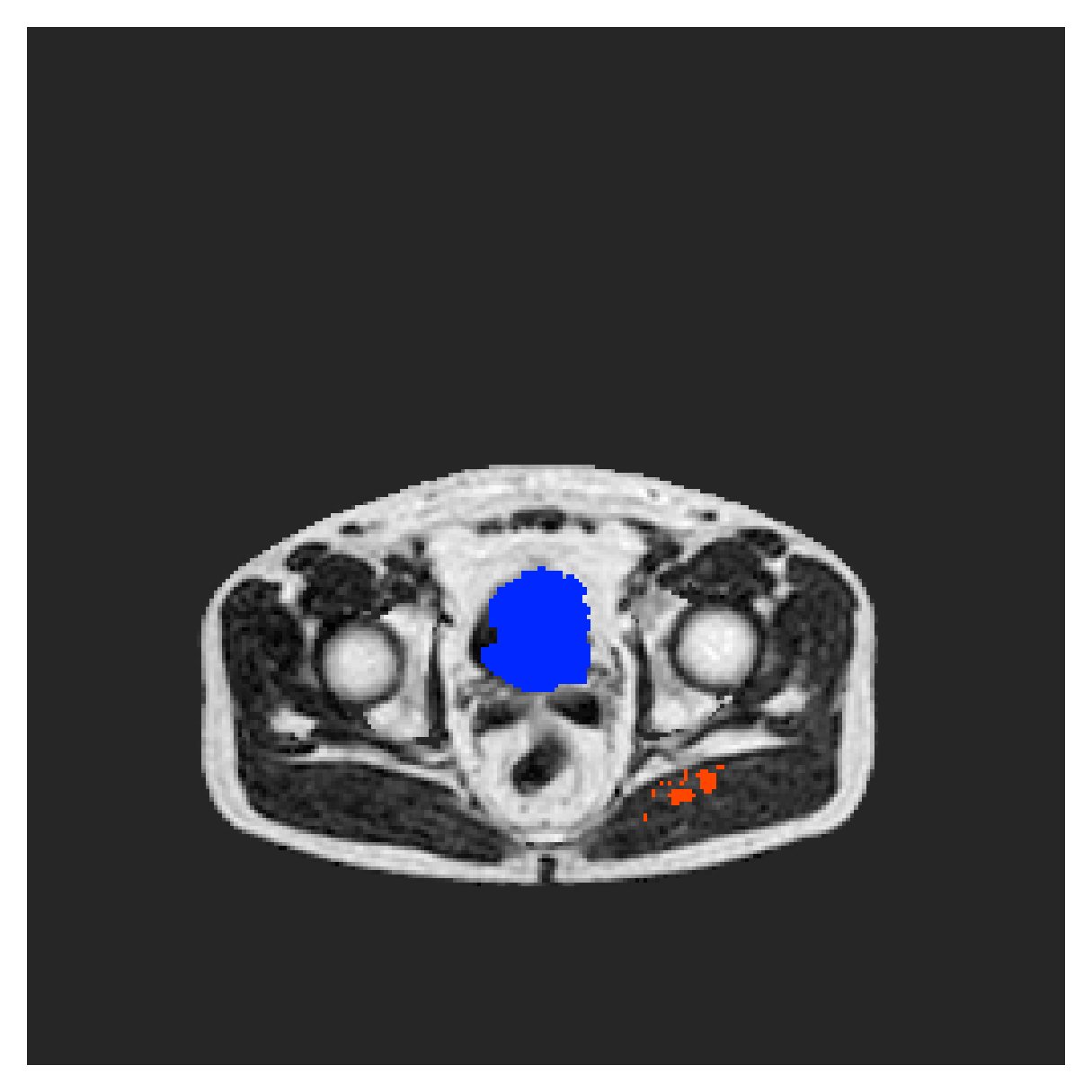}\\
                                \includegraphics[width=\textwidth]{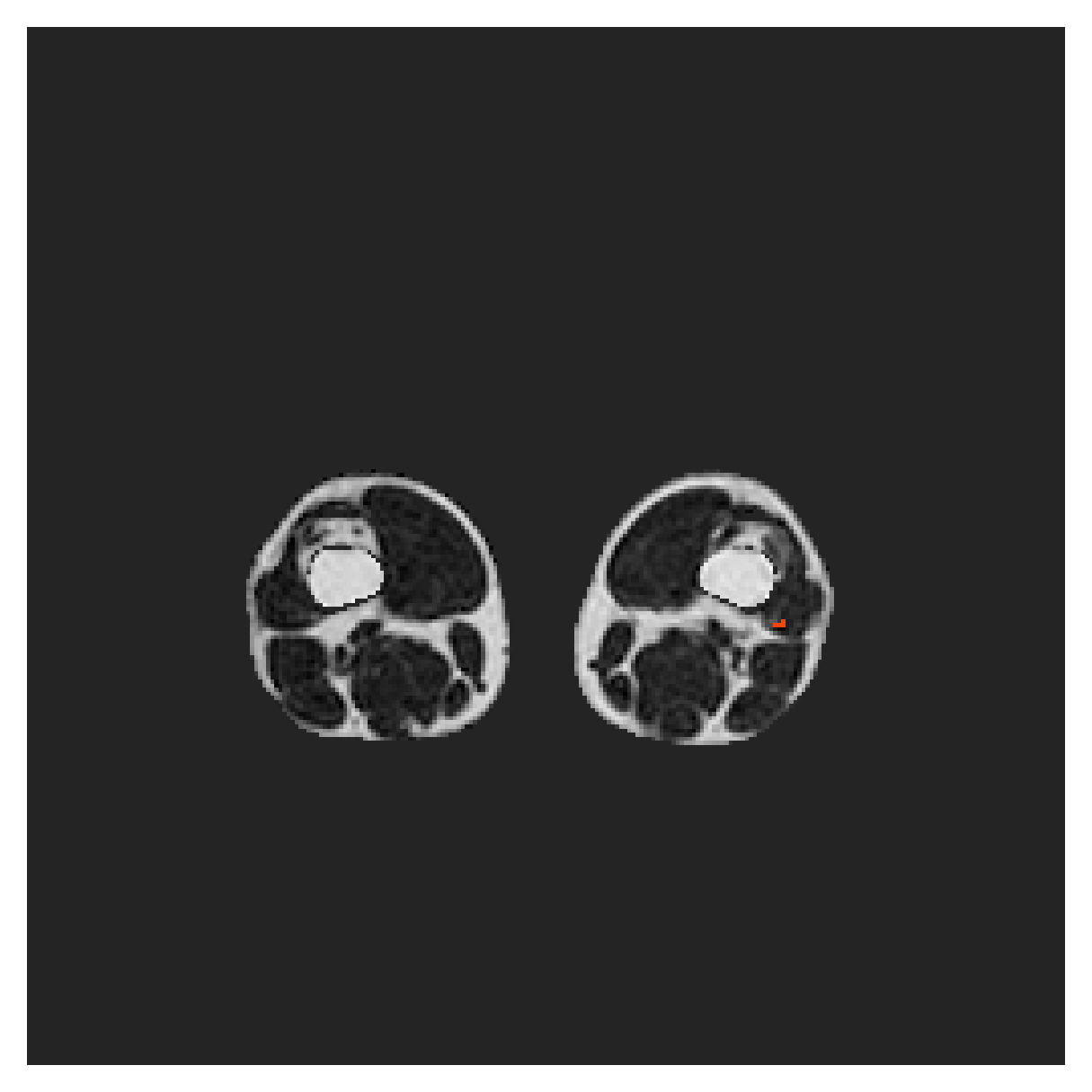}\\
                                \includegraphics[width=\textwidth]{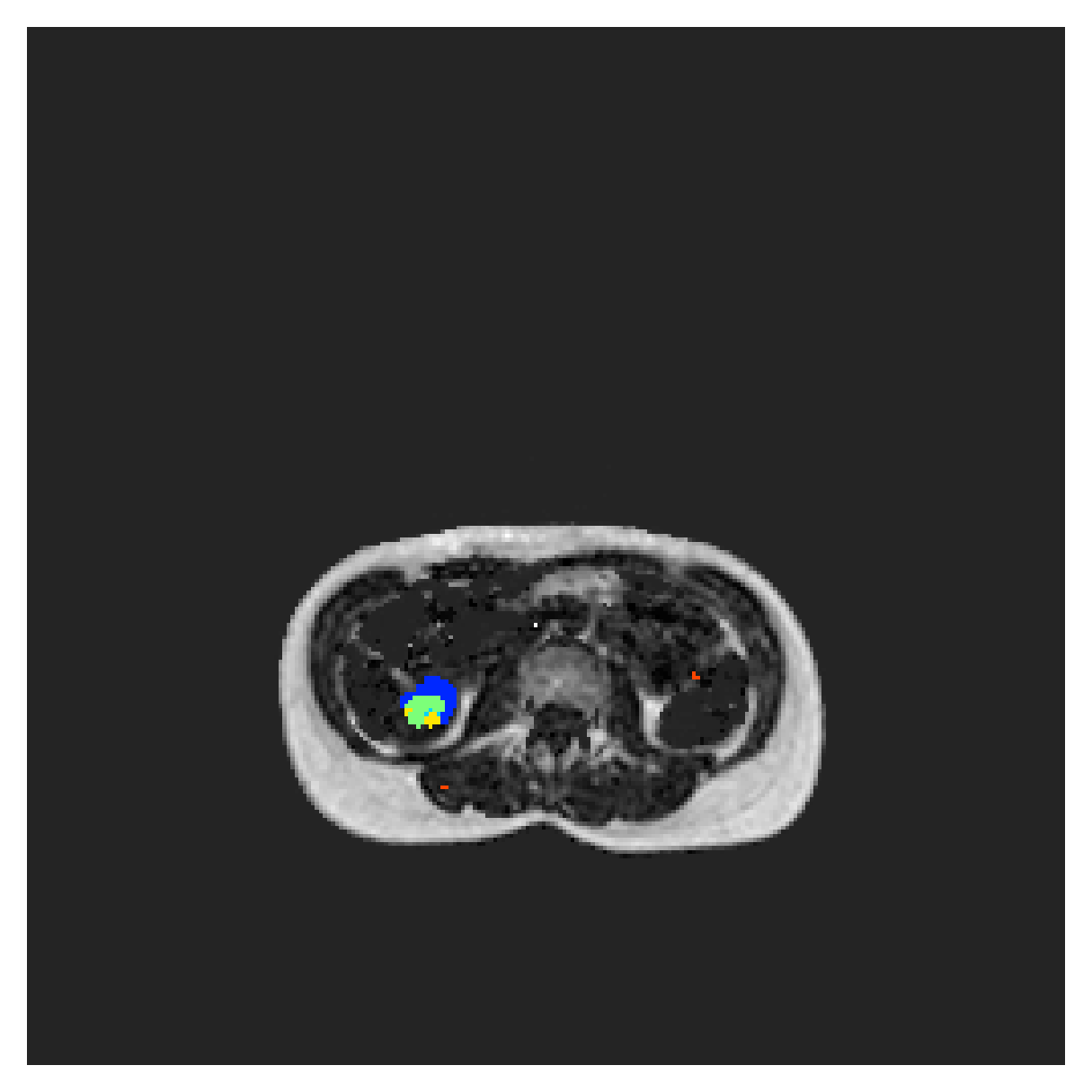}
                                \caption{$\mathcal L_{\widetilde{\text{CE}}} + \mathcal{L}_B^{int}$\\ $\qquad$}
                                \label{subfig:int1b}
                        \end{subfigure}
                        \begin{subfigure}[b]{0.11\textwidth}
                                \centering
                                \includegraphics[width=\textwidth]{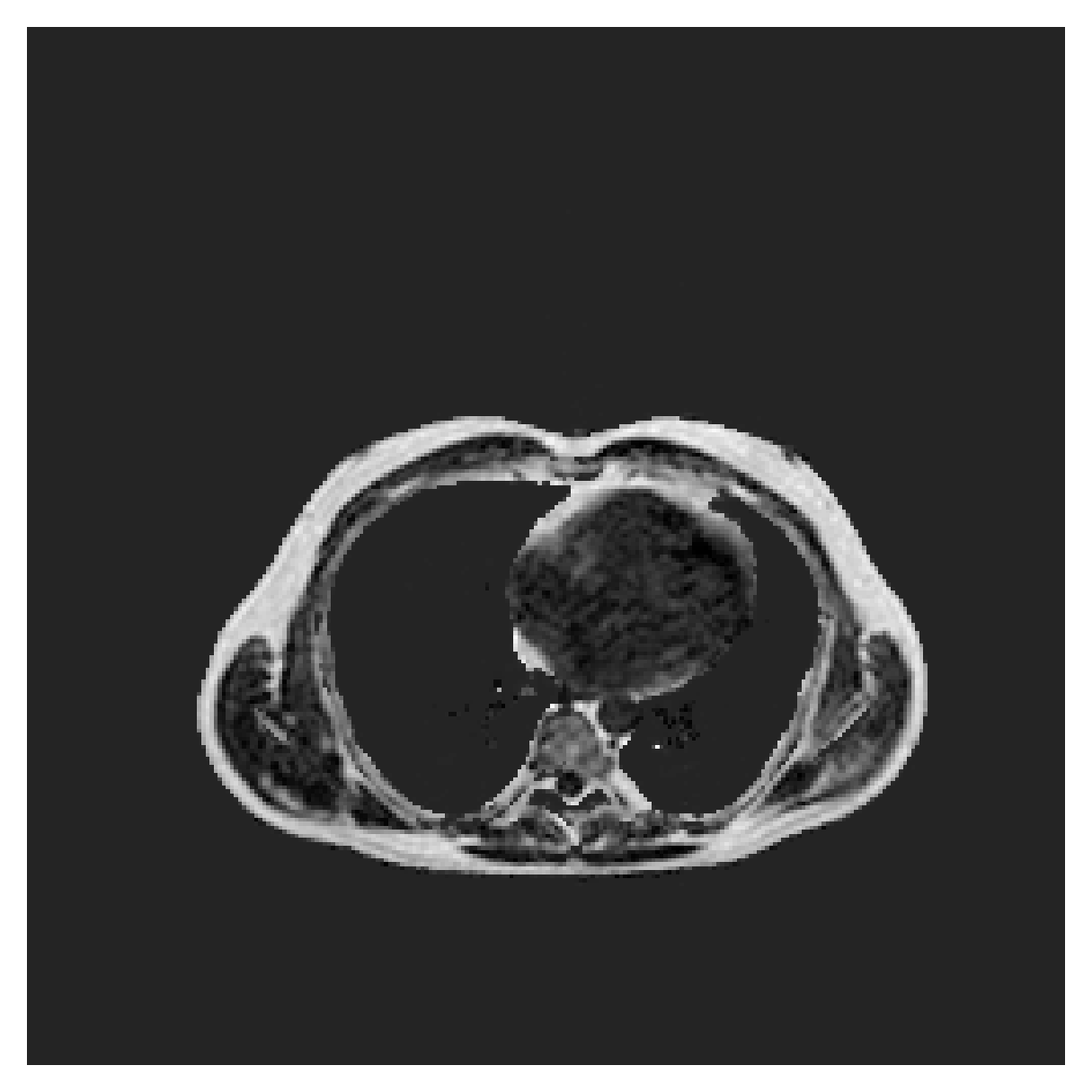}\\
                                \includegraphics[width=\textwidth]{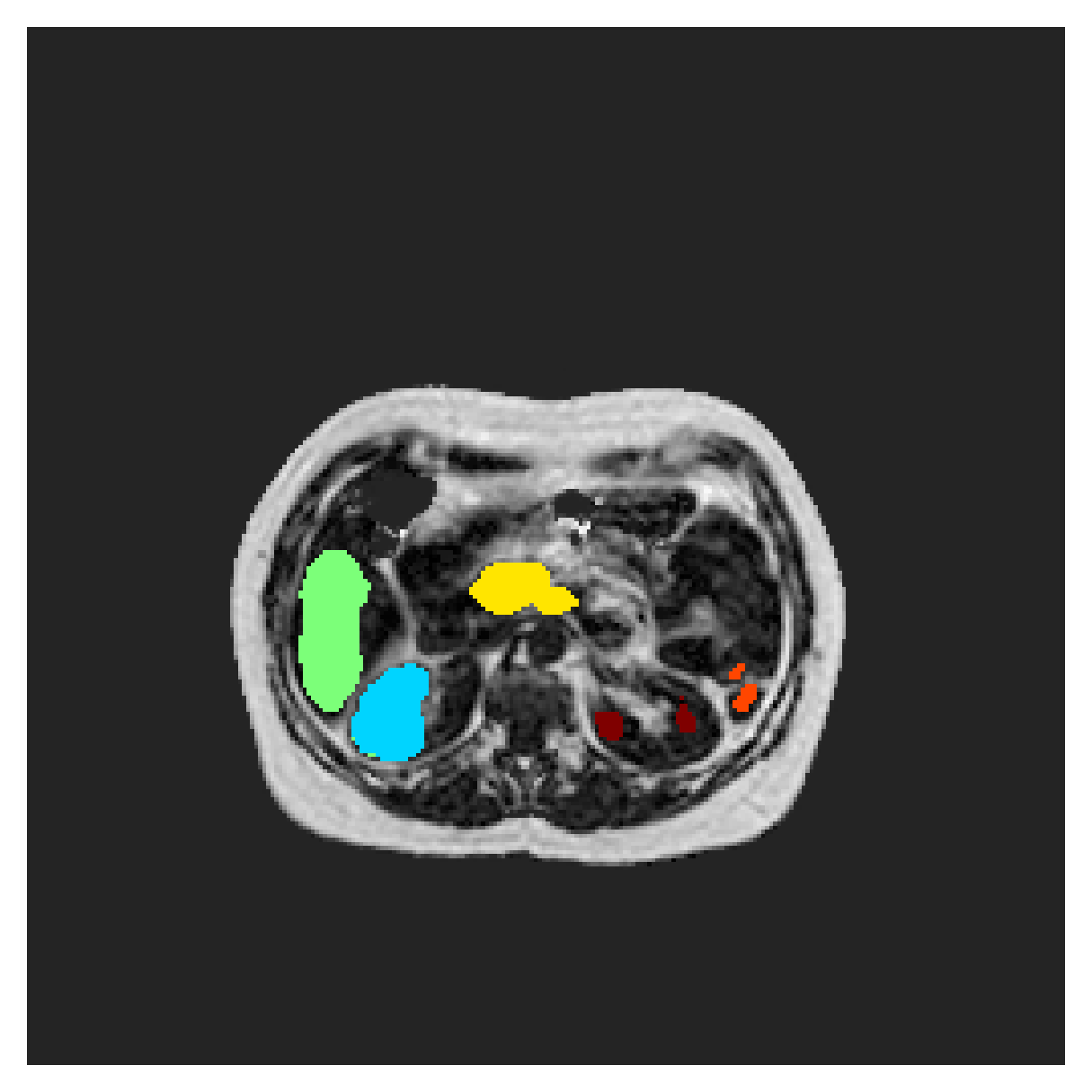}\\
                                \includegraphics[width=\textwidth]{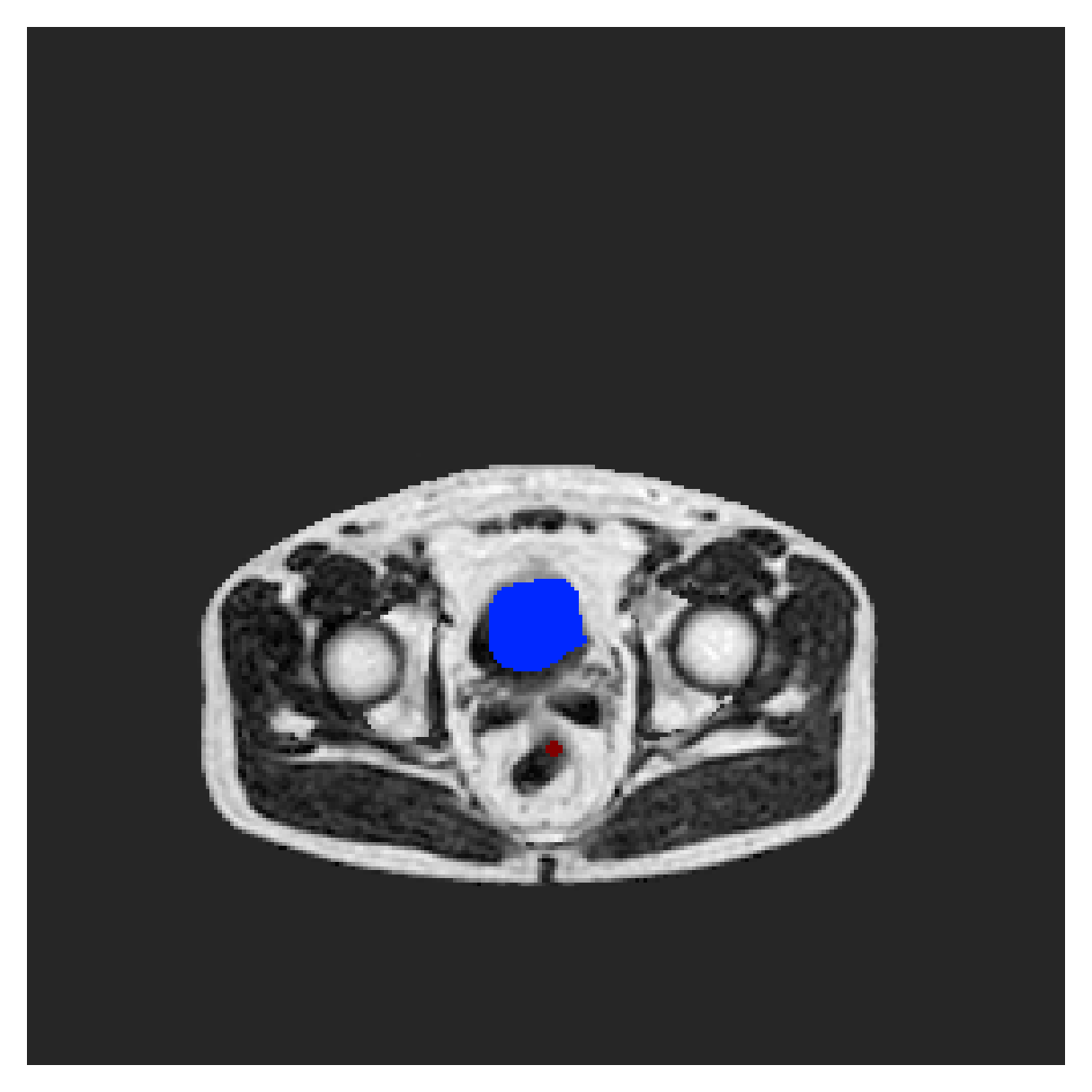}\\
                                \includegraphics[width=\textwidth]{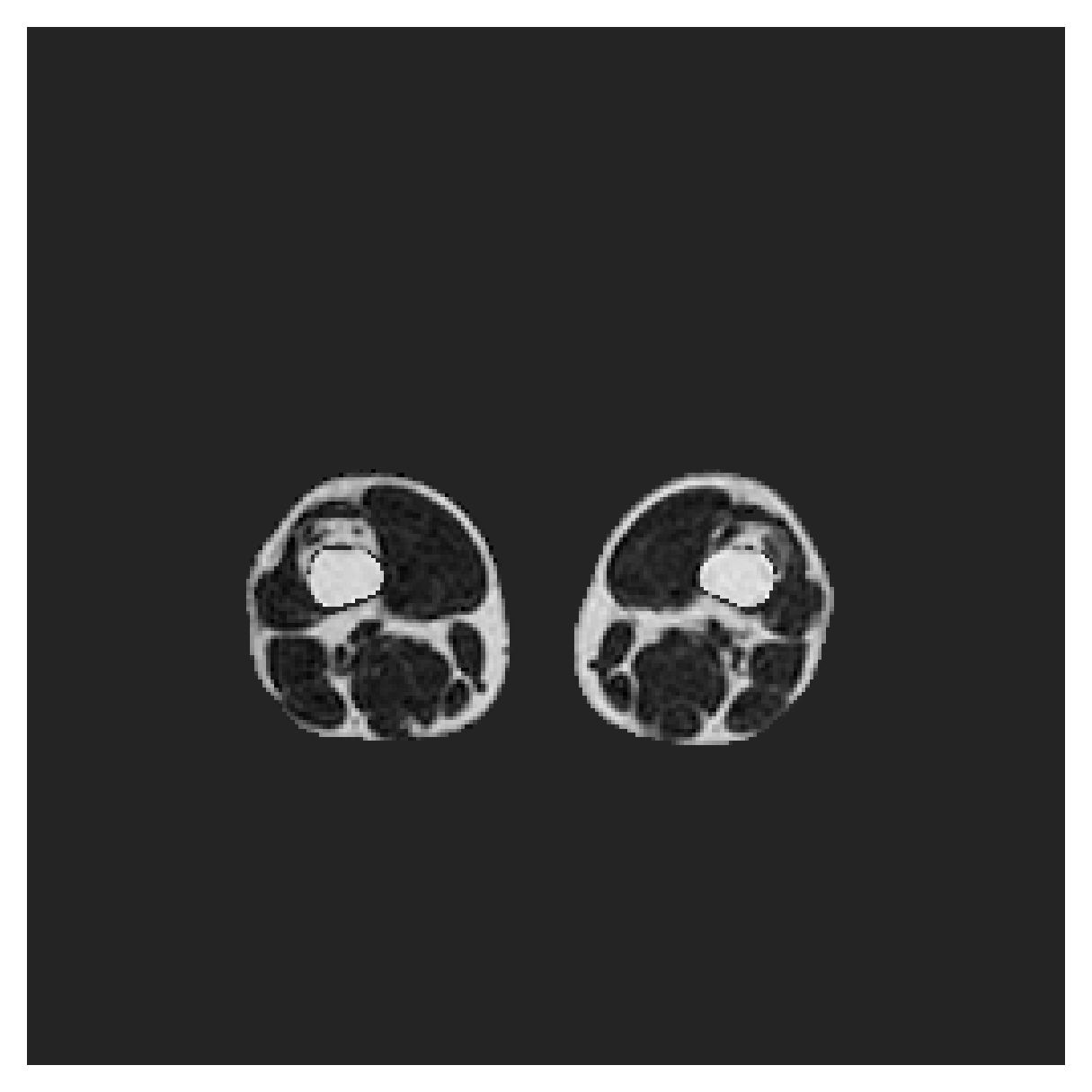}\\
                                \includegraphics[width=\textwidth]{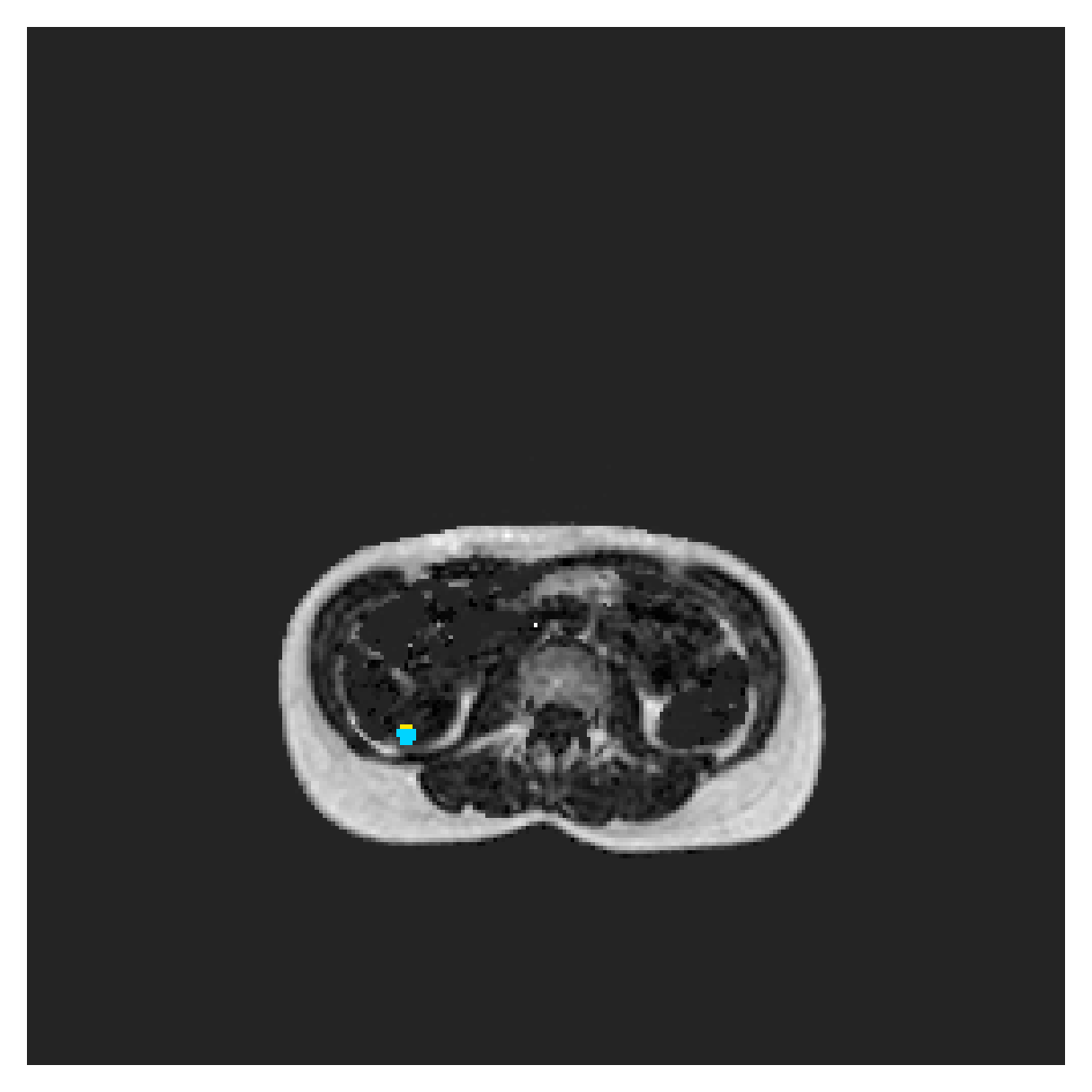}
                                \caption{$\mathcal L_{\widetilde{\text{CE}}} + \mathcal{L}_B^{mbd}$\\ $\qquad$}
                                \label{subfig:mbd1b}
                        \end{subfigure}
                        \begin{subfigure}[b]{0.11\textwidth}
                                \centering
                                \includegraphics[width=\textwidth]{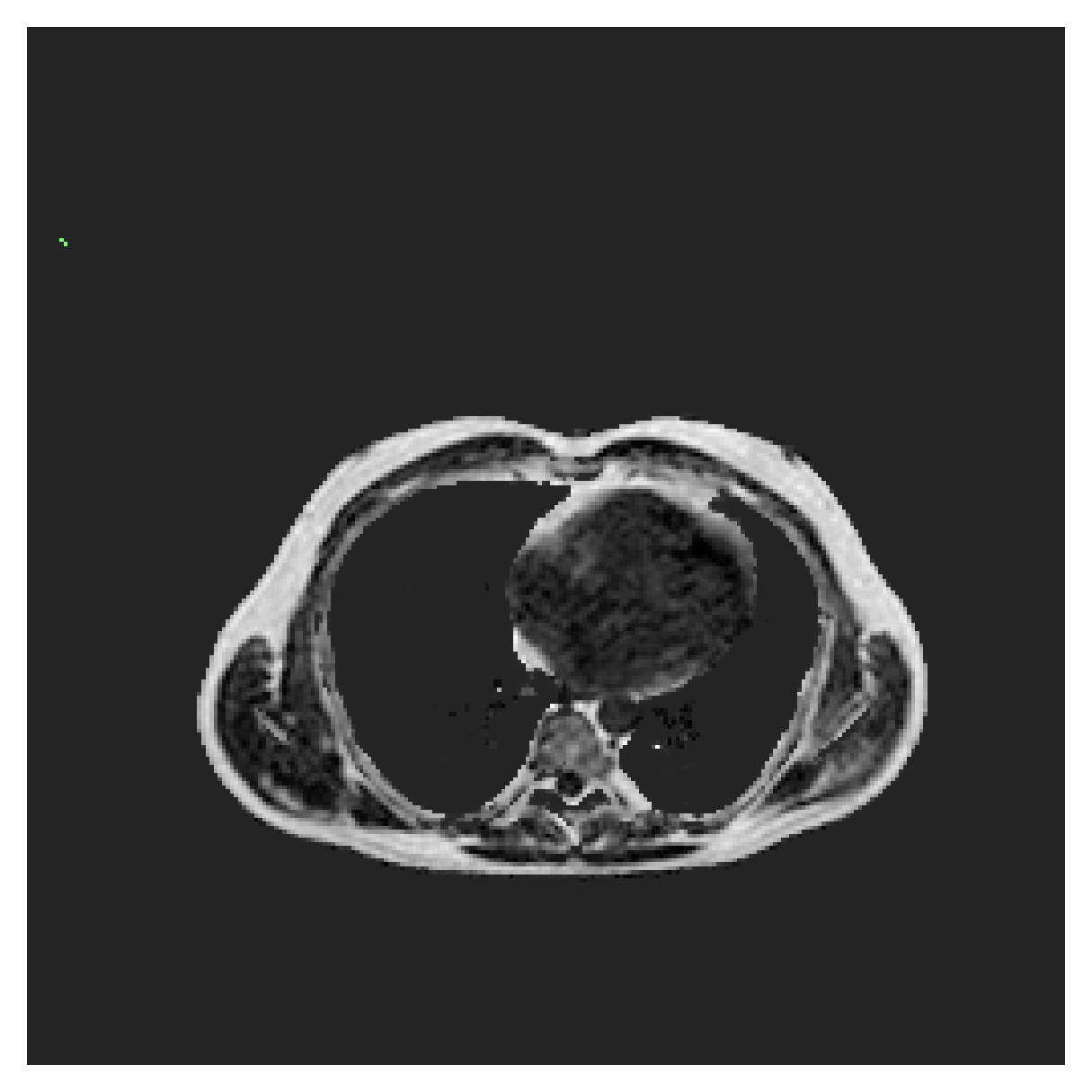}\\
                                \includegraphics[width=\textwidth]{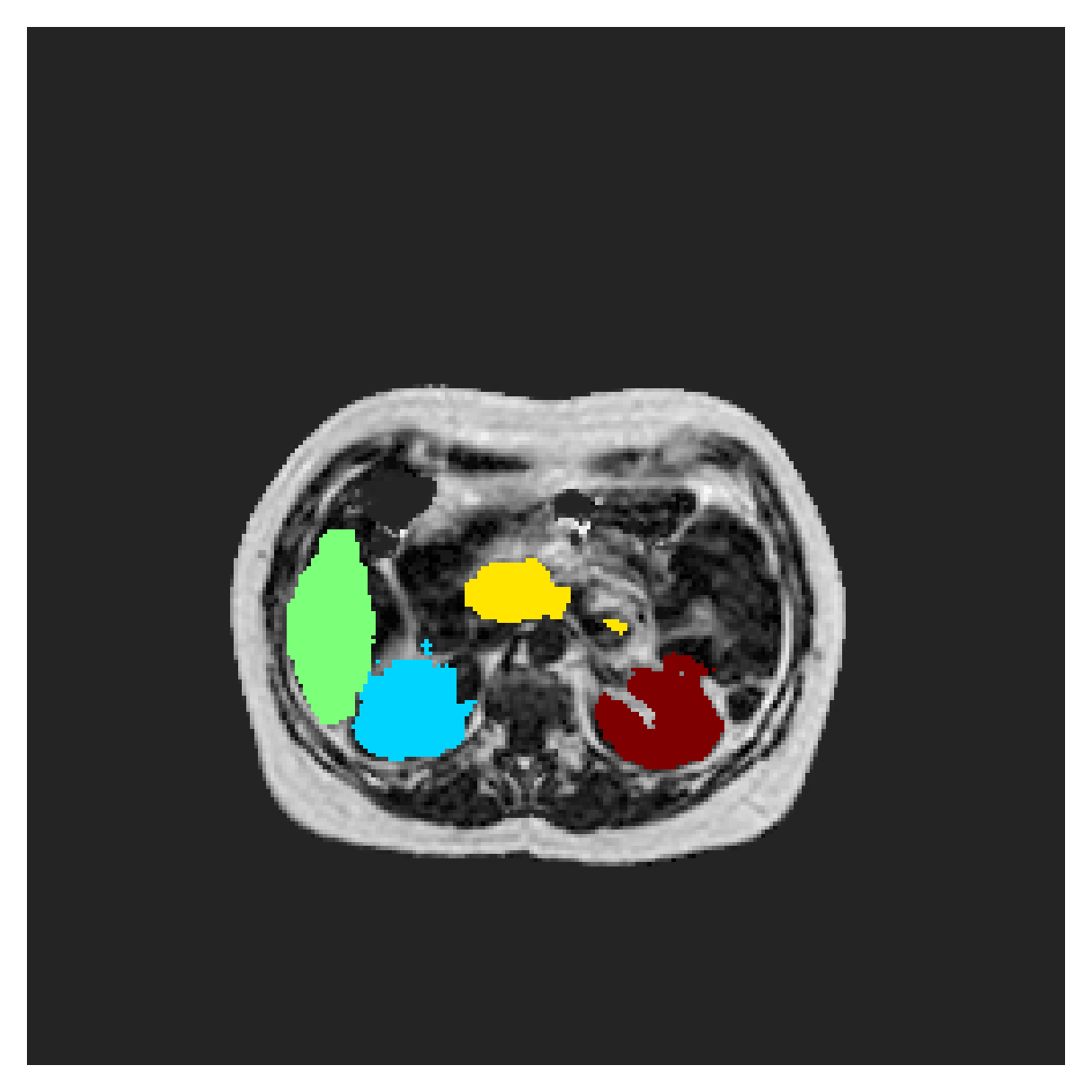}\\
                                \includegraphics[width=\textwidth]{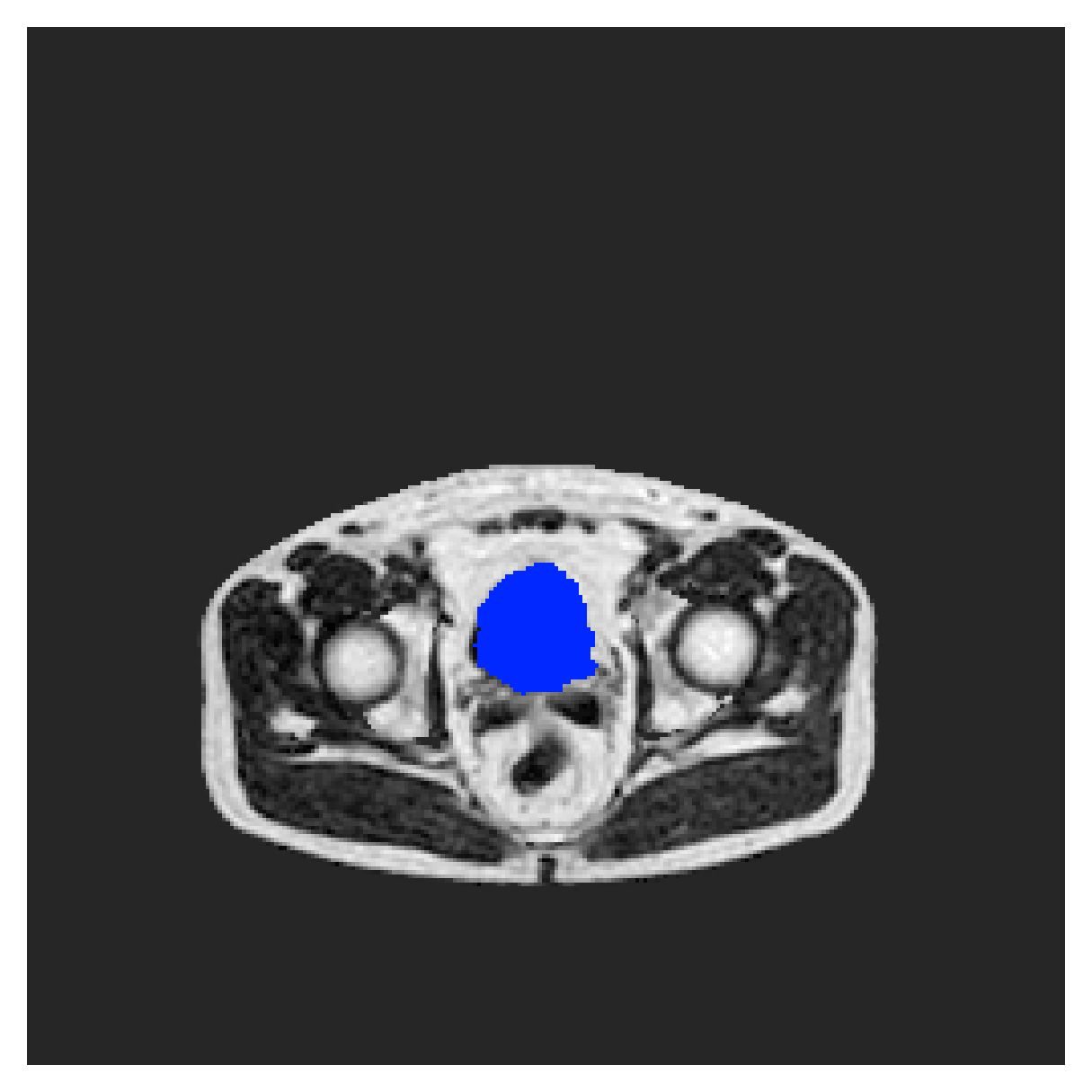}\\
                                \includegraphics[width=\textwidth]{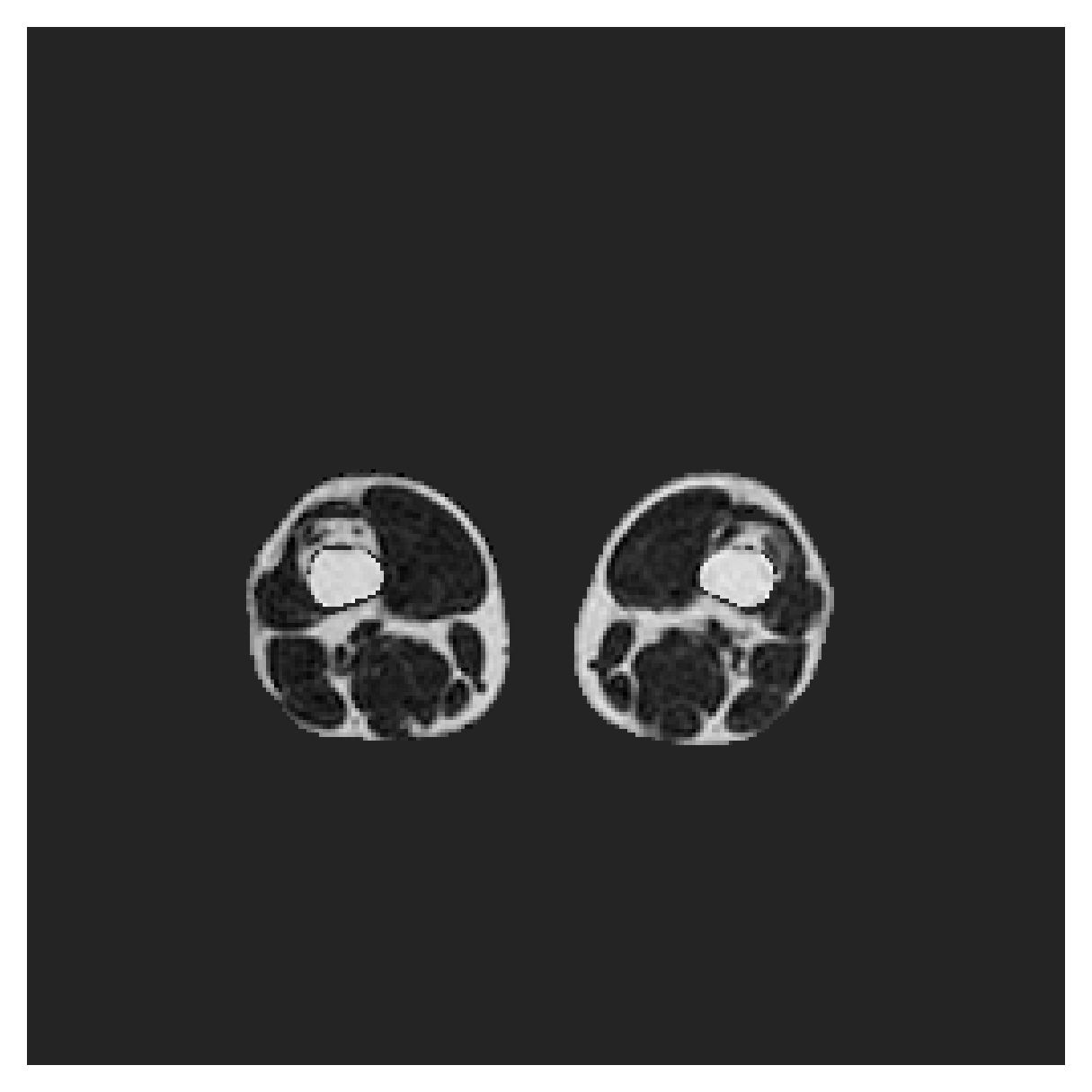}\\
                                \includegraphics[width=\textwidth]{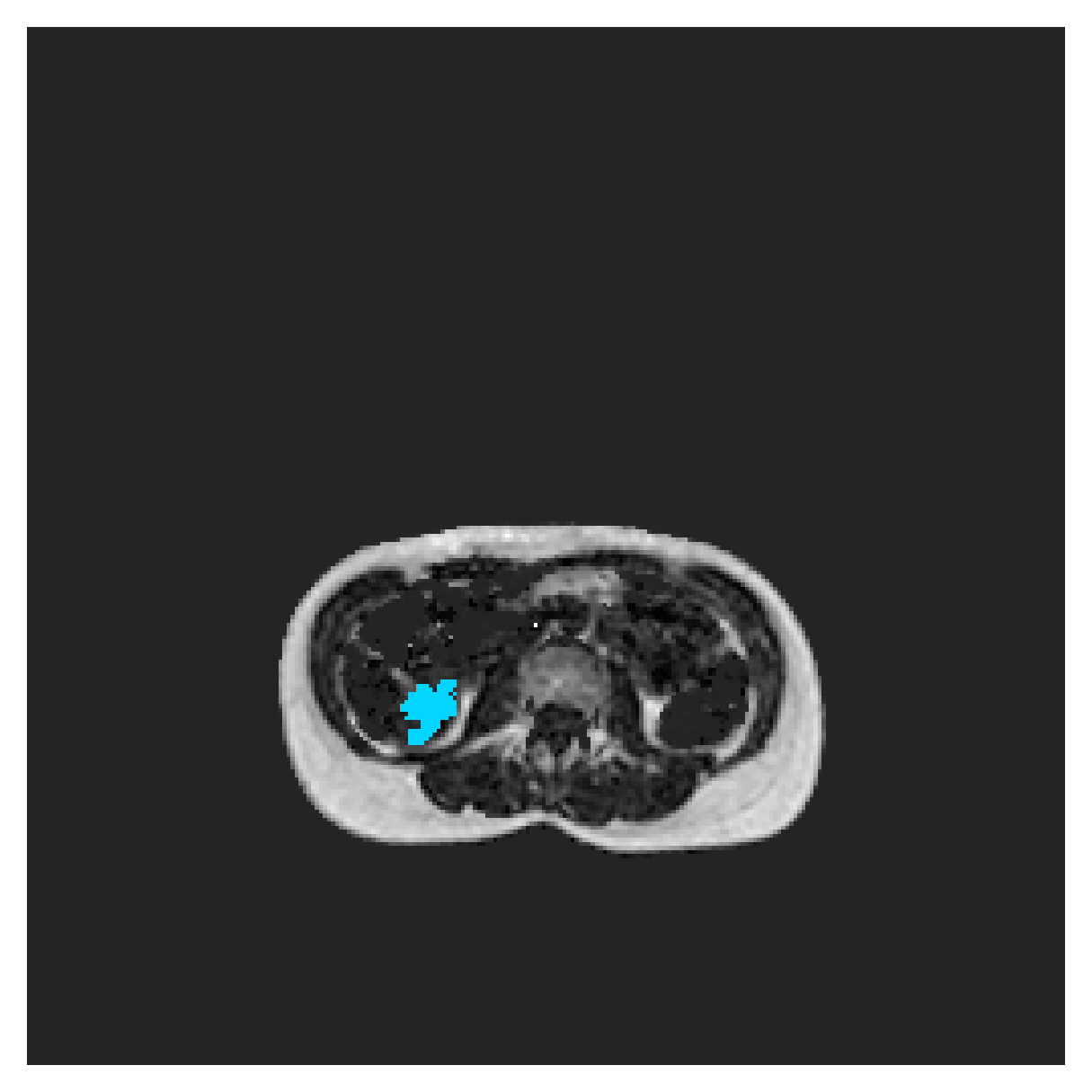}
                                \caption{$\mathcal L_{\widetilde{\text{CE}}}+$CRF-loss
                                }
                                \label{subfig:crf1b}
                        \end{subfigure}
                        \begin{subfigure}[b]{0.045\textwidth}
                                \centering
                                \includegraphics[width=\textwidth]{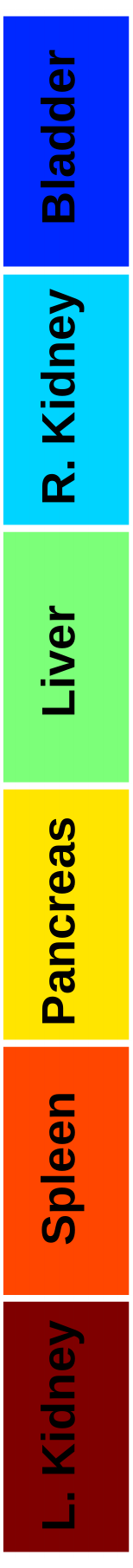}\\
                                \caption*{}
                                
                                \caption*{}
                                
                        \end{subfigure}
                        \caption{Example segmentations on the POEM test set. Training with $\mathcal{L}_\text{B}$ was done using distance maps, computed in 2D. }
                        \label{fig:POEMout}
                \end{sidewaysfigure}

%% file: examplesPOEM2.tex
\begin{figure}
                        \centering
                        \begin{subfigure}[b]{0.13\textwidth}
                                \centering
                                \includegraphics[width=\textwidth]{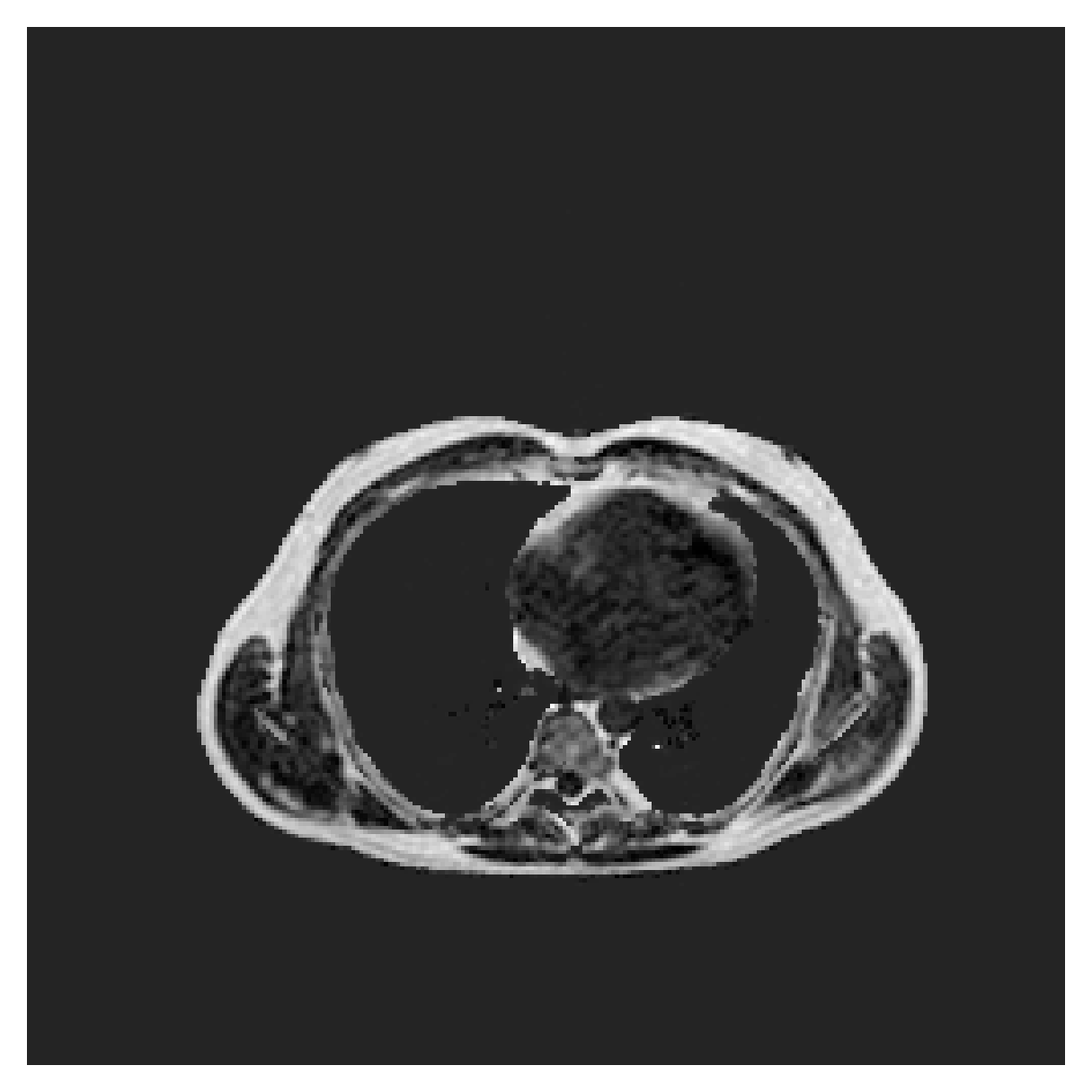}\\
                                \includegraphics[width=\textwidth]{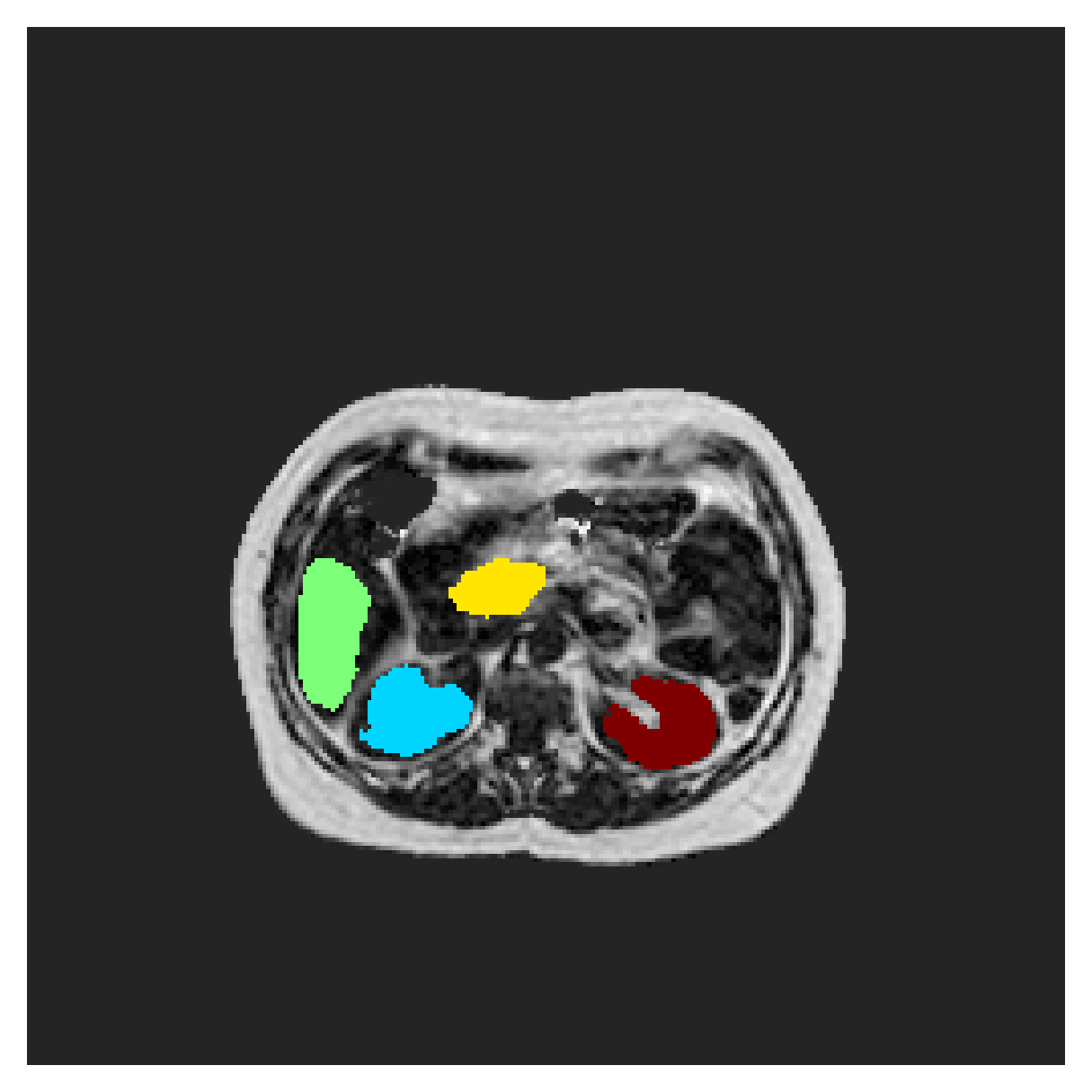}\\
                                \includegraphics[width=\textwidth]{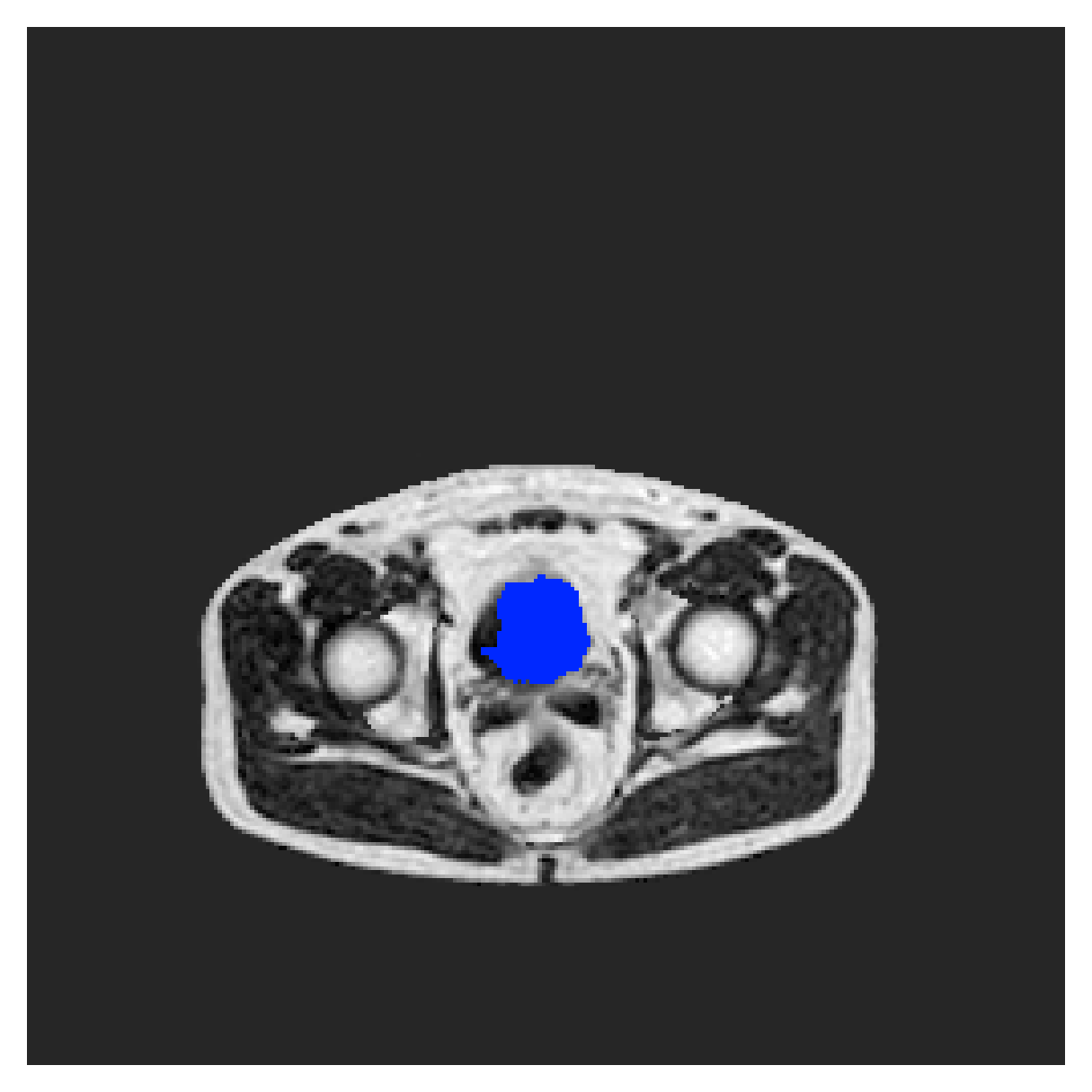}\\
                                \includegraphics[width=\textwidth]{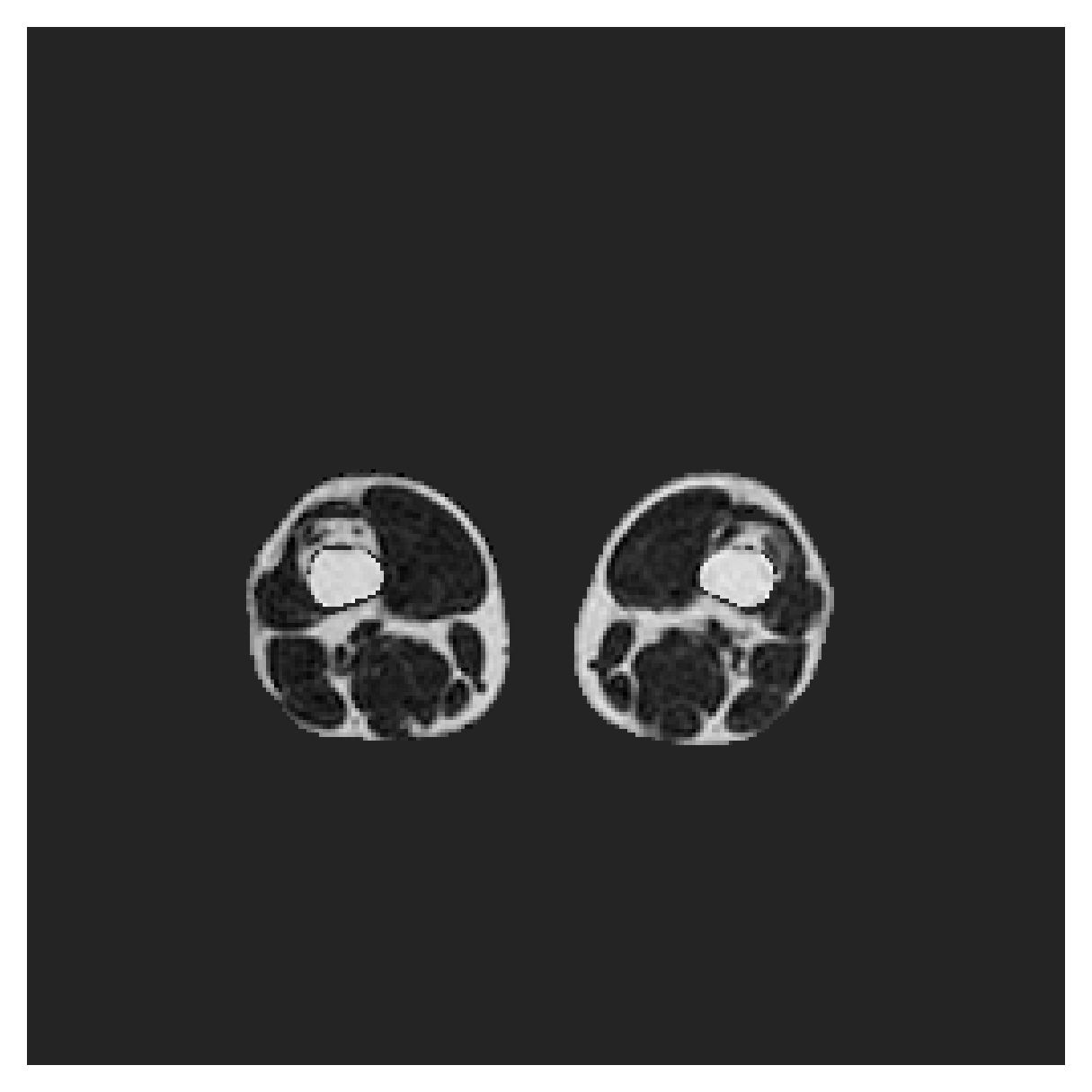}\\
                                \includegraphics[width=\textwidth]{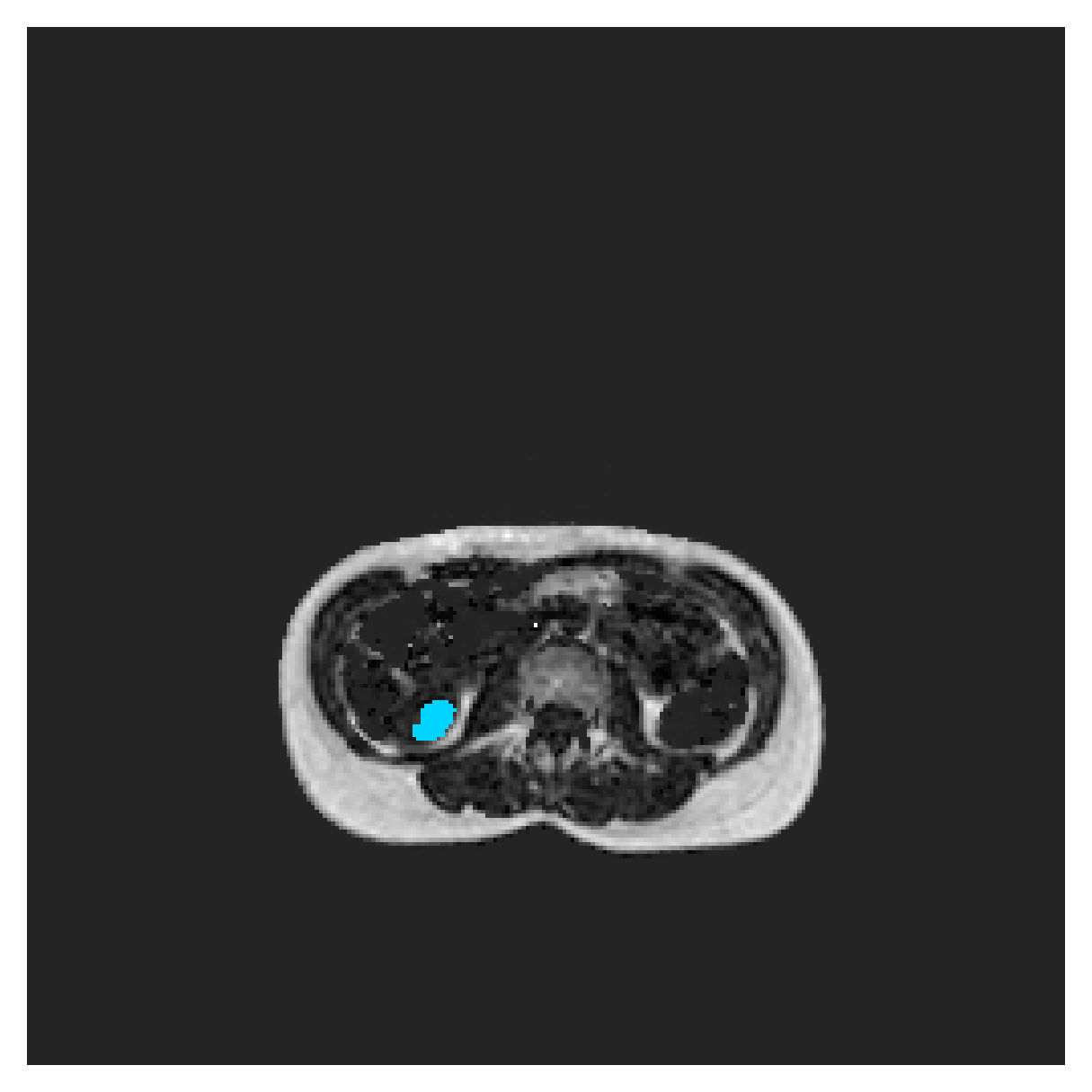}
                                \caption{$\mathcal L_{\widetilde{\text{CE}}} + \mathcal{L}_B^{euc}$\\ $\qquad$}
                                \label{subfig:euc1}
                        \end{subfigure}
                        \begin{subfigure}[b]{0.13\textwidth}
                                \centering
                                \includegraphics[width=\textwidth]{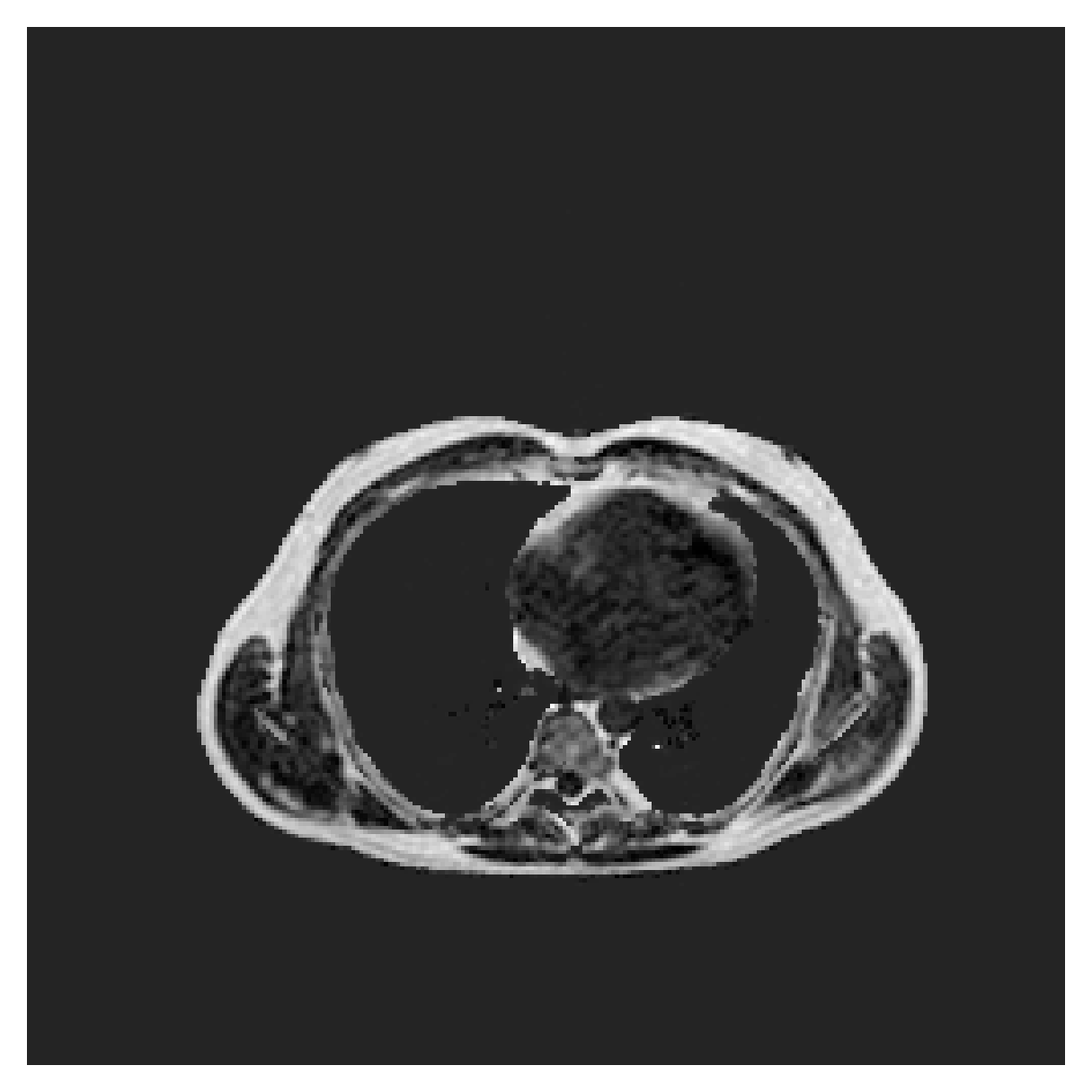}\\ 
                                \includegraphics[width=\textwidth]{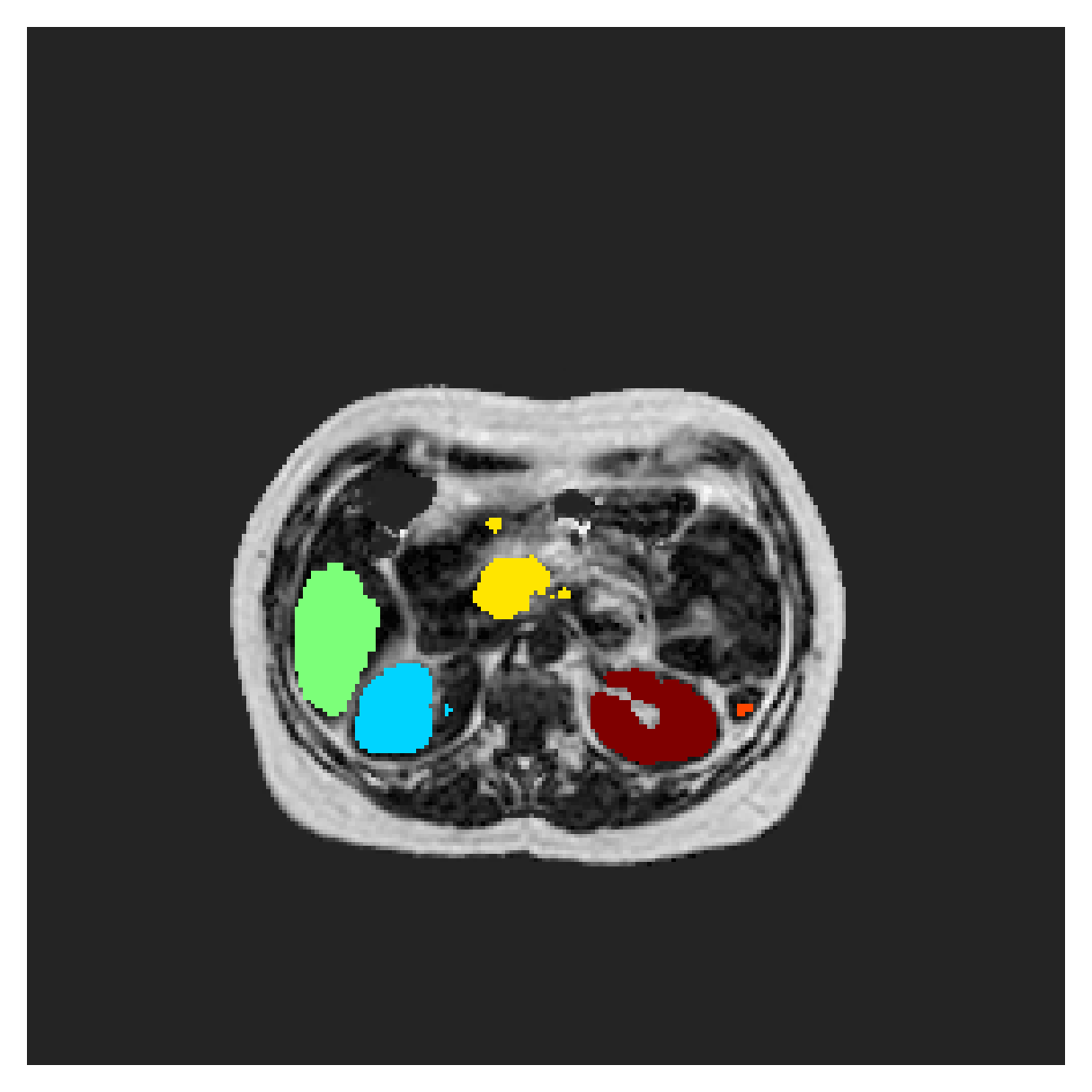}\\
                                \includegraphics[width=\textwidth]{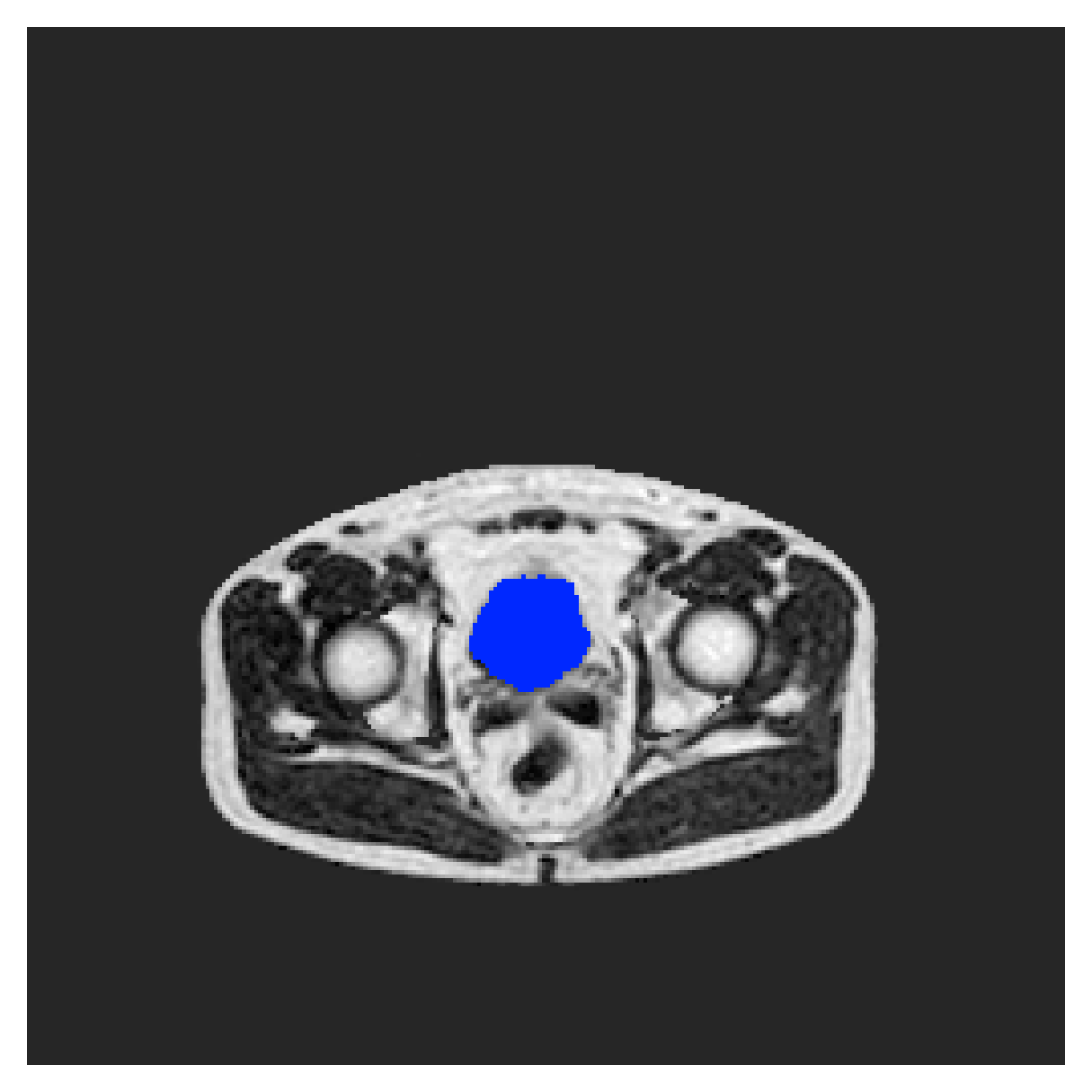}\\
                                \includegraphics[width=\textwidth]{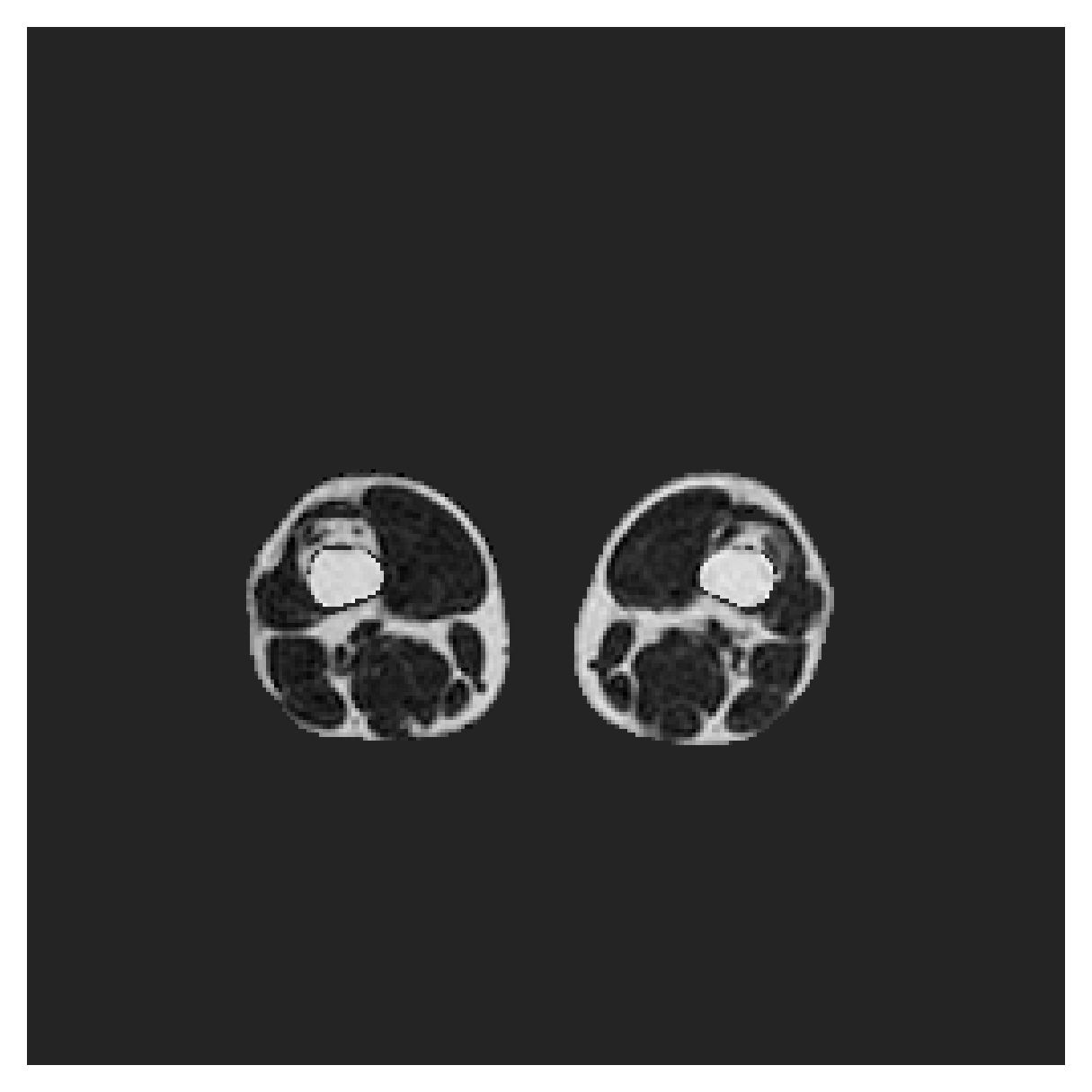}\\
                                \includegraphics[width=\textwidth]{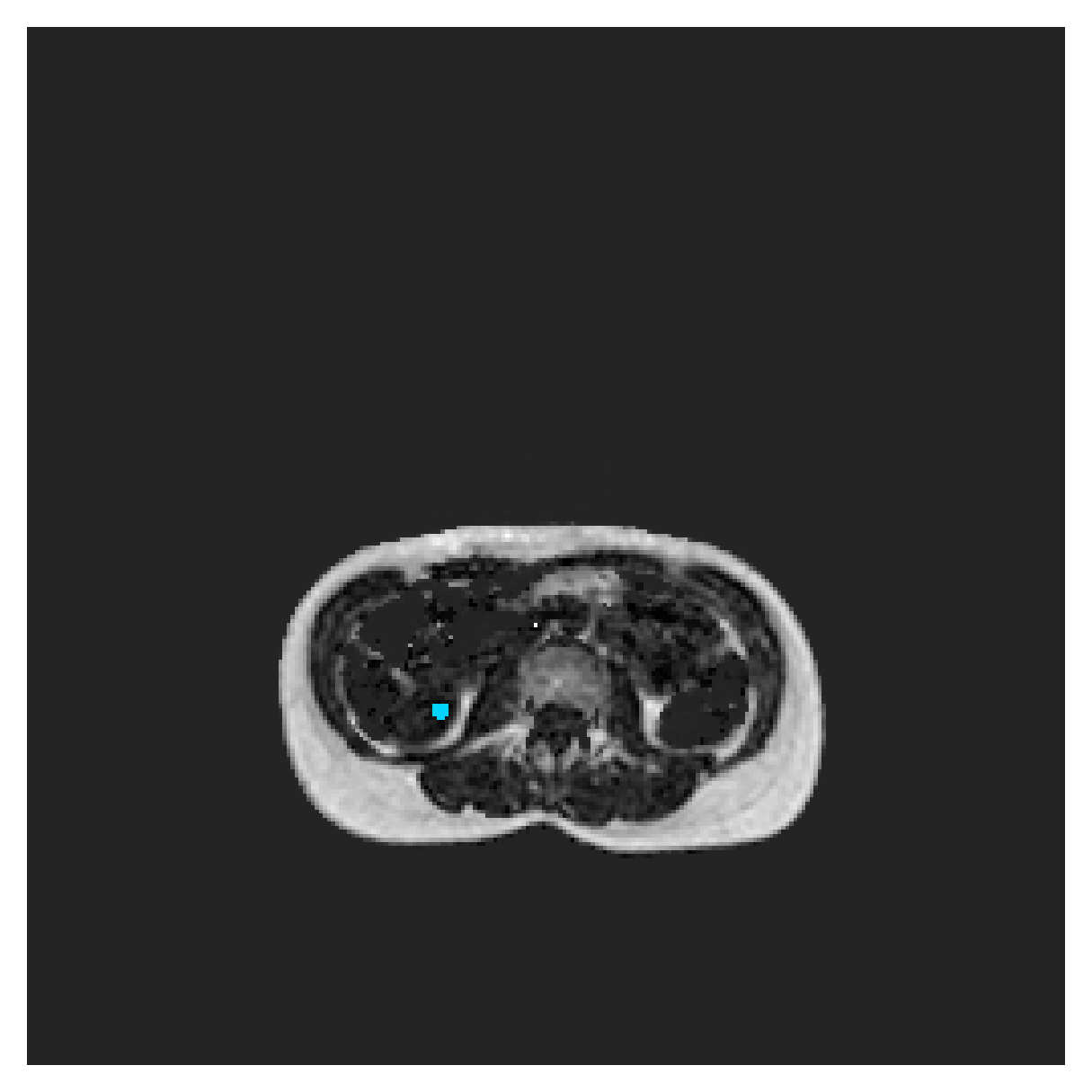}
                                \caption{$\mathcal L_{\widetilde{\text{CE}}} + \mathcal{L}_B^{geo}$\\ $\qquad$}
                                \label{subfig:geo1}
                        \end{subfigure}
                        \begin{subfigure}[b]{0.13\textwidth}
                                \centering
                                \includegraphics[width=\textwidth]{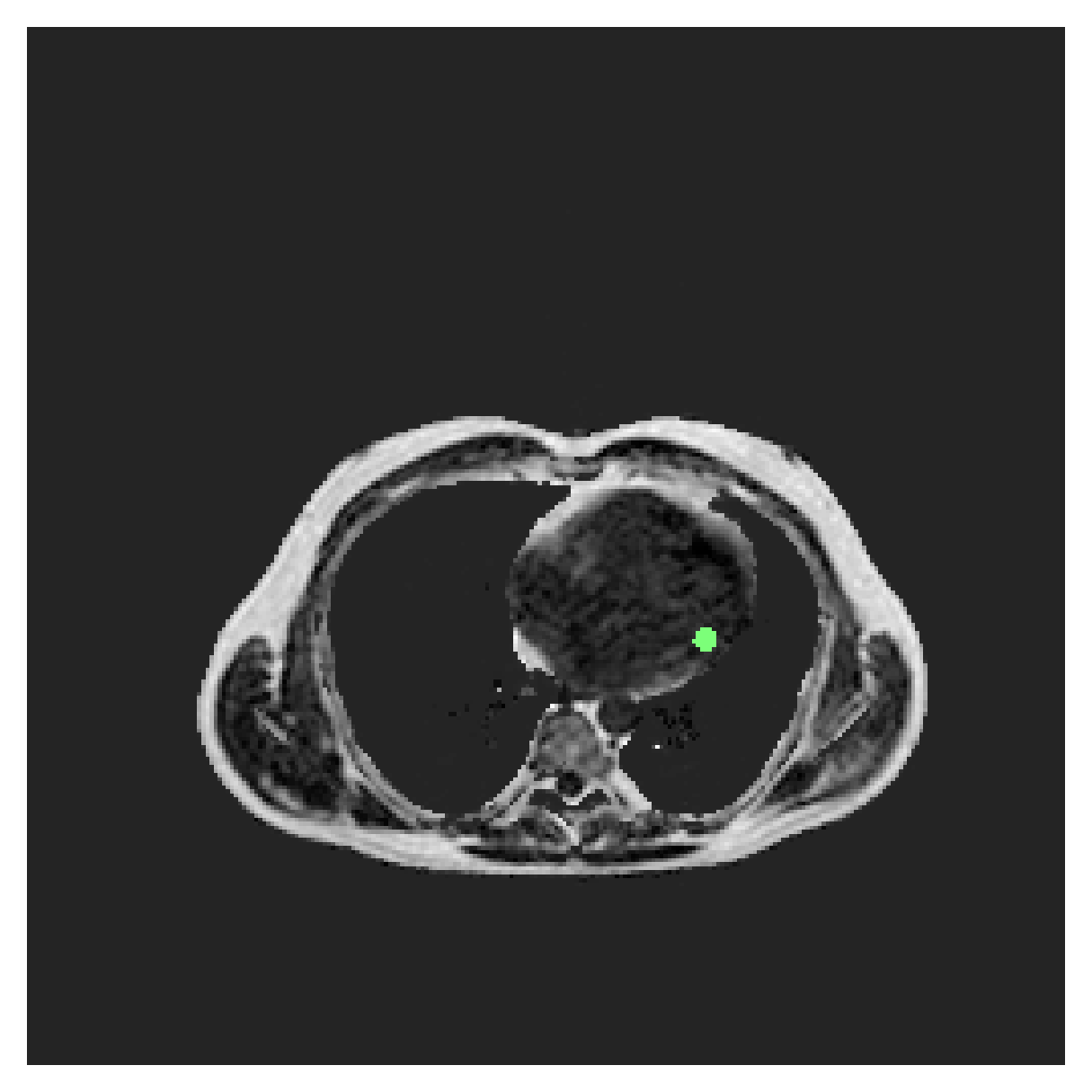}\\
                                \includegraphics[width=\textwidth]{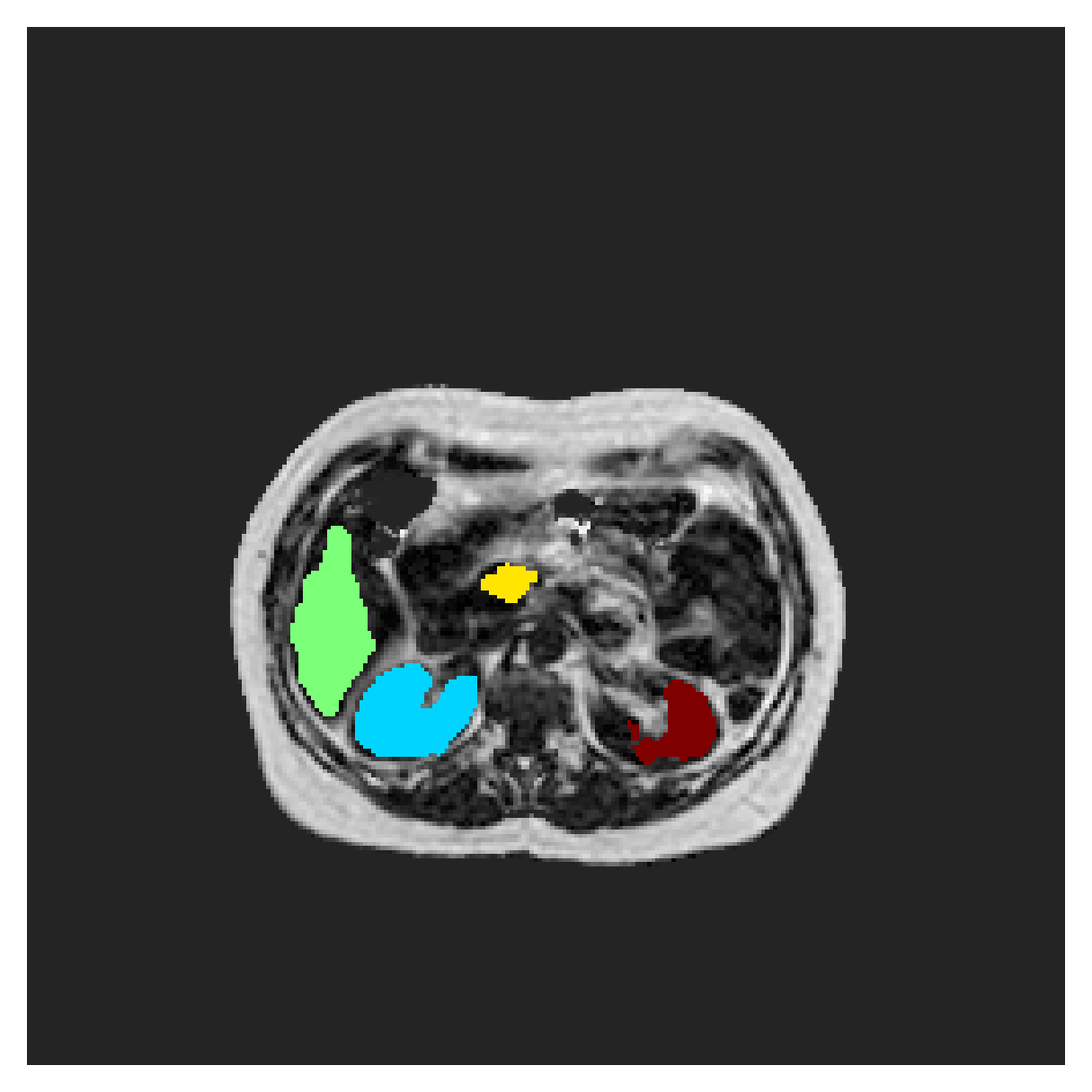}\\
                                \includegraphics[width=\textwidth]{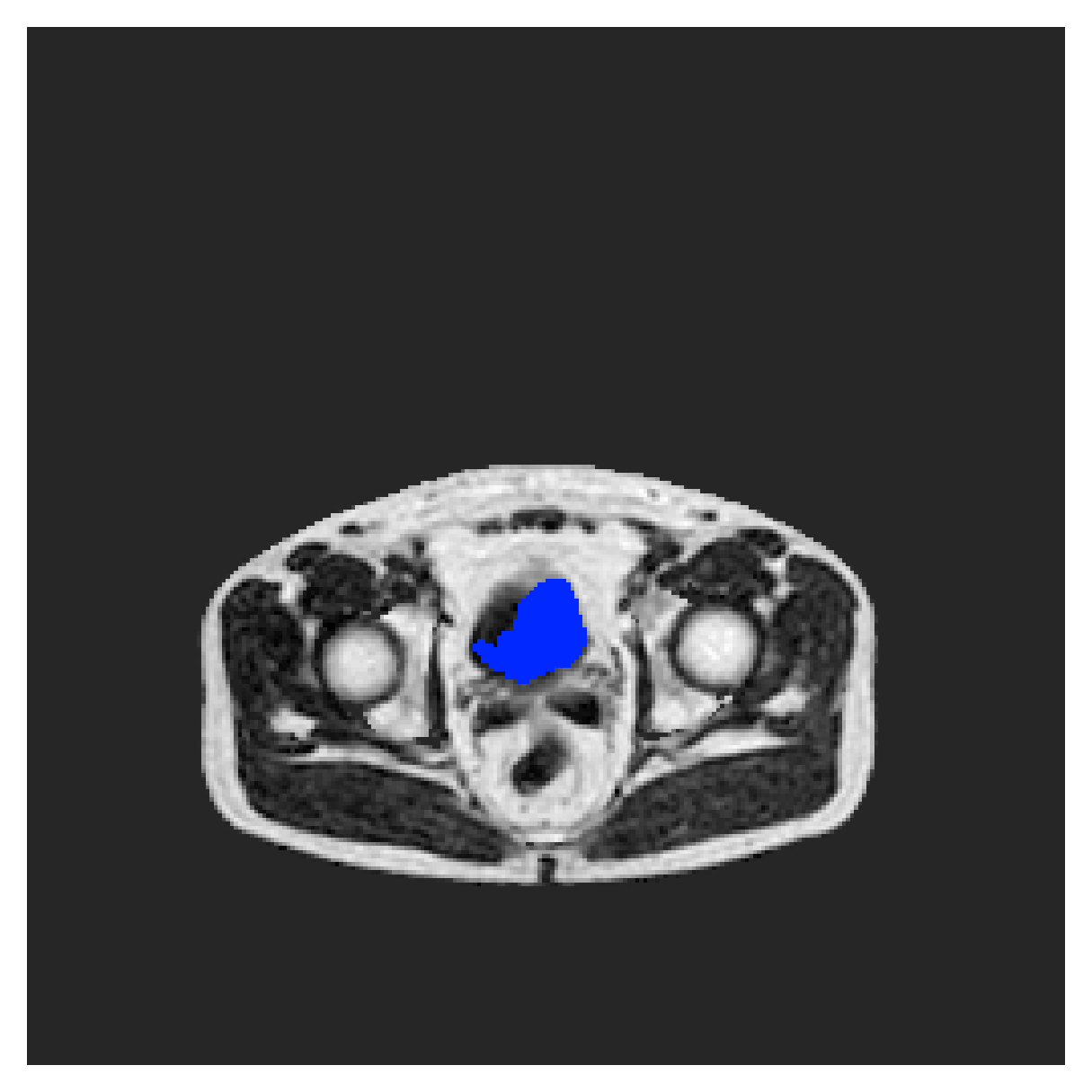}\\
                                \includegraphics[width=\textwidth]{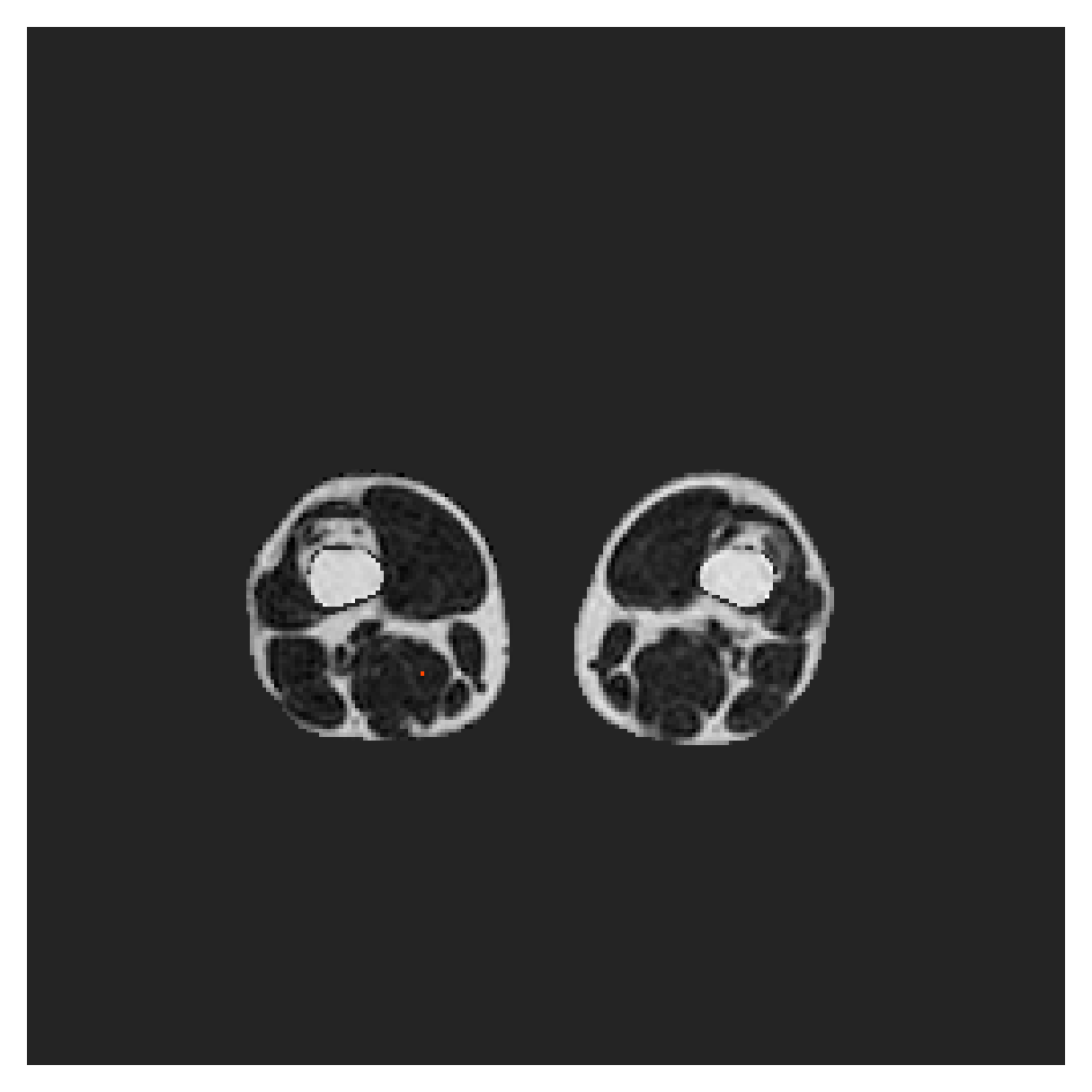}\\
                                \includegraphics[width=\textwidth]{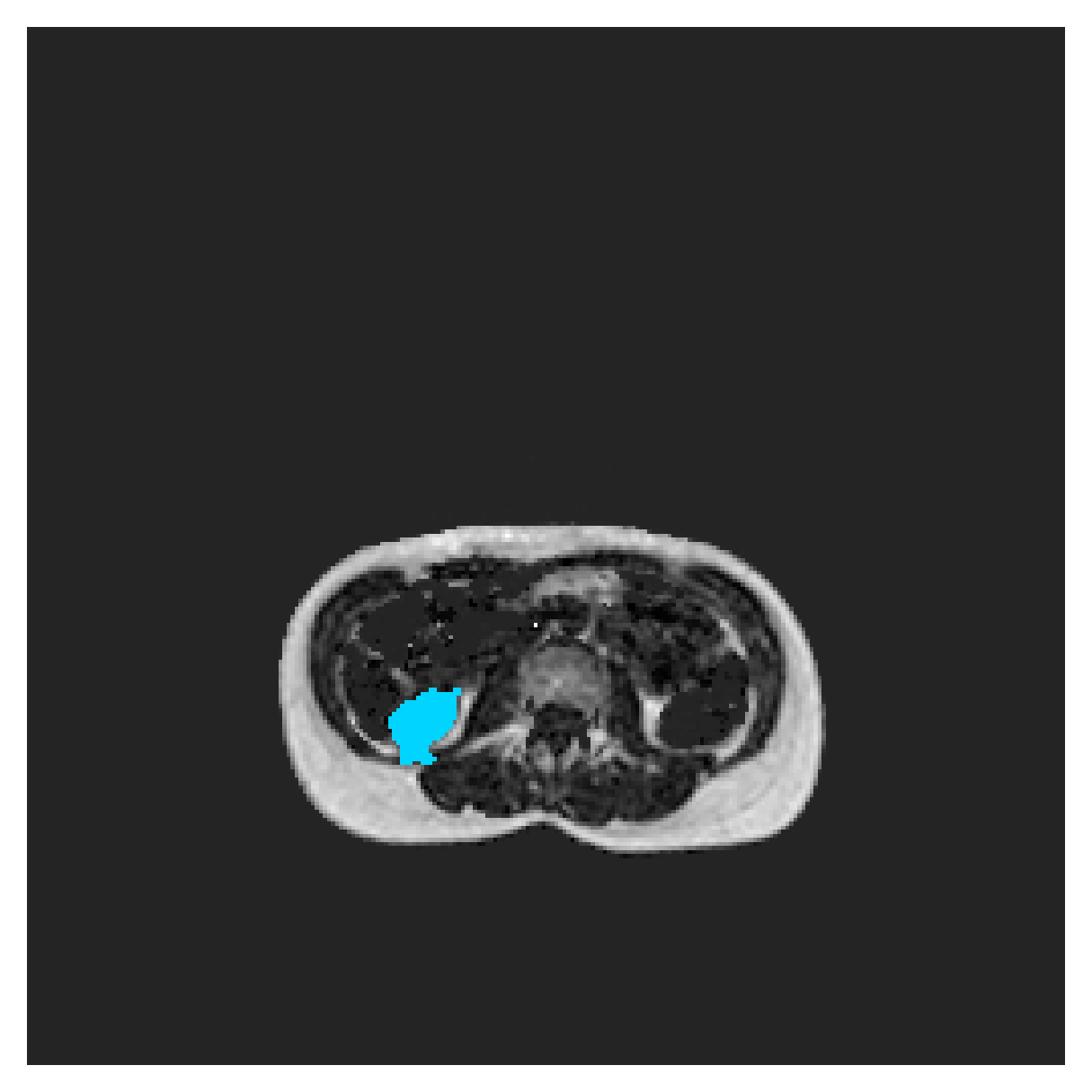}
                                \caption{$\mathcal L_{\widetilde{\text{CE}}} + \mathcal{L}_B^{int}$\\ $\qquad$}
                                \label{subfig:int1}
                        \end{subfigure}
                        \begin{subfigure}[b]{0.13\textwidth}
                                \centering
                                \includegraphics[width=\textwidth]{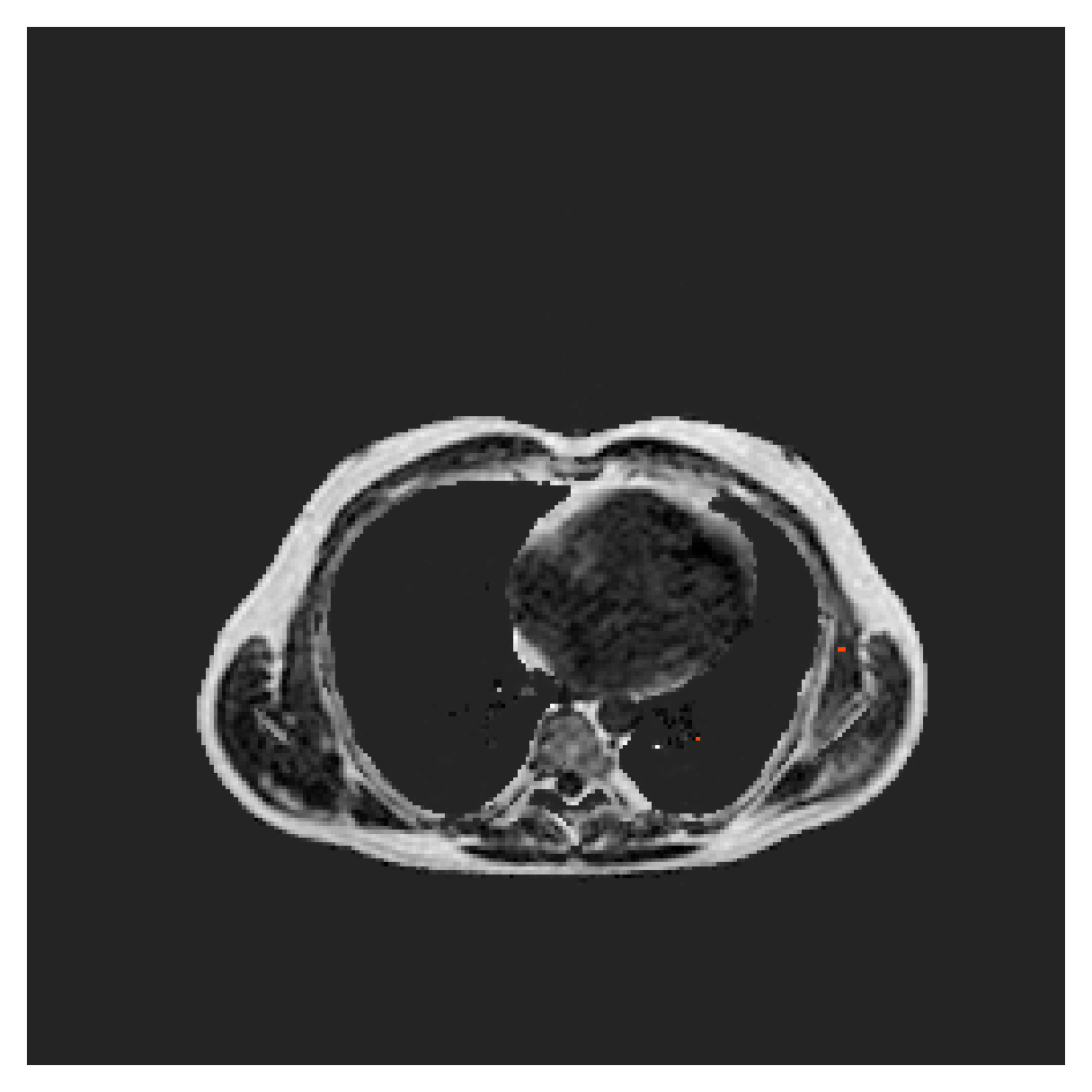}\\
                                \includegraphics[width=\textwidth]{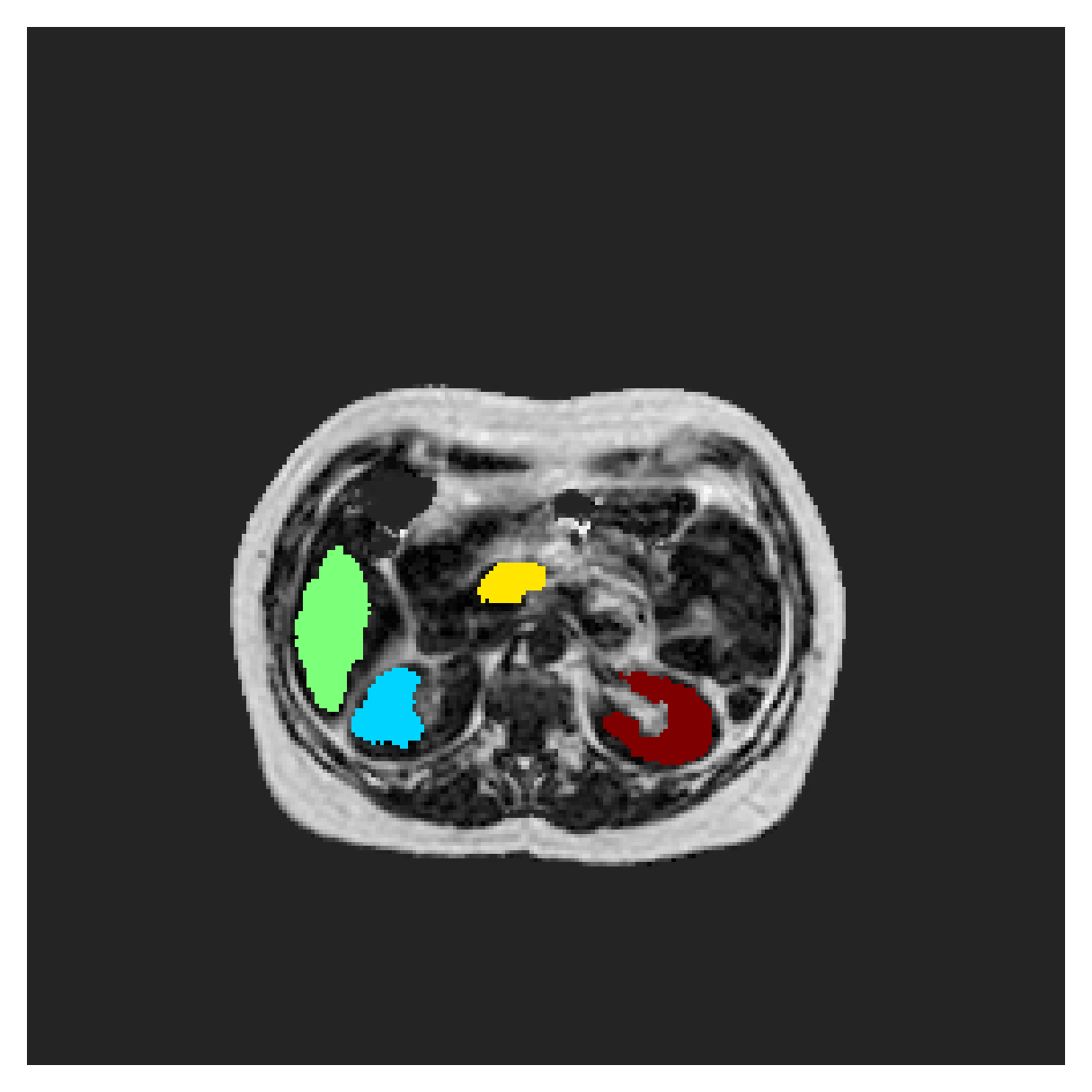}\\
                                \includegraphics[width=\textwidth]{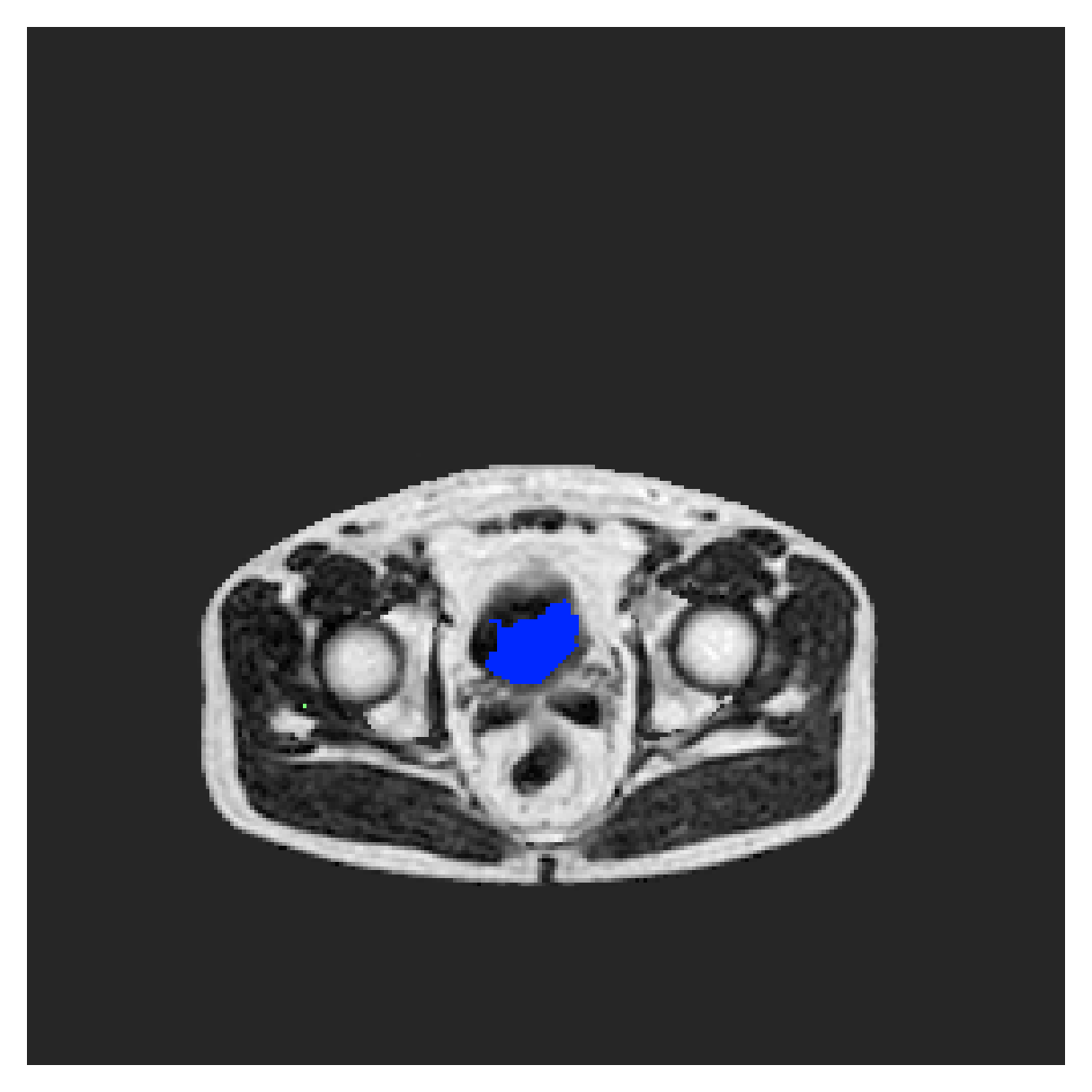}\\
                                \includegraphics[width=\textwidth]{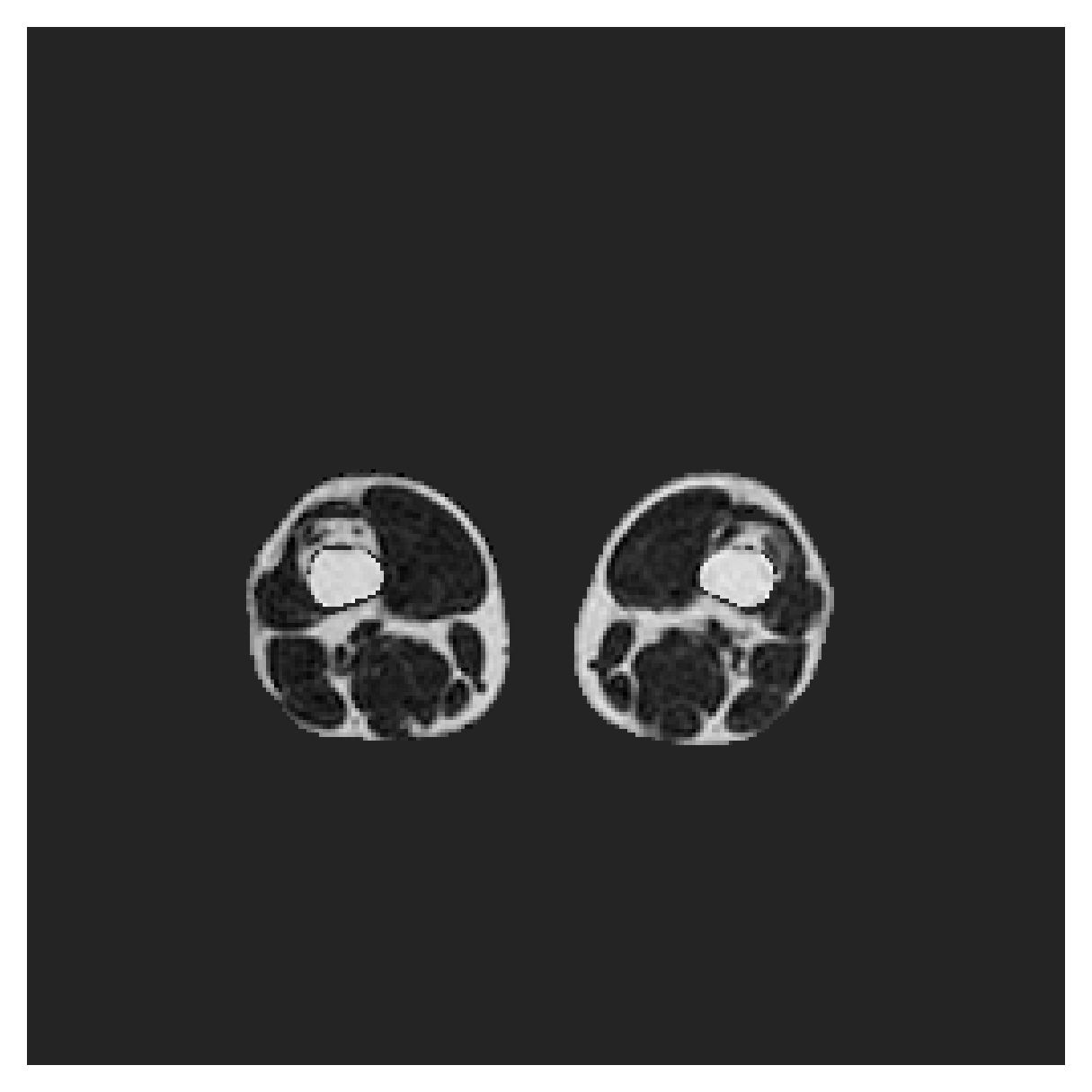}\\
                                \includegraphics[width=\textwidth]{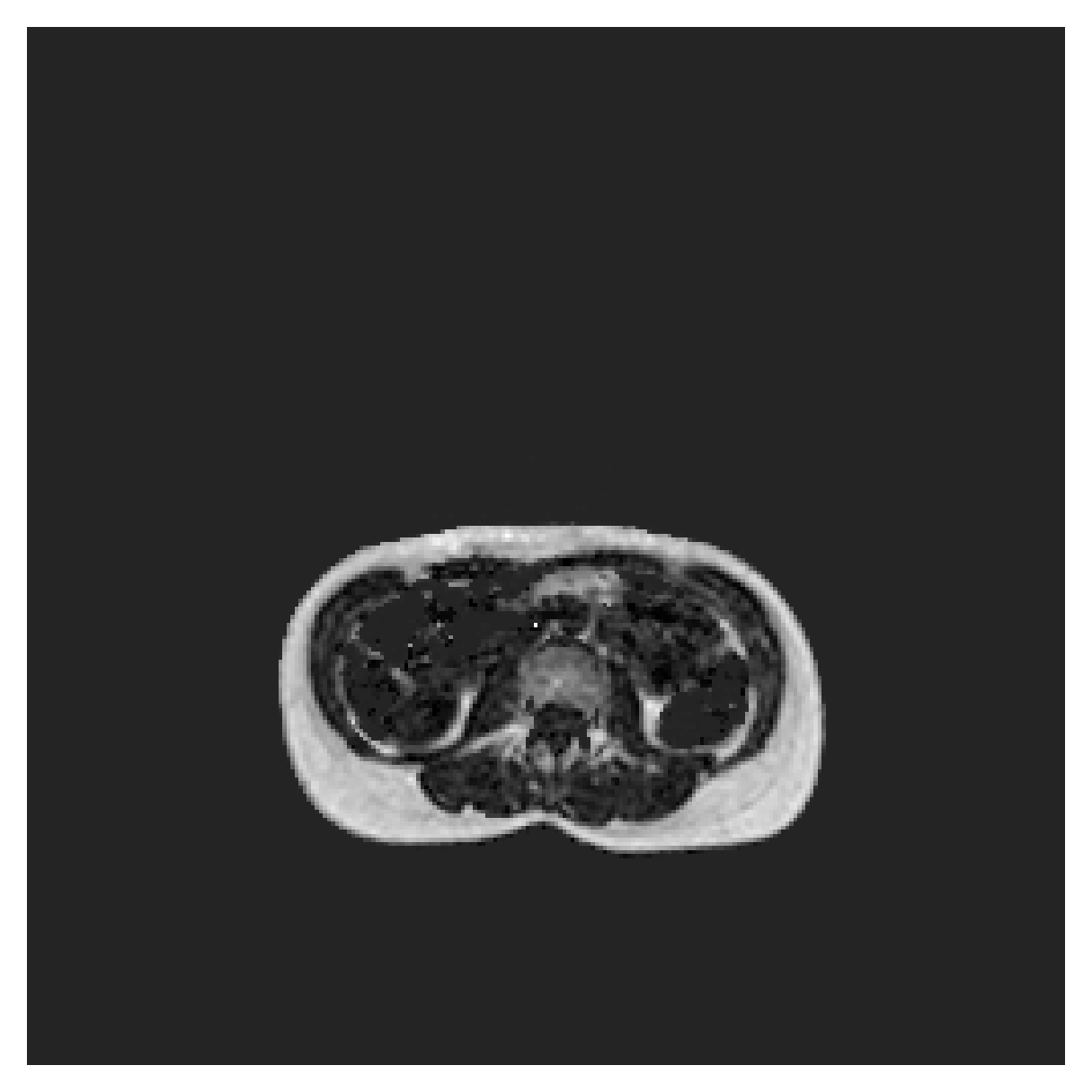}
                                \caption{$\mathcal L_{\widetilde{\text{CE}}} + \mathcal{L}_B^{mbd}$\\ $\qquad$}
                                \label{subfig:mbd1}
                        \end{subfigure}
                       \begin{subfigure}[b]{0.055\textwidth}
                                \centering
                                \includegraphics[width=\textwidth]{figures/poemtest/classes.pdf}\\
                                \vspace{0.5cm}
                                \caption*{}
                                
                        \end{subfigure}
                                
                        \caption{Example segmentations on the POEM test set. Training with $\mathcal{L}_\text{B}$ was done using  distance maps, computed in 3D. The colourmap and the slices shown are the same as the ones in Figure \ref{fig:POEMout}.}
                        \label{fig:POEMout2}
                \end{figure}